\documentclass[letterpaper]{article} 
\usepackage{aaai2026}  
\usepackage{times}  
\usepackage{helvet}  
\usepackage{courier}  
\usepackage[hyphens]{url}  
\usepackage{graphicx} 
\usepackage{natbib}  
\usepackage{caption} 
\usepackage{algorithm}
\usepackage{algorithmic}

\usepackage{newfloat}
\usepackage{listings}
\DeclareCaptionStyle{ruled}{labelfont=normalfont,labelsep=colon,strut=off} 
\floatstyle{ruled}
\newfloat{listing}{tb}{lst}{}
\floatname{listing}{Listing}
\usepackage{amsmath, amssymb}
\usepackage{array}
\usepackage{booktabs}
\usepackage{multirow}
\usepackage{subcaption}
\usepackage{pgfplots}
\pgfplotsset{compat=1.18}

\title{Gaussian Blending: Rethinking Alpha Blending in 3D Gaussian Splatting}
\author {
    Junseo Koo, 
    Jinseo Jeong, 
    Gunhee Kim 
}
\affiliations {
    Seoul National University\\
    junseo.koo@vision.snu.ac.kr, jinseo.jeong@vision.snu.ac.kr, gunhee@snu.ac.kr\\
    \url{https://1207koo.github.io/html/gaussianblending/} 
}

\begin{document}

\maketitle

\begin{abstract}
    The recent introduction of 3D Gaussian Splatting (3DGS) has significantly advanced novel view synthesis.
    Several studies have further improved the rendering quality of 3DGS, yet they still exhibit noticeable visual discrepancies when synthesizing views at sampling rates unseen during training.
    Specifically, they suffer from (i) erosion-induced blurring artifacts when zooming in and (ii) dilation-induced staircase artifacts when zooming out.
    We speculate that these artifacts arise from the fundamental limitation of the alpha blending adopted in 3DGS methods.
    Instead of the conventional alpha blending that computes alpha and transmittance as scalar quantities over a pixel, we propose to replace it with our novel \textit{Gaussian Blending} that treats alpha and transmittance as spatially varying distributions.
    Thus, transmittances can be updated considering the spatial distribution of alpha values across the pixel area, allowing nearby background splats to contribute to the final rendering.
    Our Gaussian Blending maintains real-time rendering speed and requires no additional memory cost, while being easily integrated as a drop-in replacement into existing 3DGS-based or other NVS frameworks.
    Extensive experiments demonstrate that Gaussian Blending effectively captures fine details at various sampling rates unseen during training, consistently outperforming existing novel view synthesis models across both unseen and seen sampling rates.
\end{abstract}


\section{Introduction}

\begin{figure}[ht!]
    \centering
    \renewcommand{\arraystretch}{0.0}
    \begin{tabular}{@{}c@{}c@{\hskip 2pt}c@{\hskip 2pt}c}
         & {3DGS} & {\shortstack{Analytic \\Splatting}} & {\shortstack{Gaussian\\Blending}} \\ [3pt]
        \raisebox{3.5em}{\rotatebox[origin=c]{90}{Zoom-in $\times 8$}}
         & \includegraphics[width=0.145\textwidth]{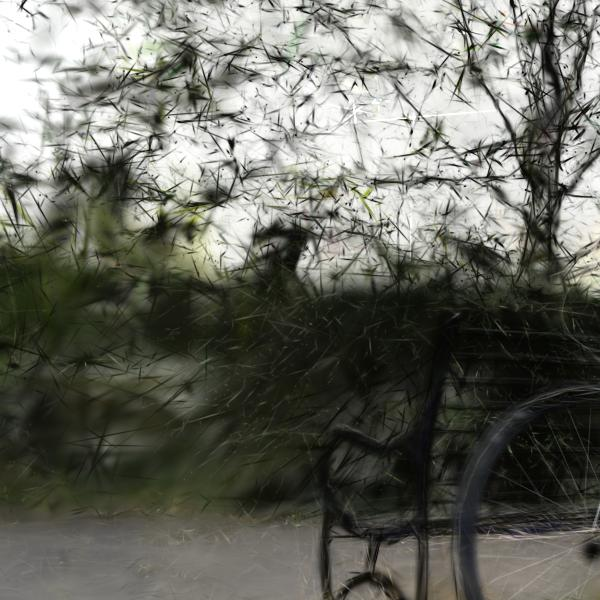}
         & \includegraphics[width=0.145\textwidth]{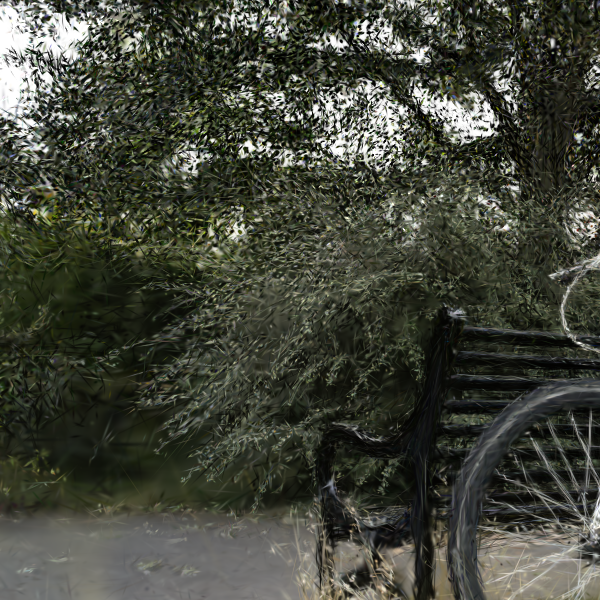}
         & \includegraphics[width=0.145\textwidth]{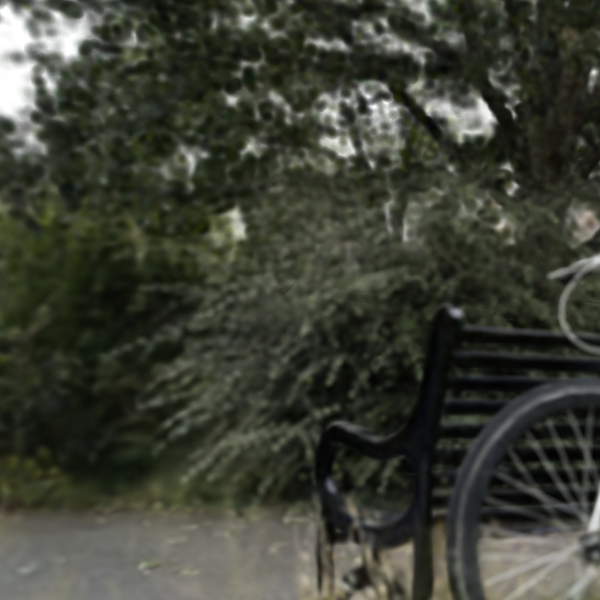}                                            \\ [0.7pt]
        \raisebox{3.4em}{\rotatebox[origin=c]{90}{Zoom-out $\times 1/8$}}
         & \includegraphics[width=0.145\textwidth]{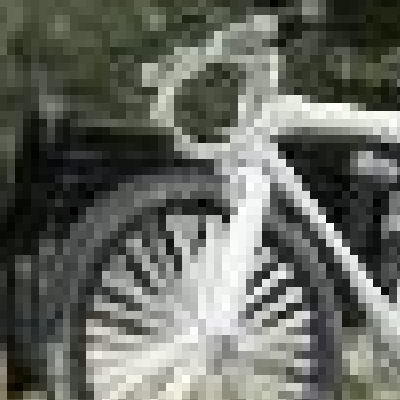}
         & \includegraphics[width=0.145\textwidth]{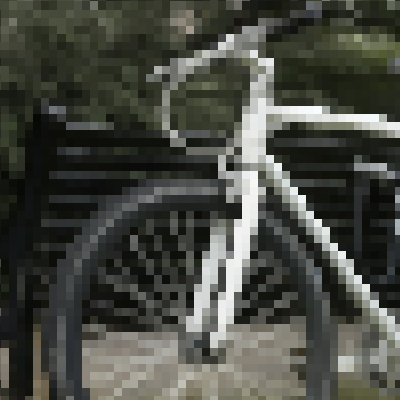}
         & \includegraphics[width=0.145\textwidth]{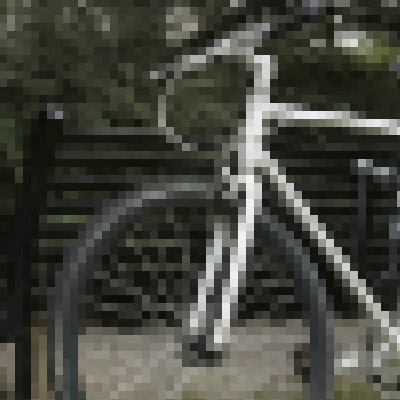}                             \\ [3pt]
    \end{tabular}
    \begin{tabular}{@{}c@{\hskip 3pt}c@{\hskip 1pt}c}
        \raisebox{2.2em}{\multirow{2}{*}{\rotatebox[origin=c]{90}{Reference}}}
         & \multirow{2}{*}[5.53em]{\includegraphics[width=0.3318\textwidth]{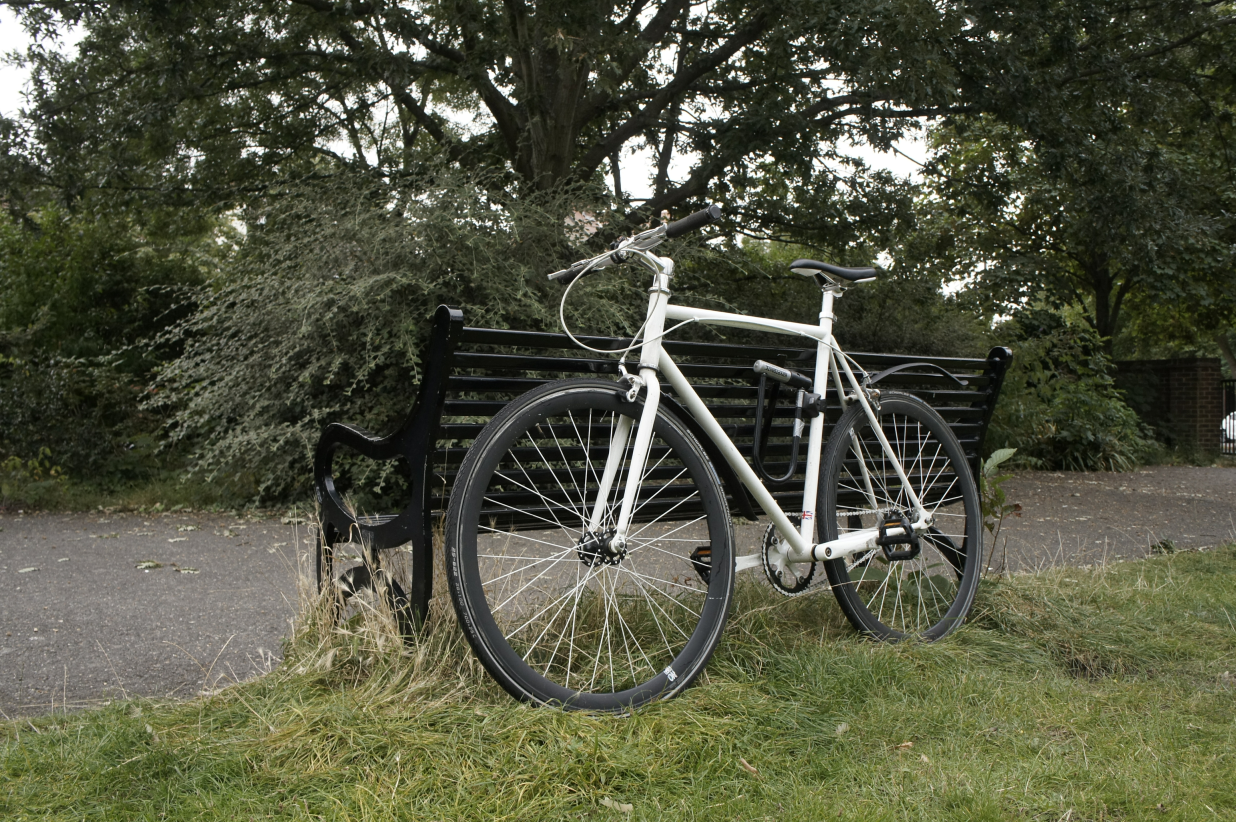}}
         & \includegraphics[width=0.10925\textwidth]{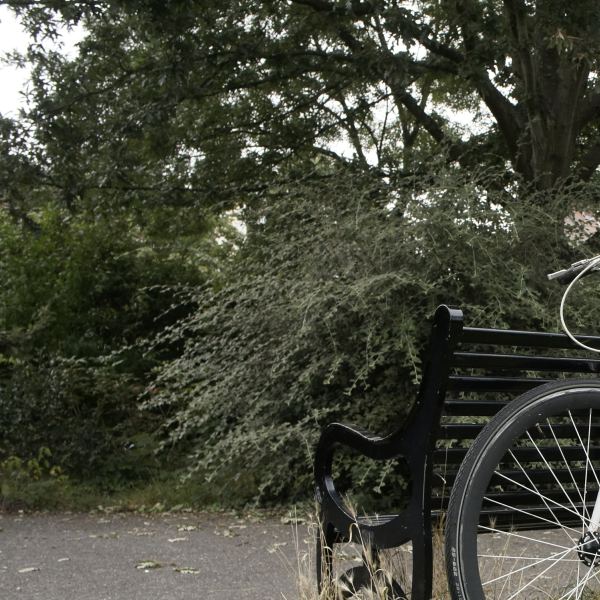} \\ [1pt]
         & & \includegraphics[width=0.10925\textwidth]{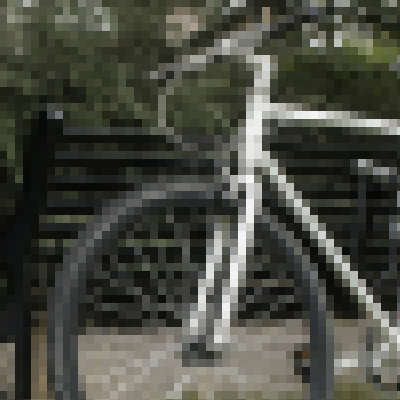}
    \end{tabular}
    \caption{
        Scalar alpha blending used in previous novel view synthesis methods (e.g., 3DGS and Analytic Splatting) exhibits aliasing artifacts when rendering at different sampling rates.
        Specifically, erosion with noisy edges occurs during zoom-in (top), while dilation with staircase artifacts occurs during zoom-out (bottom).
        In contrast, our Gaussian Blending produces consistent synthesis results at unseen sampling rates (e.g., leaves and bicycle frame) without any priors and additional training.
    }
    \label{fig:dilation}
\end{figure}

Novel view synthesis (NVS) has advanced rapidly, playing a pivotal role across diverse content generation tasks.
A key milestone was Neural Radiance Field (NeRF) \cite{nerf}, an implicit neural representation employing a neural network to estimate volumetric density and view-dependent radiance of 3D points. 
However, NeRF suffers from slow rendering speeds due to intensive ray marching.
Recently, 3D Gaussian Splatting (3DGS) \cite{3dgs} explicitly represents 3D scenes using Gaussian splats, enabling faster and finer view synthesis results. 

In the 3DGS framework, Gaussian splats are projected onto the 2D image plane, followed by the alpha blending to render the final pixel color.
However, 3DGS and its variants still suffer from noticeable artifacts when synthesizing views at unseen sampling rates (e.g., zoom-in or zoom-out).
This persistent challenge arises since existing methods treat pixels as points rather than planes, which leads to aliasing effects such as boundary erosion and dilation (see Figure~\ref{fig:dilation}).

In real-world camera systems, a sensor integrates radiances within a pixel area, whereas existing NVS methods use single ray for rendering a pixel color. 
According to the Nyquist–Shannon sampling theorem \cite{shannon}, suppressing high-frequency components at lower sampling rates can alleviate this issue, as supported by several prefiltering methods \cite{mip_nerf, mip_nerf_360, mip_splatting}.
More recently, Analytic-Splatting \cite{analytic_splatting} explicitly integrates Gaussian splat responses over pixel regions, better emulating physical camera sensors.

Despite recent advances, existing NVS approaches still exhibit noticeable aliasing, especially at object boundaries.
As illustrated in Figure~\ref{fig:dilation}, rendering at higher sampling rates (zoom-in) produces boundary erosion with blurry edges, while lower sampling rates (zoom-out) produces boundary dilation as staircase artifacts.
Even Analytic-Splatting suffers from these issues due to reliance on scalar alpha blending. 
All NVS methods, including various anti-aliasing methods, use scalar alpha values computed along single rays.
This inherently leads foreground splats (or density points in NeRF-based methods) to fully occlude background splats, disregarding their spatial overlap in the 2D screen space.
Consequently, the spatial contribution of background splats is inaccurately computed, causing persistent aliasing.
To mitigate this issue, existing NVS methods typically require retraining models at specific sampling rates, which is inefficient and impractical when training data are unavailable.

To address this limitation, we introduce \textit{Gaussian Blending} as a novel rendering methodology.
Rather than modeling alpha and transmittance as single scalar values, our approach models them as spatially varying functions across the 2D pixel area.
Our key insight is that Gaussian splats, being smooth and continuous, naturally cluster to represent distinct scene structures.
Leveraging this spatial coherence, we approximate their combined transmittance using simplified 2D uniform distributions within each pixel region.
This approximation enables efficient, dynamic updating of spatial transmittance distributions during the alpha blending process via distributional moments.

In experiments on the multi-scale Blender \cite{nerf,mip_nerf} and Mip-NeRF 360 datasets \cite{mip_nerf_360}, Gaussian Blending significantly reduces intra-pixel aliasing, mitigating erosion and dilation at previously unseen sampling rates.
Moreover, Gaussian Blending achieves real-time rendering speeds comparable to the original 3DGS method via optimized CUDA implementation, improving visual fidelity without extra memory overhead.
In addition, it can be seamlessly integrated as a drop-in replacement for the rendering kernels of existing NVS frameworks.

Finally, our key contributions are as follows.
\begin{itemize}
    \item We revisit existing scalar alpha blending methods and identify that aliasing arises from inadequate handling of the spatial variations within pixel regions.
    \item To the best of our knowledge, our work is the first to integrate intra-pixel anti-aliasing into the alpha blending process explicitly. Our proposed \textit{Gaussian Blending} efficiently models and dynamically tracks spatial variations within pixels to effectively suppress aliasing.
    \item Extensive experiments demonstrate that Gaussian Blending significantly reduces aliasing, yielding higher-quality synthesized views at both unseen and seen sampling rates during training, without additional priors or retraining, all while preserving real-time performance. Our Gaussian Blending can be easily integrated as a drop-in replacement into existing NVS frameworks.
\end{itemize}

\begin{figure*}[t]
    \centering
    \includegraphics[width=0.98\textwidth]{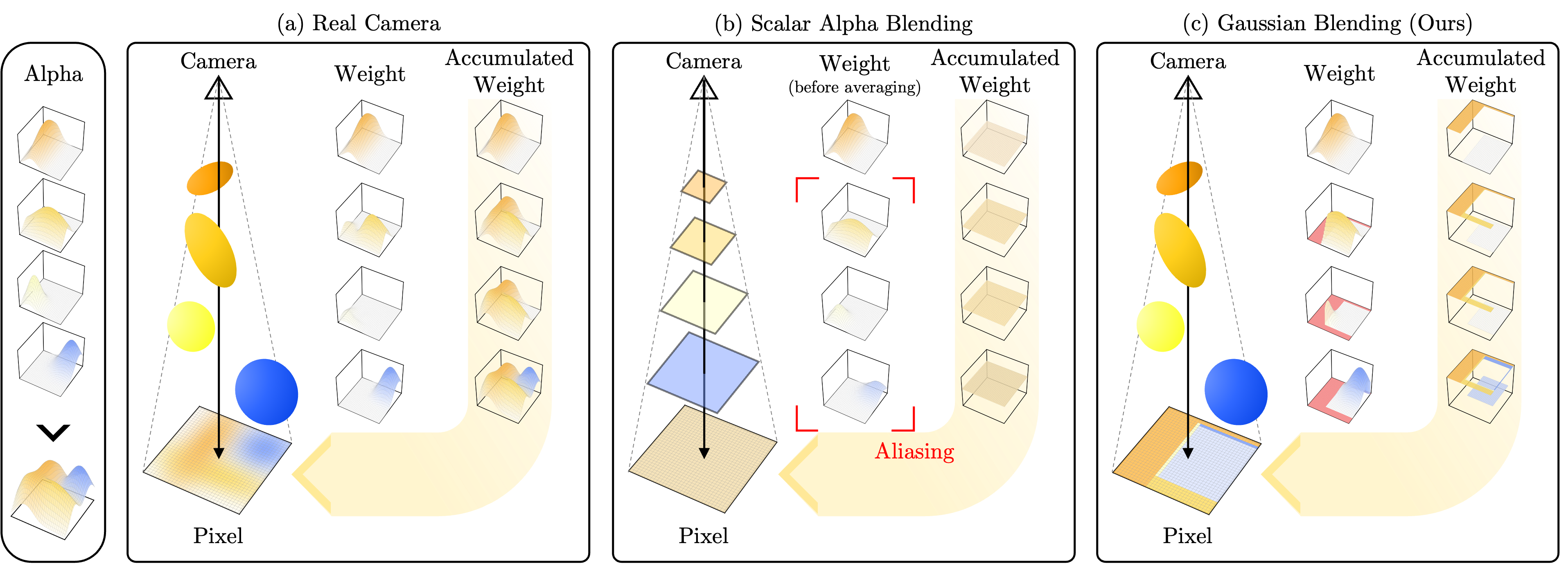} \\
    \caption{
        Comparison of blending behaviors when rendering overlapping splats. 
        (a) In a real camera system, spatial occlusion is properly handled—the yellow splat is attenuated by closer splats, while the blue splat, which does not overlap in screen space, retains high transmittance and remains visible. 
        (b) Scalar alpha blending ignores spatial occlusion by averaging alpha values across the pixel, causing dilation of the overlapping yellow splat and undesired suppression of the non-overlapping blue splat. 
        (c) Our Gaussian Blending dynamically adjusts the transmittance window within each pixel, preserving high transmittance in regions without splats and effectively maintaining the visibility of the blue splat while reducing the influence of the occluded yellow splat.
    }
    \label{fig:rendering}
\end{figure*}

\section{Related Work}

\subsection{Novel View Synthesis (NVS)}
NVS aims to reconstruct the geometry and appearance of 3D scenes from multi-view images, enabling rendering from unseen viewpoints \cite{d_nerf, hypernerf, 4dgs, esr_nerf, gs_ir, gs_rt}.
While Neural Radiance Fields (NeRF) \cite{nerf} have laid the groundwork, they suffer from slow rendering speeds due to intensive ray marching.
To overcome these limitations, subsequent methods adopt explicit grid-based scene representations \cite{dvgo, tensorf}, significantly enhancing rendering efficiency through strategies like free-space skipping.
Hybrid methods further combine NeRF with Signed Distance Functions (SDF) \cite{neus, voxurf}, extending implicit representations to support accurate surface reconstruction.

3D Gaussian Splatting (3DGS) \cite{3dgs} represents scenes as collections of Gaussian splats, enabling real-time, high-quality rendering.
Unlike implicit neural methods involving computationally heavy ray marching, 3DGS rasterizes Gaussian splats directly into screen space, allowing efficient rendering via GPU-optimized rasterization with explicit alpha blending.
Building upon this framework, several variants have been proposed to further enhance rendering quality and efficiency.
2D Gaussian Splatting (2DGS) \cite{2dgs} and 3D Half-Gaussian Splatting (3D-HGS) \cite{3d_hgs} respectively introduce 2D Gaussian splats and half-Gaussian kernels, to better capture geometric details.
Scaffold-GS and Octree-GS \cite{scaffold_gs, octree_gs} employ anchor-based or Level-of-Detail-structured 3D Gaussians to reduce redundancy and enhance efficiency in large-scale view-varying scenarios. 
These advances have broadened the applicability of NVS, including dynamic scene reconstruction \cite{d_nerf, hypernerf, 4dgs}, generative modeling \cite{dreamfusion, instruct_nerf2nerf, diffsplat}, and inverse rendering tasks \cite{esr_nerf, gs_ir, gs_rt}, inspiring numerous downstream developments.

\subsection{NVS at Different Sampling Rates}
Despite rapid progress in NVS, most existing methods assume fixed resolutions, camera distances, and camera intrinsics.
However, real-world applications often involve varying sampling rates, causing noticeable aliasing artifacts due to single ray rendering that neglects the visible frustum.

Several methods have addressed this issue through prefiltering-based anti-aliasing approaches following the Nyquist–Shannon Sampling Theorem \cite{shannon}.
For example, Mip-NeRF \cite{mip_nerf} reduces aliasing by integrating positional encoding over spatial regions, effectively applying implicit low-pass filtering.
Mip-NeRF 360 \cite{mip_nerf_360} further extends this approach to handle unbounded scenes, addressing aliasing more effectively in realistic scenarios.
With the emergence of 3DGS,
Mip-Splatting \cite{mip_splatting} combines 2D and 3D filtering strategies, mitigating dilation artifacts caused by varying sampling rates.
Going beyond these filtering-based approaches, Analytic-Splatting \cite{analytic_splatting} analytically computes Gaussian splat responses over pixel areas, explicitly integrating splat contributions to achieve effective and accurate anti-aliasing.

Mipmap-based approach is another popular strategy to handle anti-aliasing. They often train separate scene representations for each sampling rate.
Instant-NGP \cite{instant_ngp} pioneers resolution-adaptive rendering through multi-resolution hash-grid features.
Tri-MipRF \cite{tri_miprf} introduces learnable mipmap representations, dynamically retrieving resolution-specific features during rendering.
Recent adaptations \cite{mipmap_gs, ms_gs} group Gaussian splats based on sampling rates.
However, these mipmap-based approaches can only handle specific resolutions seen during training, limiting their generalization.

Additionally, several studies focus on enhancing high-frequency details or resolving artifacts when rendering at higher sampling rates beyond training conditions.
While straightforward techniques such as supersampling show limited effectiveness \cite{nerf_sr}, recent approaches leverage generative diffusion models \cite{gaussiansr, cu_gs} or single-image super-resolution (SISR) frameworks \cite{srgs} to generate pseudo-ground-truth details of test-time resolutions.
Unlike these methods that improve image quality by generating high-frequency details, our approach directly addresses aliasing artifacts caused by scalar alpha blending.

\section{Approach}

\subsection{Preliminaries}
3D Gaussian Splatting (3DGS) \cite{3dgs} explicitly represents a 3D scene using a set of Gaussian splats.
Given a total of $N$ splats, the $i$-th closest splat to the camera is parameterized by its center position $\mu_i \in \mathbb{R}^{3}$, scale vector $s_i \in \mathbb{R}^{3}$, rotation quaternion $r_i \in \mathbb{R}^{4}$, and opacity value $o_i \in \mathbb{R}$.
The scale vector $s_i$ forms a diagonal scale matrix $S_i \in \mathbb{R}^{3\times 3}$, and the quaternion $r_i$ defines a rotation matrix $R_i \in \mathbb{R}^{3\times 3}$.
Consequently, the 3D covariance matrix of the Gaussian splat is computed as
\begin{equation}
    \Sigma_i = R_i S_i S_i^\top R_i^\top.
\end{equation}
The influence of the $i$-th Gaussian splat at a 3D position $x$ is defined as
\begin{equation}
    G_i(x) = o_i \exp\left(-\frac{1}{2}(x-\mu_i)^\top \Sigma_i^{-1}(x-\mu_i)\right).
\end{equation}

Each 3D Gaussian splat is projected onto the 2D screen space, resulting in the corresponding 2D Gaussian splat.
Given a camera extrinsic matrix $W$ and an intrinsic projection matrix $K$, the projected mean vector and covariance matrix are
\begin{equation}
    \mu^{\prime}_i = K W [\mu_i, 1]^\top, \quad \Sigma^{\prime}_i = J_i W \Sigma_i W^\top J_i^\top,
\end{equation}
where $J_i$ is the Jacobian of the local affine approximation of the perspective projection at $\mu_i$.
By removing its third row and column, $\Sigma^{\prime}_i$ becomes a 2D covariance matrix in $\mathbb{R}^{2\times2}$.
Thus, the screen space influence of the projected Gaussian splat at a 2D position $x$ is
\begin{equation}
    G^{\prime}_i(x) = o_i \exp\left(-\frac{1}{2}(x-\mu^{\prime}_i)^\top \Sigma_i^{\prime -1}(x-\mu^{\prime}_i)\right).
\end{equation}

Given a pixel center $p$, each splat’s alpha value is computed as the Gaussian influence evaluated at the pixel center $G^{\prime}_i(p)$.
Rendering is performed by blending these splats in the front-to-back depth order using alpha blending.
As $c_i$ is the view-dependent color of the $i$-th splat, the final pixel color $C_p$ is computed as
\begin{equation}
    C_p = \sum_{i=1}^{N} c_i G^{\prime}_i(p) T_i, \quad T_i = \prod_{j=1}^{i-1}(1 - G^{\prime}_j(p)),
\end{equation}
where the contribution of each splat can be interpreted as the weight $w_i = G^{\prime}_i(p) T_i$.

Analytic-Splatting \cite{analytic_splatting} extends the 3DGS approach by incorporating anti-aliasing through analytic integration of each Gaussian splat’s influence over the entire pixel area, rather than evaluating it only at the pixel center.
This integral of 2D Gaussian cannot be simplified with a closed form; thus, Analytic-Splatting employs an efficient approximation, which
performs an eigen-decomposition on each 2D Gaussian splat to identify principal axes.
Then, the pixel coordinate frame is rotated to align with these axes, eliminating the correlation term in the Gaussian function.
Consequently, the integral can be factorized into two separable 1D Gaussian integrals along these axes, each computed analytically.
Further details can be found in the original papers \cite{3dgs, analytic_splatting}.

\subsection{Scalar Alpha Blending}

Although Analytic-Splatting theoretically eliminates aliasing artifacts, as illustrated in Figure~\ref{fig:dilation}, it still exhibits persistent aliasing issues like erosion and dilation. 
We identify that these problems fundamentally originate from scalar alpha blending, and subsequently describe how our proposed approach overcomes this limitation.

Alpha blending is a fundamental rendering technique used by most NVS methods, including anti-aliasing approaches.
However, conventional alpha blending represents alpha and transmittance as scalar values, ignoring critical spatial variations within pixels.

Physically, rendering should involve integrating spatially varying alpha and color distributions over each pixel region.
The physically correct pixel color $C_{p}^{p}$ is computed by integrating the contributions of all splats and resulting transmittance $T_i^{p}$ within the pixel area $p$:
\begin{align}
    C_{p}^{p}  & = \int_{p} \sum_{i=1}^{N} T_i^{p}(x)\alpha_i(x)c_i dx, \notag \\
    T_i^{p}(x) & = \prod_{j=1}^{i-1}\left(1 - \alpha_j(x)\right).
\end{align}
Existing methods approximate this integration by computing scalar alpha integrals independently for each splat:
\begin{align}
    C_{p}^{s} & = \sum_{i=1}^{N} T_i^{s} c_i\int_{p}\alpha_i(x) dx, \notag \\
    T_i^{s}   & = \prod_{j=1}^{i-1}\left(1 - \int_{p}\alpha_j(x) dx\right),
\end{align}
where $C_{p}^{s}$ is the approximated pixel color, and $T_i^{s}$ is the scalar transmittance after rendering the $i$-th splat.


Figure~\ref{fig:rendering} illustrates that such scalar approximations inevitably lead to aliasing artifacts. 
In a real camera system, overlapping splats that lie behind other splats should attenuate their alpha values due to spatial occlusion. 
However, the scalar approximation in alpha and transmittance ignores the spatial context such as occlusions, computing alpha values regardless of the spatial occlusion. 
This results in the object edges, which partially overlap within the pixel area, being rendered with higher alpha values than they should be, even though object splats are highly occluding each other.
This rapidly saturates transmittance, leading to pixel-level dilation artifacts,  where foreground objects appear artificially enlarged, and background objects (e.g., the blue splat in Figure~\ref{fig:rendering}) become overly occluded, as shown in Figure~\ref{fig:dilation}.
At lower sampling rates (zoom-out), saturated transmittance causes dilation artifacts, despite the broader visible frustum per pixel. 
Conversely, at higher sampling rates (zoom-in), scalar alpha blending induces erosion artifacts with blurry boundaries due to dilation bias learned during training.

This limitation persists across all existing NVS approaches, including standard rendering, and prefiltering or mipmap-based anti-aliasing, since all rely on scalar alpha blending.
Only supersampling approaches, which explicitly compute spatial variations in alpha and transmittance, accurately capture these effects.
However, supersampling is computationally expensive, requiring multiple rendering passes, and is thus challenging to employ in real-time applications.

\subsection{Efficient Spatial Alpha Blending}

\begin{figure*}[t]
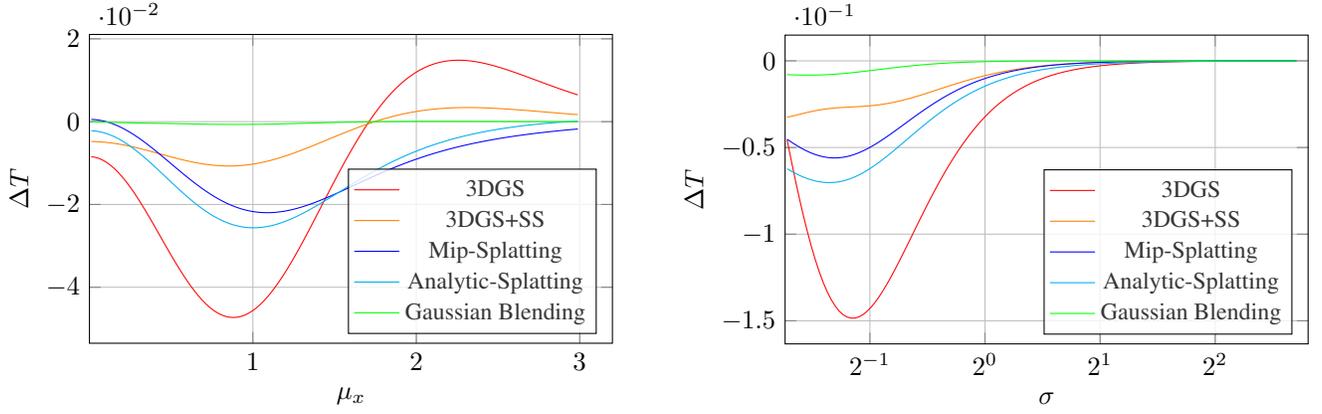

    \centering
    \setlength{\tabcolsep}{12pt}

    \caption{Comparison of the remaining transmittance error $\Delta T$ after rendering two splats symmetrically placed at $\mu_1 = [\mu_x, -0.1]^\top$ and $\mu_2 = [\mu_x, 0.1]^\top$, with $o_1 = o_2 = 1$ and $\sigma_1 = \sigma_2 = \sigma$. \textbf{(Left)} Varying $\mu_x$ while fixing $\sigma = 1$. \textbf{(Right)} Varying $\sigma$ while fixing $\mu_x = 0.5$. A negative $\Delta T$ indicates dilation, where a lower transmittance remains after rendering.} 
    \label{fig:error_analysis}
\end{figure*}

A straightforward and physically accurate approach to spatial variations in transmittance is to integrate the transmittance $T_i$ and the corresponding weight $w_i$ across each pixel's 2D screen space coverage.
However, computing these integrals requires exponential computational complexity, as the transmittance $T_i = \prod_{j=1}^{i-1}(1 - \alpha_j)$ and the weight $w_i = \alpha_i T_i$ consist of $2^{i-1}-1$ and $2^{i-1}$ terms, respectively.

Our key insight to mitigate this computational challenge is that Gaussian splats collectively represent continuous object surfaces, typically forming uniform alpha distributions across 2D screen space, such as  $\alpha = 1$ for opaque surfaces.
Thus, despite the transmittance involving an exponentially large number of Gaussian terms, these terms spatially cluster into compact regions, effectively approximated by a uniform distribution within each pixel area.
Leveraging this observation, we approximate the transmittance distribution using a simpler, spatially uniform 2D representation.

By adaptively controlling the spatial extent of this uniform distribution, we dynamically modulate the effective transmittance, reducing the influence of splats located in regions already occluded by previously rendered surfaces, where transmittance values are low.
This approach enables real-time rendering without compromising visual accuracy.

Moreover, representing transmittance as a 2D uniform distribution offers two physical advantages.
First, it inherently satisfies $0 \leq w_i \leq T_i$ throughout the pixel, ensuring physically consistent blending.
Second, the initial pixel state naturally conforms to a uniform distribution, simplifying the computation of subsequent rendering windows.

Formally, we represent transmittance $T_i$ by a window centered at $x_i \in \mathbb{R}^2$ with size $l_i=[l_{i, 1}, l_{i, 2}]^\top \in \mathbb{R}^2$ and uniform transmittance value $t_i \in \mathbb{R}$.
Initially, before any splats are rendered, $T_1$ is defined as $x_1=p$, $l_1=[1,1]^\top$, and $t_1=1$.
Assuming splats composing an object form uniform distributions in the pixel region, dynamically adjusting the window $(x_i, l_i)$ enables accurate representation of lower-transmittance subregions.

When rendering splat $\alpha_i$ onto distribution $T_i$, we compute two quantities:
(i) the integrated splat response within the current window region, $\int w_i(x)dx$, and
(ii) the optimal distribution $T_{i+1}$ approximating the remaining transmittance $T_i(x)(1-\alpha_i(x))$.

\textbf{Weight Computation}.
To compute a splat's response $\int w_i(x)dx$, we first perform eigen-decomposition on the 2D covariance matrix $\Sigma^{\prime}_i$, yielding eigenvalues $\lambda_1$, $\lambda_2$ and unit eigenvectors $e_1$, $e_2$, defining principal axes.
The standard deviations along these axes are $\sigma_1 = \sqrt{\lambda_1}$ and $\sigma_2 = \sqrt{\lambda_2}$.
Next, we rotate the current window $T_i$ slightly (within $45^\circ$) to align it with the Gaussian axes defined by $e_1$, $e_2$ to efficiently approximate the integral.
Without loss of generality, let us assume $\lambda_1, e_1$ align with the first axis, and $\lambda_2, e_2$ with the second axis.
In the coordinate system defined by the splat $\alpha_i$, centered at $\mu_i'$, the window center is located at $[u,v]^\top = [(x_i-\mu_i')\cdot e_1, (x_i-\mu_i')\cdot e_2]^\top$, and the window region spans $[u_1, u_2]\times[v_1, v_2]$ with boundaries:
\begin{align}
    \begin{bmatrix}u_1\\v_1\end{bmatrix} & = \begin{bmatrix}u\\v\end{bmatrix} - \frac{1}{2} l_i, \quad \begin{bmatrix}u_2\\v_2\end{bmatrix} = \begin{bmatrix}u\\v\end{bmatrix} + \frac{1}{2} l_i.
\end{align}

To simplify the integrals, we introduce the notation for the $k$-th order moments of the 1D Gaussian as $I^k_{\sigma}(a, b) = \int_{a}^{b} x^k \exp(-x^2 / 2 \sigma^2) dx$.
Then, we express the integrated weight as
\begin{align}
    \int w_i(x)dx & = t_i o_i I^0_{\sigma_1}(u_1, u_2) I^0_{\sigma_2}(v_1, v_2).
\end{align}

\begin{table*}[ht!]
    \centering
    \caption{Single-scale training ($\times 1$ resolution) and multi-scale testing results on the multi-scale Blender dataset. The best results are highlighted in \textbf{bold}.}
    \label{tab:stmt1_blender}
    \resizebox{\textwidth}{!}{
        \begin{tabular}{l|ccccc|ccccc|ccccc}
            \toprule
            \multirow{2}{*}{} & \multicolumn{5}{c|}{PSNR $\uparrow$} & \multicolumn{5}{c|}{SSIM $\uparrow$} & \multicolumn{5}{c}{LPIPS $\downarrow$}                                                                                                                                                                                                             \\
                                    & $\times 1$                           & $\times 1/2$                         & $\times 1/4$                           & $\times 1/8$   & Avg.
                                    & $\times 1$                           & $\times 1/2$                         & $\times 1/4$                           & $\times 1/8$   & Avg.
                                    & $\times 1$                           & $\times 1/2$                         & $\times 1/4$                           & $\times 1/8$   & Avg.                                                                                                                                                                                     \\
            \midrule
            TensoRF
                                    & 33.22                                & 33.27                                & 30.23                                  & 26.77          & 30.87          & 0.964          & 0.972          & 0.968          & 0.953          & 0.964          & 0.047          & 0.034          & 0.049          & 0.079          & 0.052          \\
            3DGS
                                    & 33.57                                & 27.04                                & 21.43                                  & 17.74          & 24.95          & 0.970          & 0.950          & 0.876          & 0.767          & 0.891          & 0.037          & 0.036          & 0.071          & 0.130          & 0.068          \\
            Scaffold-GS
                                    & 31.76                                & 27.46                                & 22.14                                  & 18.34          & 24.92          & 0.961          & 0.949          & 0.886          & 0.784          & 0.895          & 0.050          & 0.041          & 0.067          & 0.121          & 0.070          \\
            2DGS
                                    & 31.81                                & 26.73                                & 20.21                                  & 16.41          & 23.79          & 0.965          & 0.946          & 0.846          & 0.711          & 0.867          & 0.048          & 0.052          & 0.102          & 0.177          & 0.095          \\
            3D-HGS
                                    & 33.70                                & 27.34                                & 21.71                                  & 17.99          & 25.19          & 0.970          & 0.953          & 0.883          & 0.777          & 0.896          & 0.037          & 0.035          & 0.067          & 0.124          & 0.066          \\
            \midrule
            Tri-MipRF
                                    & 32.86                                & 32.80                                & 28.44                                  & 24.22          & 29.58          & 0.959          & 0.968          & 0.955          & 0.924          & 0.951          & 0.056          & 0.039          & 0.051          & 0.075          & 0.055          \\
            3DGS+EWA
                                    & 33.60                                & 32.05                                & 28.28                                  & 25.11          & 29.76          & 0.970          & 0.972          & 0.963          & 0.946          & 0.962          & 0.039          & 0.027          & 0.033          & 0.045          & 0.036          \\
            3DGS+SS
                                    & 33.86                                & 31.47                                & 26.53                                  & 22.41          & 28.57          & \textbf{0.971} & 0.972          & 0.952          & 0.906          & 0.950          & \textbf{0.036} & 0.024          & 0.035          & 0.064          & 0.040          \\
            Mip-Splatting
                                    & 33.54                                & 34.09                                & 31.50                                  & 27.80          & 31.73          & 0.969          & 0.976          & 0.977          & 0.968          & 0.973          & 0.038          & 0.022          & 0.022          & 0.032          & 0.028          \\
            Analytic-Splatting
                                    & 33.78                                & 34.20                                & 31.16                                  & 27.22          & 31.59          & 0.970          & 0.977          & 0.977          & 0.965          & 0.972          & \textbf{0.036} & 0.022          & 0.025          & 0.037          & 0.030          \\
            \midrule
            Scaffold-GS+GB
                                    & 30.22                                & 32.02                                & 33.84                                  & 33.80          & 32.47          & 0.949          & 0.964          & 0.976          & 0.983          & 0.968          & 0.070          & 0.041          & 0.025          & 0.019          & 0.039          \\
            2DGS+GB
                                    & 31.96                                & 33.53                                & 34.93                                  & 34.74          & 33.79          & 0.960          & 0.970          & 0.979          & 0.985          & 0.974          & 0.055          & 0.033          & 0.022          & 0.015          & 0.031          \\
            Mip-Splatting+GB\textsubscript{test}
                                    & 33.45                                & 35.37                                & 36.98                                  & 36.36          & 35.54          & 0.969          & 0.978          & 0.984          & 0.988          & 0.980          & 0.038          & 0.021          & 0.014          & 0.012          & 0.021          \\
            Analytic-Splatting+GB\textsubscript{test}
                                    & 33.62                                & 35.72                                & \textbf{37.36}                         & \textbf{36.51} & \textbf{35.80} & 0.970          & \textbf{0.979} & \textbf{0.985} & \textbf{0.989} & \textbf{0.981} & 0.037          & \textbf{0.020} & \textbf{0.013} & \textbf{0.011} & \textbf{0.020} \\
            \midrule
            Gaussian Blending
                                    & \textbf{33.92}                       & \textbf{35.80}                       & 36.82                                  & 35.79          & 35.58          & 0.970          & \textbf{0.979} & \textbf{0.985} & 0.988          & \textbf{0.981} & \textbf{0.036} & \textbf{0.020} & \textbf{0.013} & 0.012          & \textbf{0.020} \\
            \bottomrule
        \end{tabular}
    }
\end{table*}

\begin{figure*}[ht!]
    \centering
    \setlength{\tabcolsep}{2pt}
    \begin{tabular}{ccccccc}
                                                                                                      & {Ground Truth} & {Tri-MipRF} & {3DGS} & {Mip-Splatting} & {Analytic-Splatting} & {Gaussian Blending} \\

        \raisebox{0.68em}{\rotatebox[origin=c]{90}{\textbf{1}}}                                        &
        \includegraphics[width=0.153\textwidth]{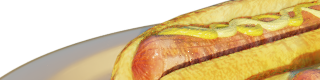}                 &
        \includegraphics[width=0.153\textwidth]{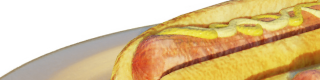}          &
        \includegraphics[width=0.153\textwidth]{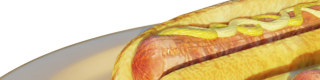}               &
        \includegraphics[width=0.153\textwidth]{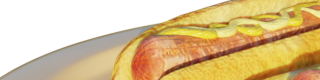}      &
        \includegraphics[width=0.153\textwidth]{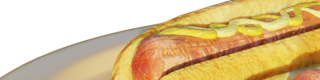} &
        \includegraphics[width=0.153\textwidth]{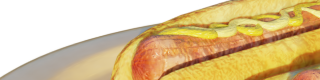}                                                                                                         \\ [-2.5pt]
        \raisebox{0.68em}{\rotatebox[origin=c]{90}{1/2}}                                               &
        \includegraphics[width=0.153\textwidth]{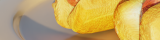}                 &
        \includegraphics[width=0.153\textwidth]{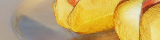}          &
        \includegraphics[width=0.153\textwidth]{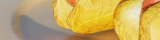}               &
        \includegraphics[width=0.153\textwidth]{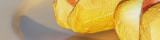}      &
        \includegraphics[width=0.153\textwidth]{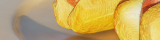} &
        \includegraphics[width=0.153\textwidth]{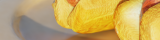}                                                                                                         \\ [-2.5pt]
        \raisebox{0.68em}{\rotatebox[origin=c]{90}{1/4}}                                               &
        \includegraphics[width=0.153\textwidth]{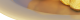}                 &
        \includegraphics[width=0.153\textwidth]{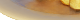}          &
        \includegraphics[width=0.153\textwidth]{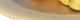}               &
        \includegraphics[width=0.153\textwidth]{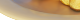}      &
        \includegraphics[width=0.153\textwidth]{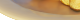} &
        \includegraphics[width=0.153\textwidth]{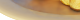}                                                                                                         \\ [-2.5pt]
        \raisebox{0.68em}{\rotatebox[origin=c]{90}{1/8}}                                               &
        \includegraphics[width=0.153\textwidth]{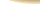}                 &
        \includegraphics[width=0.153\textwidth]{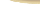}          &
        \includegraphics[width=0.153\textwidth]{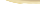}               &
        \includegraphics[width=0.153\textwidth]{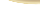}      &
        \includegraphics[width=0.153\textwidth]{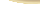} &
        \includegraphics[width=0.153\textwidth]{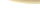}                                                                                                         \\

        \raisebox{0.68em}{\rotatebox[origin=c]{90}{\textbf{1}}}                                        &
        \includegraphics[width=0.153\textwidth]{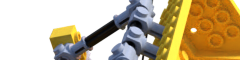}                 &
        \includegraphics[width=0.153\textwidth]{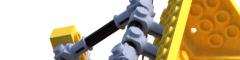}          &
        \includegraphics[width=0.153\textwidth]{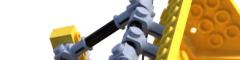}               &
        \includegraphics[width=0.153\textwidth]{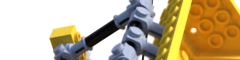}      &
        \includegraphics[width=0.153\textwidth]{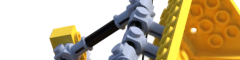} &
        \includegraphics[width=0.153\textwidth]{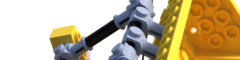}                                                                                                         \\ [-2.5pt]
        \raisebox{0.68em}{\rotatebox[origin=c]{90}{1/2}}                                               &
        \includegraphics[width=0.153\textwidth]{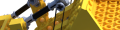}                 &
        \includegraphics[width=0.153\textwidth]{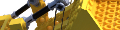}          &
        \includegraphics[width=0.153\textwidth]{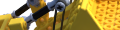}               &
        \includegraphics[width=0.153\textwidth]{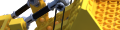}      &
        \includegraphics[width=0.153\textwidth]{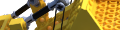} &
        \includegraphics[width=0.153\textwidth]{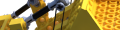}                                                                                                         \\ [-2.5pt]
        \raisebox{0.68em}{\rotatebox[origin=c]{90}{1/4}}                                               &
        \includegraphics[width=0.153\textwidth]{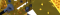}                 &
        \includegraphics[width=0.153\textwidth]{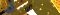}          &
        \includegraphics[width=0.153\textwidth]{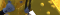}               &
        \includegraphics[width=0.153\textwidth]{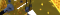}      &
        \includegraphics[width=0.153\textwidth]{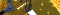} &
        \includegraphics[width=0.153\textwidth]{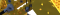}                                                                                                         \\ [-2.5pt]
        \raisebox{0.68em}{\rotatebox[origin=c]{90}{1/8}}                                               &
        \includegraphics[width=0.153\textwidth]{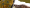}                 &
        \includegraphics[width=0.153\textwidth]{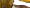}          &
        \includegraphics[width=0.153\textwidth]{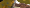}               &
        \includegraphics[width=0.153\textwidth]{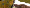}      &
        \includegraphics[width=0.153\textwidth]{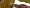} &
        \includegraphics[width=0.153\textwidth]{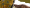}                                                                                                         \\

        \raisebox{0.68em}{\rotatebox[origin=c]{90}{\textbf{1}}}                                        &
        \includegraphics[width=0.153\textwidth]{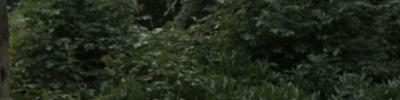}                 &
        \raisebox{1.18em}{\multirow{4}{*}{\includegraphics[width=0.153\textwidth]{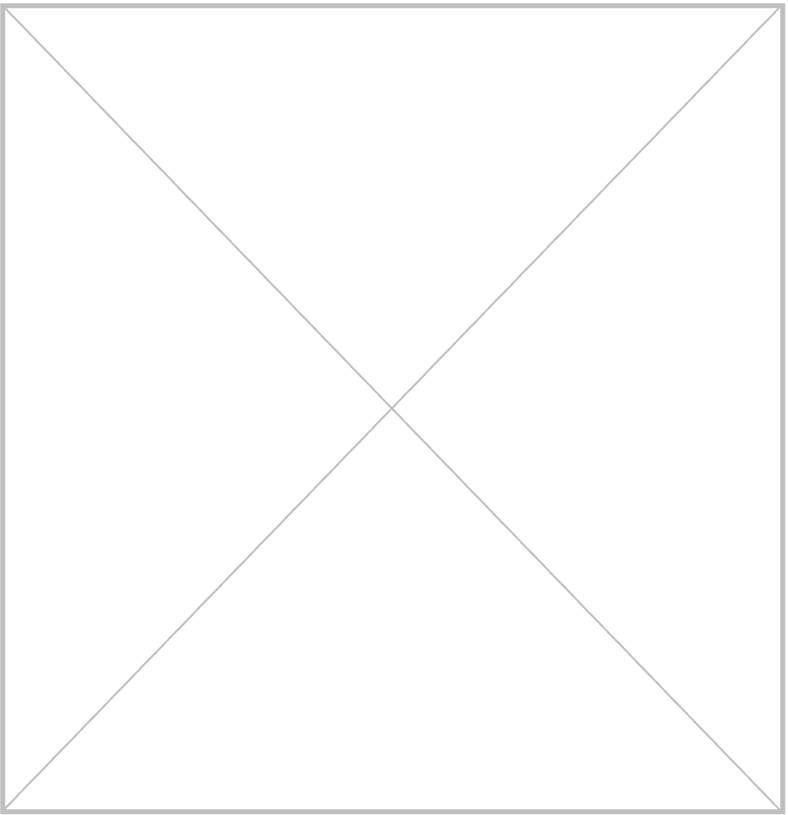}}}          &
        \includegraphics[width=0.153\textwidth]{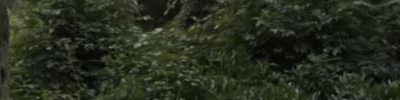}               &
        \includegraphics[width=0.153\textwidth]{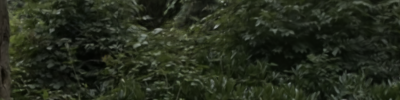}      &
        \includegraphics[width=0.153\textwidth]{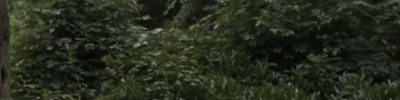} &
        \includegraphics[width=0.153\textwidth]{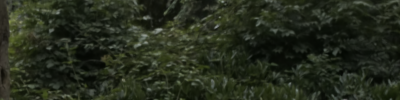}                                                                                                         \\ [-2.5pt]
        \raisebox{0.68em}{\rotatebox[origin=c]{90}{1/2}}                                               &
        \includegraphics[width=0.153\textwidth]{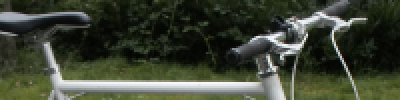}                 &
        &
        \includegraphics[width=0.153\textwidth]{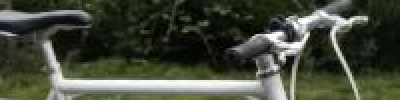}               &
        \includegraphics[width=0.153\textwidth]{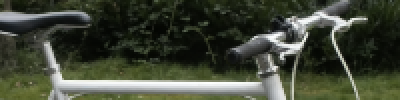}      &
        \includegraphics[width=0.153\textwidth]{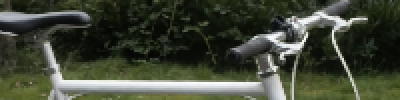} &
        \includegraphics[width=0.153\textwidth]{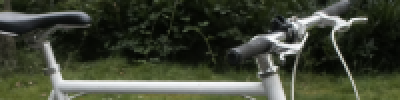}                                                                                                         \\ [-2.5pt]
        \raisebox{0.68em}{\rotatebox[origin=c]{90}{1/4}}                                               &
        \includegraphics[width=0.153\textwidth]{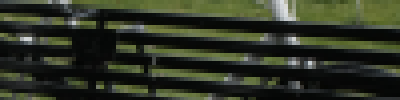}                 &
        &
        \includegraphics[width=0.153\textwidth]{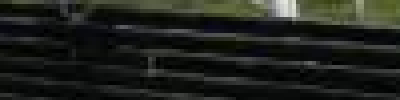}               &
        \includegraphics[width=0.153\textwidth]{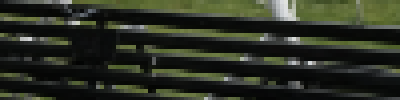}      &
        \includegraphics[width=0.153\textwidth]{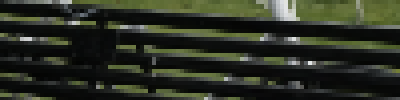} &
        \includegraphics[width=0.153\textwidth]{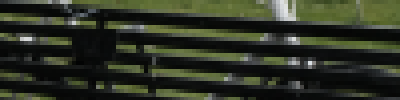}                                                                                                         \\ [-2.5pt]
        \raisebox{0.68em}{\rotatebox[origin=c]{90}{1/8}}                                               &
        \includegraphics[width=0.153\textwidth]{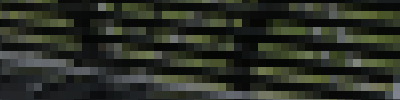}                 &
        &
        \includegraphics[width=0.153\textwidth]{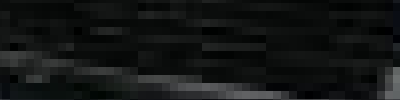}               &
        \includegraphics[width=0.153\textwidth]{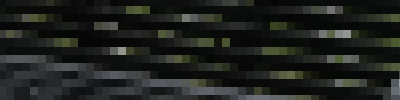}      &
        \includegraphics[width=0.153\textwidth]{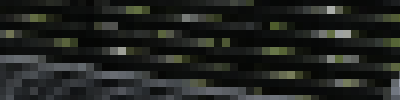} &
        \includegraphics[width=0.153\textwidth]{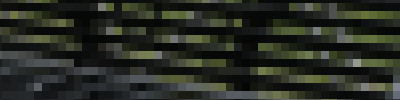}                                                                                                         \\
    \end{tabular}
    \caption{
        Qualitative results of zoom-out setting (Single-scale training ($\times 1$ resolution) and multi-scale testing).
        The training resolution is highlighted in \textbf{bold}.
        Our Gaussian Blending effectively suppresses pixel-level dilation in zoom-out scenarios (e.g., the disk edge in the top scene at $\times 1/8$ resolution).
    }
    \label{fig:qualitative_zoomout}
\end{figure*}

\textbf{Transmittance Computation}.
Next, we compute the optimal distribution $T_{i+1}$ to approximate the remaining transmittance, $T_i(x)(1-\alpha_i(x))$, after blending the current splat.
Specifically, we calculate the zeroth-, first-, and second-order moments ($M^0_i, M^1_i, M^2_i$) of the resulting distribution within the current window to preserve the total transmittance mass, mean, and variance after each blending step.
They are computed in the rotated coordinate system aligned with the splat’s principal axes by
\begin{align}
    M^0_i & = t_i l_{i, 1} l_{i, 2} - \int w_i(x)dx, \notag \\
    M^1_i & = t_i l_{i, 1} l_{i, 2} \begin{bmatrix}u\\v\end{bmatrix} - t_i o_i
    \begin{bmatrix}
        I^1_{\sigma_1}(u_1, u_2) \cdot I^0_{\sigma_2}(v_1, v_2) \\
        I^0_{\sigma_1}(u_1, u_2) \cdot I^1_{\sigma_2}(v_1, v_2)
    \end{bmatrix}, \notag \\
    M^2_i & = t_i l_{i, 1} l_{i, 2} \left(\begin{bmatrix}u^2\\v^2\end{bmatrix} + \frac{1}{12} l_i^2\right) \notag \\
             & \quad - t_i o_i
    \begin{bmatrix}
        I^2_{\sigma_1}(u_1, u_2) \cdot I^0_{\sigma_2}(v_1, v_2) \\
        I^0_{\sigma_1}(u_1, u_2) \cdot I^2_{\sigma_2}(v_1, v_2)
    \end{bmatrix}.
\end{align}

Using these moments, we obtain the parameters of the new 2D uniform transmittance distribution $T_{i+1}$ as
\begin{align}
    x_{i+1} & = \mu_i' + [e_1, e_2] M^1_i / M^0_i, \notag \\
    l_{i+1} & = \sqrt{12\left(M^2_i / M^0_i - \left(M^1_i / M^0_i\right)^2\right)}, \notag \\
    t_{i+1} & = M^0_i / l_{i+1,1} l_{i+1,2}.
\end{align}

Through this process, starting from an initially fully visible pixel area, we iteratively integrate the response of each splat and dynamically update the rendering window as in Figure~\ref{fig:rendering}(c).
This dynamic update allows us to handle spatial occlusions by adaptively shrinking the rendering window in regions of lower transmittance, reducing the influence of overlapping splats.
Consequently, each splat is blended according to its spatial distribution, effectively and efficiently reducing intra-pixel aliasing within a single rendering pass. 

Finally, we compute the final pixel color as
\begin{equation}
    C_p=\sum_{i=1}^{N} c_i\int w_i(x)dx.
\end{equation}

\begin{table*}[ht!]
    \centering
    \caption{Single-scale training ($\times 1/8$ resolution) and multi-scale testing results on the multi-scale Blender dataset. The best results are highlighted in \textbf{bold}.}
    \label{tab:stmt8_blender}
    \resizebox{\textwidth}{!}{
        \begin{tabular}{l|ccccc|ccccc|ccccc}
            \toprule
            \multirow{2}{*}{} & \multicolumn{5}{c|}{PSNR $\uparrow$} & \multicolumn{5}{c|}{SSIM $\uparrow$} & \multicolumn{5}{c}{LPIPS $\downarrow$}                                                                                                                                                                                                             \\
                                    & $\times 1$                           & $\times 1/2$                         & $\times 1/4$                           & $\times 1/8$   & Avg.
                                    & $\times 1$                           & $\times 1/2$                         & $\times 1/4$                           & $\times 1/8$   & Avg.
                                    & $\times 1$                           & $\times 1/2$                         & $\times 1/4$                           & $\times 1/8$   & Avg.                                                                                                                                                                                     \\
            \midrule
            TensoRF
                                    & 25.42                                & 26.57                                & 29.02                                  & 32.87          & 28.47          & 0.874          & 0.902          & 0.946          & 0.977          & 0.925          & 0.159          & 0.132          & 0.078          & 0.028          & 0.099          \\
            3DGS
                                    & 18.51                                & 19.90                                & 23.33                                  & 34.72          & 24.11          & 0.824          & 0.832          & 0.905          & 0.984          & 0.886          & 0.163          & 0.136          & 0.072          & 0.016          & 0.097          \\
            Scaffold-GS
                                    & 16.37                                & 17.71                                & 21.35                                  & 34.06          & 22.37          & 0.802          & 0.797          & 0.867          & 0.980          & 0.862          & 0.187          & 0.172          & 0.106          & 0.020          & 0.121          \\
            2DGS
                                    & 22.87                                & 23.59                                & 25.46                                  & 31.85          & 25.94          & 0.864          & 0.879          & 0.920          & 0.973          & 0.909          & 0.159          & 0.139          & 0.087          & 0.031          & 0.104          \\
            3D-HGS
                                    & 18.87                                & 20.16                                & 23.39                                  & 35.08          & 24.38          & 0.826          & 0.834          & 0.904          & 0.984          & 0.887          & 0.169          & 0.142          & 0.073          & 0.015          & 0.100          \\
            \midrule
            Tri-MipRF
                                    & 21.99                                & 22.91                                & 25.84                                  & 33.93          & 26.17          & 0.809          & 0.821          & 0.888          & 0.979          & 0.874          & 0.240          & 0.232          & 0.170          & 0.023          & 0.166          \\
            3DGS+EWA
                                    & 21.76                                & 23.15                                & 26.64                                  & 34.32          & 26.47          & 0.834          & 0.871          & 0.938          & 0.983          & 0.907          & 0.231          & 0.184          & 0.095          & 0.019          & 0.132          \\
            3DGS+SS
                                    & 20.98                                & 22.47                                & 26.33                                  & \textbf{36.96} & 26.68          & 0.851          & 0.877          & 0.942          & \textbf{0.987} & 0.914          & 0.136          & 0.100          & 0.048          & \textbf{0.012} & 0.074          \\
            Mip-Splatting
                                    & 25.97                                & 27.24                                & 30.06                                  & 35.29          & 29.64          & \textbf{0.892} & 0.920          & 0.959          & 0.985          & 0.939          & 0.150          & 0.117          & 0.056          & 0.015          & 0.084          \\
            Analytic-Splatting
                                    & 25.45                                & 26.98                                & 30.04                                  & 35.63          & 29.53          & 0.878          & 0.914          & 0.956          & 0.984          & 0.933          & 0.138          & 0.102          & 0.051          & 0.015          & 0.077          \\
            \midrule
            Scaffold-GS+GB
                                    & 26.57                                & 28.02                                & 31.11                                  & 35.92          & 30.41          & 0.886          & 0.918          & 0.958          & 0.985          & 0.937          & 0.149          & 0.116          & 0.061          & 0.016          & 0.085          \\
            2DGS+GB
                                    & 25.49                                & 26.94                                & 29.47                                  & 33.05          & 28.74          & 0.880          & 0.908          & 0.947          & 0.976          & 0.928          & 0.151          & 0.120          & 0.067          & 0.025          & 0.091          \\
            Mip-Splatting+GB\textsubscript{test}
                                    & 25.95                                & 27.17                                & 29.67                                  & 33.69          & 29.12          & 0.891          & 0.919          & 0.958          & 0.984          & 0.938          & 0.151          & 0.120          & 0.059          & 0.016          & 0.087          \\
            Analytic-Splatting+GB\textsubscript{test}
                                    & 25.37                                & 26.87                                & 29.51                                  & 33.34          & 28.78          & 0.876          & 0.912          & 0.954          & 0.982          & 0.931          & 0.139          & 0.104          & 0.053          & 0.017          & 0.078          \\
            \midrule
            Gaussian Blending
                                    & \textbf{26.74}                       & \textbf{28.53}                       & \textbf{31.97}                         & 36.92          & \textbf{31.04} & 0.891          & \textbf{0.927} & \textbf{0.965} & \textbf{0.987} & \textbf{0.943} & \textbf{0.132} & \textbf{0.092} & \textbf{0.042} & \textbf{0.012} & \textbf{0.069} \\
            \bottomrule
        \end{tabular}
    }
\end{table*}

\begin{figure*}[ht!]
    \centering
    \setlength{\tabcolsep}{2pt}
    \begin{tabular}{ccccccc}
                                                                                                      & {Ground Truth} & {Tri-MipRF} & {3DGS} & {Mip-Splatting} & {Analytic-Splatting} & {Gaussian Blending} \\

        \raisebox{0.68em}{\rotatebox[origin=c]{90}{1}}                                                 &
        \includegraphics[width=0.153\textwidth]{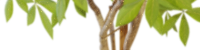}                  &
        \includegraphics[width=0.153\textwidth]{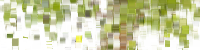}           &
        \includegraphics[width=0.153\textwidth]{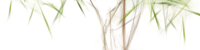}                &
        \includegraphics[width=0.153\textwidth]{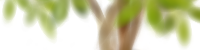}       &
        \includegraphics[width=0.153\textwidth]{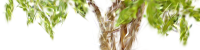}  &
        \includegraphics[width=0.153\textwidth]{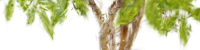}                                                                                                          \\ [-2.5pt]
        \raisebox{0.68em}{\rotatebox[origin=c]{90}{1/2}}                                               &
        \includegraphics[width=0.153\textwidth]{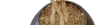}                  &
        \includegraphics[width=0.153\textwidth]{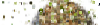}           &
        \includegraphics[width=0.153\textwidth]{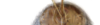}                &
        \includegraphics[width=0.153\textwidth]{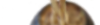}       &
        \includegraphics[width=0.153\textwidth]{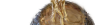}  &
        \includegraphics[width=0.153\textwidth]{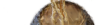}                                                                                                          \\ [-2.5pt]
        \raisebox{0.68em}{\rotatebox[origin=c]{90}{1/4}}                                               &
        \includegraphics[width=0.153\textwidth]{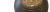}                  &
        \includegraphics[width=0.153\textwidth]{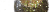}           &
        \includegraphics[width=0.153\textwidth]{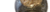}                &
        \includegraphics[width=0.153\textwidth]{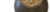}       &
        \includegraphics[width=0.153\textwidth]{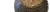}  &
        \includegraphics[width=0.153\textwidth]{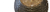}                                                                                                          \\ [-2.5pt]
        \raisebox{0.68em}{\rotatebox[origin=c]{90}{\textbf{1/8}}}                                      &
        \includegraphics[width=0.153\textwidth]{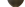}                  &
        \includegraphics[width=0.153\textwidth]{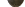}           &
        \includegraphics[width=0.153\textwidth]{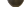}                &
        \includegraphics[width=0.153\textwidth]{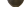}       &
        \includegraphics[width=0.153\textwidth]{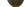}  &
        \includegraphics[width=0.153\textwidth]{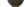}                                                                                                          \\

        \raisebox{0.68em}{\rotatebox[origin=c]{90}{1}}                                                 &
        \includegraphics[width=0.153\textwidth]{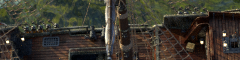}                  &
        \includegraphics[width=0.153\textwidth]{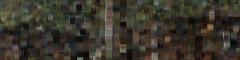}           &
        \includegraphics[width=0.153\textwidth]{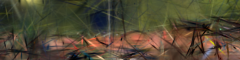}                &
        \includegraphics[width=0.153\textwidth]{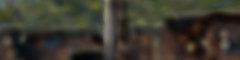}       &
        \includegraphics[width=0.153\textwidth]{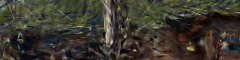}  &
        \includegraphics[width=0.153\textwidth]{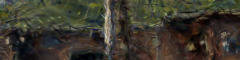}                                                                                                          \\ [-2.5pt]
        \raisebox{0.68em}{\rotatebox[origin=c]{90}{1/2}}                                               &
        \includegraphics[width=0.153\textwidth]{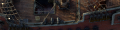}                  &
        \includegraphics[width=0.153\textwidth]{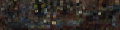}           &
        \includegraphics[width=0.153\textwidth]{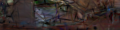}                &
        \includegraphics[width=0.153\textwidth]{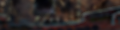}       &
        \includegraphics[width=0.153\textwidth]{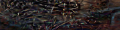}  &
        \includegraphics[width=0.153\textwidth]{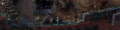}                                                                                                          \\ [-2.5pt]
        \raisebox{0.68em}{\rotatebox[origin=c]{90}{1/4}}                                               &
        \includegraphics[width=0.153\textwidth]{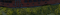}                  &
        \includegraphics[width=0.153\textwidth]{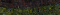}           &
        \includegraphics[width=0.153\textwidth]{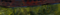}                &
        \includegraphics[width=0.153\textwidth]{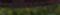}       &
        \includegraphics[width=0.153\textwidth]{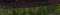}  &
        \includegraphics[width=0.153\textwidth]{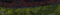}                                                                                                          \\ [-2.5pt]
        \raisebox{0.68em}{\rotatebox[origin=c]{90}{\textbf{1/8}}}                                      &
        \includegraphics[width=0.153\textwidth]{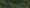}                  &
        \includegraphics[width=0.153\textwidth]{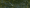}           &
        \includegraphics[width=0.153\textwidth]{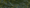}                &
        \includegraphics[width=0.153\textwidth]{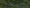}       &
        \includegraphics[width=0.153\textwidth]{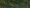}  &
        \includegraphics[width=0.153\textwidth]{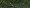}                                                                                                          \\

        \raisebox{0.68em}{\rotatebox[origin=c]{90}{1}}                                                 &
        \includegraphics[width=0.153\textwidth]{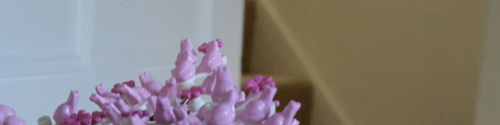}                 &
        \raisebox{1.18em}{\multirow{4}{*}{\includegraphics[width=0.153\textwidth]{png_figures/qualitative_mipnerf360/empty.png}}}          &
        \includegraphics[width=0.153\textwidth]{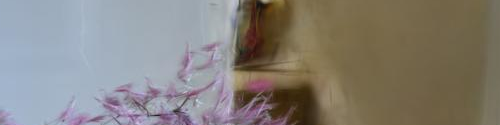}               &
        \includegraphics[width=0.153\textwidth]{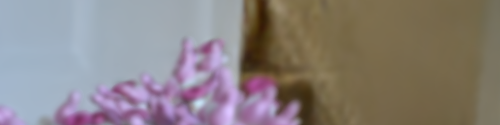}      &
        \includegraphics[width=0.153\textwidth]{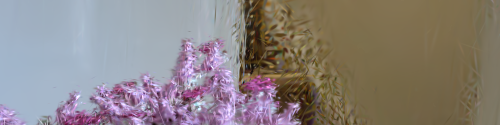} &
        \includegraphics[width=0.153\textwidth]{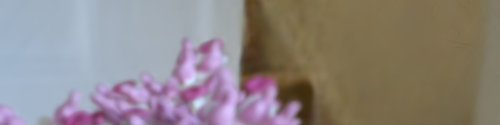}                                                                                                         \\ [-2.5pt]
        \raisebox{0.68em}{\rotatebox[origin=c]{90}{1/2}}                                               &
        \includegraphics[width=0.153\textwidth]{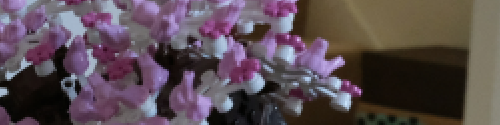}                 &
        &
        \includegraphics[width=0.153\textwidth]{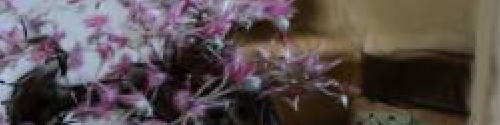}               &
        \includegraphics[width=0.153\textwidth]{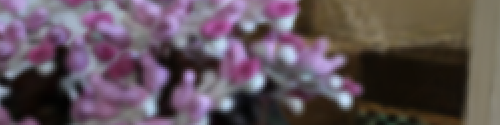}      &
        \includegraphics[width=0.153\textwidth]{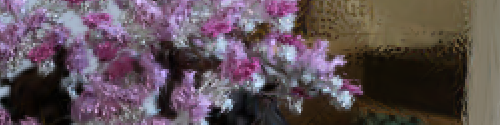} &
        \includegraphics[width=0.153\textwidth]{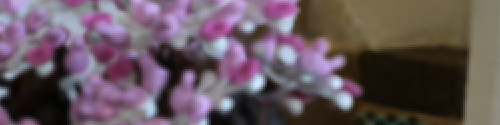}                                                                                                         \\ [-2.5pt]
        \raisebox{0.68em}{\rotatebox[origin=c]{90}{1/4}}                                               &
        \includegraphics[width=0.153\textwidth]{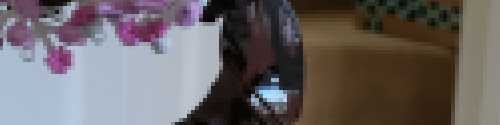}                 &
        &
        \includegraphics[width=0.153\textwidth]{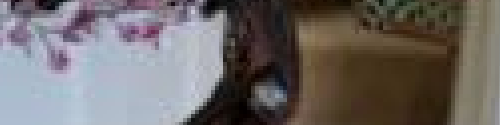}               &
        \includegraphics[width=0.153\textwidth]{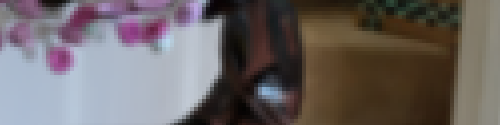}      &
        \includegraphics[width=0.153\textwidth]{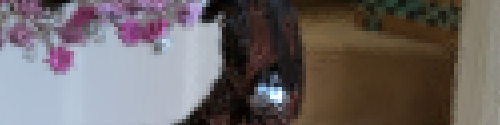} &
        \includegraphics[width=0.153\textwidth]{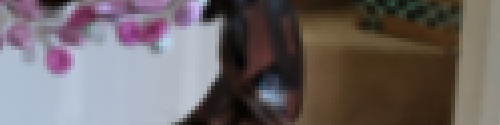}                                                                                                         \\ [-2.5pt]
        \raisebox{0.68em}{\rotatebox[origin=c]{90}{\textbf{1/8}}}                                      &
        \includegraphics[width=0.153\textwidth]{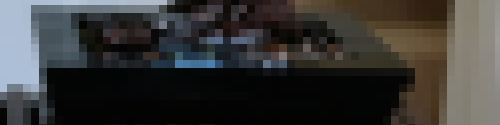}                 &
        &
        \includegraphics[width=0.153\textwidth]{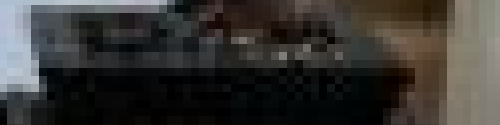}               &
        \includegraphics[width=0.153\textwidth]{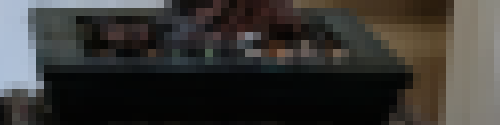}      &
        \includegraphics[width=0.153\textwidth]{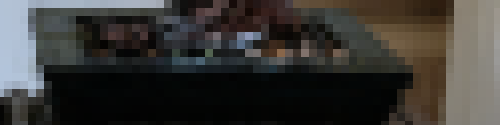} &
        \includegraphics[width=0.153\textwidth]{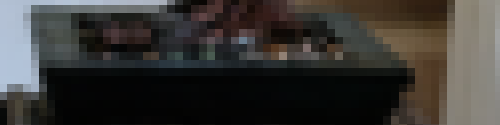}                                                                                                         \\
    \end{tabular}
    \caption{
        Qualitative results of zoom-in setting (Single-scale training ($\times 1/8$ resolution) and multi-scale testing).
        The training resolution is highlighted in \textbf{bold}.
        Our Gaussian Blending effectively prevents erosion in zoom-in scenarios (e.g., the leaves in the top scene at $\times 1$ resolution).
    }
    \label{fig:qualitative_zoomin}
\end{figure*}

\section{Experiments}

\subsection{Experiment Setup}
\label{sec:experiment_setup}

\begin{table*}[ht!]
    \centering
    \caption{PSNR scores for multi-scale testing on the multi-scale Mip-NeRF 360 dataset. The best results are highlighted in \textbf{bold}.}
    \label{tab:mipnerf360}
    \resizebox{\textwidth}{!}{
        \begin{tabular}{l|ccccc|ccccc|ccccc}
            \toprule
            \multirow{2}{*}{} & \multicolumn{5}{c|}{(a) Single-scale Training ($\times1$ res)} & \multicolumn{5}{c|}{(b) Single-scale Training ($\times1/8$ res)} & \multicolumn{5}{c}{(c) Multi-scale Training} \\
                                    & $\times 1$                           & $\times 1/2$                         & $\times 1/4$                           & $\times 1/8$   & Avg.
                                    & $\times 1$                           & $\times 1/2$                         & $\times 1/4$                           & $\times 1/8$   & Avg.
                                    & $\times 1$                           & $\times 1/2$                         & $\times 1/4$                           & $\times 1/8$   & Avg. \\
            \midrule
            3DGS
                                    & 27.63                                & 25.76                                & 21.98                                  & 19.34          & 23.68          & 17.25                                & 18.73                                & 22.41                                  & 30.60          & 22.25          & 26.85                                & 27.89                                & 28.28                                  & 27.17          & 27.55 \\
            Scaffold-GS
                                    & 27.27                                & 26.05                                & 22.48                                  & 19.70          & 23.87          & 17.09                                & 18.43                                & 21.98                                  & 31.06          & 22.14          & 26.61                                & 27.75                                & 28.42                                  & 27.23          & 27.50 \\
            2DGS
                                    & 26.71                                & 25.99                                & 21.84                                  & 18.41          & 23.24          & 20.92                                & 21.66                                & 23.85                                  & 29.49          & 23.98          & 25.70                                & 26.73                                & 27.72                                  & 26.52          & 26.67 \\
            3D-HGS
                                    & \textbf{28.07}                       & 26.12                                & 22.13                                  & 19.36          & 23.92          & 18.16                                & 19.59                                & 23.10                                  & \textbf{31.26} & 23.03          & 27.47                                & 28.62                                & 29.40                                  & 28.75          & 28.56 \\
            \midrule
            3DGS+EWA
                                    & 27.67                                & 28.40                                & 28.23                                  & 27.19          & 27.87          & 20.27                                & 21.76                                & 24.90                                  & 29.38          & 24.08          & 26.50                                & 27.75                                & 28.93                                  & 29.25          & 28.11 \\
            3DGS+SS
                                    & 27.67                                & 27.71                                & 25.47                                  & 22.59          & 25.86          & 20.14                                & 21.50                                & 24.84                                  & 31.22          & 24.42          & 27.37                                & 28.54                                & 29.69                                  & 29.44          & 28.76 \\
            Mip-Splatting
                                    & 27.50                                & 28.27                                & 29.22                                  & 28.89          & 28.47          & 24.35                                & 25.51                                & 27.79                                  & 30.95          & 27.15          & 27.41                                & 28.46                                & 30.01                                  & 31.13          & 29.25 \\
            Analytic-Splatting
                                    & 27.36                                & 28.12                                & 28.92                                  & 28.51          & 28.23          & 23.00                                & 24.37                                & 27.12                                  & 30.86          & 26.34          & 27.31                                & 28.40                                & 29.98                                  & 31.17          & 29.21 \\
            \midrule
            Gaussian Blending
                                    & 27.55                                & \textbf{28.62}                       & \textbf{29.92}                         & \textbf{30.58} & \textbf{29.17} & \textbf{24.86}                       & \textbf{25.97}                       & \textbf{28.13}                         & 30.90          & \textbf{27.47} & \textbf{27.58}                       & \textbf{28.77}                       & \textbf{30.24}                         & \textbf{31.54} & \textbf{29.53} \\
            \bottomrule
        \end{tabular}
    }
\end{table*}

\textbf{Datasets}.
We evaluate novel view synthesis performance on two standard datasets: multi-scale Blender \cite{nerf,mip_nerf} and multi-scale Mip-NeRF 360 dataset \cite{mip_nerf_360}.
The Blender dataset comprises eight synthetic scenes, each containing 100 training and 200 test images, with a resolution of $800\times800$ pixels.
The Mip-NeRF 360 dataset includes nine real-world scenes (five outdoor and four indoor scenes).
Following standard practice, we use every eighth image for testing, and the other images for training.
For both datasets, to assess rendering quality at varying sampling rates, we also include downsampled versions of each image by factors of 2, 4, and 8.

\textbf{Performance Measures}.
We report results using three standard metrics:
Peak Signal-to-Noise Ratio (PSNR), Structural Similarity Index Measure (SSIM), and Learned Perceptual Image Patch Similarity (LPIPS).

\textbf{Implementation}.
Our Gaussian Blending algorithm is implemented in CUDA. 
Through careful optimization, it achieves identical time complexity and comparable runtime performance to 3DGS models without additional memory overhead.
Following prior works, we train all models for 30k iterations using identical hyperparameters across all scenes.

\textbf{Baselines}.
We compare the performance against state-of-the-art real-time NVS and anti-aliased NVS methods, including (i) five standard NVS approaches: TensoRF \cite{tensorf}, 3DGS \cite{3dgs}, Scaffold-GS \cite{scaffold_gs}, 2DGS \cite{2dgs}, and 3D-HGS \cite{3d_hgs}, (ii) one mipmap-based anti-aliasing method: Tri-MipRF \cite{tri_miprf}, and (iii) three prefiltering-based anti-aliasing methods: 3DGS+EWA \cite{ewa}, Mip-Splatting \cite{mip_splatting}, and Analytic-Splatting \cite{analytic_splatting}.
We also include a supersampling baseline, 3DGS+SS, which renders images at the $2\times$ target resolution and then downsamples it by half.
Note that TensoRF and Tri-MipRF are tested only in the Blender dataset, since they are not applicable to unbounded scenes.

Gaussian Blending is a general rendering formulation that can function both as a standalone model and as a drop-in renderer for existing NVS frameworks.
We demonstrate its flexibility across three representative integration settings:
(i) for anti-aliased NVS methods such as Mip-Splatting and Analytic-Splatting, we retain each model’s original training pipeline and simply substitute their rendering module with our kernel at test time, denoted as \textit{Mip-Splatting+GB\textsubscript{test}} and \textit{Analytic-Splatting+GB\textsubscript{test}};
(ii) for general 3DGS-based frameworks such as Scaffold-GS, we replace the original CUDA kernel with our Gaussian Blending kernel during both training and inference, denoted as \textit{Scaffold-GS+GB}; and
(iii) for models using non-3DGS rendering backbones, such as 2DGS, we reimplement our spatial alpha blending concept within their native kernel, yielding \textit{2DGS+GB}.

\subsection{Approximation Error Analysis}
As shown in Figure~\ref{fig:error_analysis}, scalar alpha blending methods—such as 3DGS, 3DGS+SS, Mip-Splatting, and Analytic-Splatting—exhibit noticeable transmittance errors.
In particular, their transmittance error $\Delta T$ become negative, indicating dilation, where the occluded splat contributes excessively to the pixel color.
Although Analytic-Splatting analytically integrates each splat’s response over the pixel area, it still relies on scalar alpha blending when compositing multiple splats.
As a result, the first splat’s contribution is computed accurately, but the second splat, which spatially overlaps with the first splat, is blended using an averaged transmittance value.
This averaging neglects spatial occlusion, causing the second splat to receive an erroneously higher weight and leading to over-attenuated (dilated) transmittance after rendering.
In contrast, our Gaussian Blending dynamically adjusts the transmittance window to explicitly account for spatial occlusion.
As a result, Gaussian Blending achieves on average more than $5\times$ lower transmittance error compared to previous methods.

\begin{table}[t]
    \centering
    \caption{Multi-scale training and multi-scale testing results on the multi-scale Blender dataset.}
    \label{tab:mtmt_blender}
    \resizebox{0.45\textwidth}{!}{
        \begin{tabular}{l|ccccc}
            \toprule
            \multirow{2}{*}{} & \multicolumn{5}{c}{PSNR $\uparrow$}                                                                              \\
                                    & $\times 1$                           & $\times 1/2$   & $\times 1/4$   & $\times 1/8$   & Avg. \\
            \midrule
            TensoRF
                                    & 33.03                                & 33.92                                & 30.99                                  & 27.26          & 31.30 \\
            3DGS
                                    & 30.19                                & 31.38                                & 30.59                                  & 27.05          & 29.90 \\
            Scaffold-GS
                                    & 28.56                                & 29.85                                & 29.41                                  & 26.08          & 28.48 \\
            2DGS
                                    & 27.95                                & 29.39                                & 30.31                                  & 26.23          & 28.47 \\
            3D-HGS
                                    & 30.96                                & 32.40                                & 31.90                                  & 28.70          & 30.99 \\
            \midrule
            Tri-MipRF
                                    & 32.60                                & 34.18                                & 34.97                                  & 35.32          & 34.27 \\
            3DGS+EWA
                                    & 30.34                                & 31.78                                & 32.45                                  & 31.50          & 31.52 \\
            3DGS+SS
                                    & 32.12                                & 33.81                                & 33.98                                  & 31.13          & 32.76 \\
            Mip-Splatting
                                    & 32.93                                & 34.68                                & 35.79                                  & 35.48          & 34.72 \\
            Analytic-Splatting
                                    & 33.28                                & 35.02                                & 36.07                                  & 35.90          & 35.07 \\
            \midrule
            Scaffold-GS+GB
                                    & 29.68                                & 31.56                                & 33.81                                  & 34.87          & 32.48 \\
            2DGS+GB
                                    & 31.12                                & 32.89                                & 34.99                                  & 36.26          & 33.81 \\
            Mip-Splatting+GB\textsubscript{test}
                                    & 32.85                                & 34.69                                & 36.32                                  & 37.00          & 35.22 \\
            Analytic-Splatting+GB\textsubscript{test}
                                    & 33.15                                & 35.09                                & 36.52                                  & 36.44          & 35.30 \\
            \midrule
            Gaussian Blending
                                    & \textbf{33.50}                       & \textbf{35.48}                       & \textbf{37.38}                         & \textbf{38.39} & \textbf{36.19} \\
            \bottomrule
        \end{tabular}
    }
\end{table}

\subsection{Single-Scale Training and Multi-Scale Testing}
To evaluate anti-aliasing performance at sampling rates not encountered during training, we experiment with a single-scale training and multi-scale testing setup.
Specifically, models are trained at each of the resolutions ($\times1$, $\times1/2$, $\times1/4$, or $\times1/8$), and their rendering quality is tested across all resolutions.
Quantitative results demonstrate that Gaussian Blending consistently outperforms competing methods in both zoom-out (Table \ref{tab:stmt1_blender} and \ref{tab:mipnerf360}(a)) and zoom-in (Table \ref{tab:stmt8_blender} and \ref{tab:mipnerf360}(b)) scenarios, achieving particularly impressive gains in zoom-out settings.
Moreover, our qualitative results in Figure \ref{fig:qualitative_zoomout} and \ref{fig:qualitative_zoomin} further illustrate that Gaussian Blending successfully mitigates aliasing artifacts at sampling rates unseen during training.

Models employing scalar alpha blending suffer from pixel-level dilation and erosion artifacts.
Mipmap-based models struggle with rendering at unseen sampling rates due to extrapolation issues, thus creating floaters from the unseen resolution mipmap.
Mip-Splatting avoids sparsity issues during zoom-in due to its 3D smoothing filter, but it produces blurry renderings and fails to accurately reconstruct high-frequency details.
In contrast, our model prevents saturation of the transmittance by avoiding rendering in a low transmittance area.
As shown in Table~\ref{tab:efficiency}, Gaussian Blending achieves real-time rendering without any additional memory overhead and converges within only 7K iterations—over $3\times$ faster than the baselines—while already surpassing them in anti-aliasing performance.

\begin{table}[t]
    \centering
    \caption{Efficiency comparison across different NVS methods. We report average zoom-out quality (PSNR), $\times 1$ resolution rendering speed (FPS), training time, and storage memory usage in the multi-scale Blender dataset.}
    \label{tab:efficiency}
    \resizebox{0.47\textwidth}{!}{
        \begin{tabular}{l|c|ccc}
            \toprule
            Method                  & PSNR $\uparrow$ & FPS $\uparrow$ & Training Time $\downarrow$ & Memory (MB) $\downarrow$ \\
            \midrule
            TensoRF
                                    & 30.87 & 0.83 & 18m 18s & 72.59 \\
            3DGS
                                    & 24.95 & 131.80 & 16m 2s & 61.86 \\
            Scaffold-GS
                                    & 24.92 & 134.68 & 12m 2s & 8.14 \\
            2DGS
                                    & 23.79 & 95.35 & 19m 12s & 25.97 \\
            3D-HGS
                                    & 25.19 & 184.33 & 10m 43s & 61.52 \\
            \midrule
            Tri-MipRF
                                    & 29.58 & 3.18 & 6m 33s & 58.31 \\
            3DGS+EWA
                                    & 29.76 & 104.71 & 13m 18s & 56.19 \\
            3DGS+SS
                                    & 28.57 & 79.21 & 21m 56s & 60.84 \\
            Mip-Splatting
                                    & 31.73 & 131.58 & 10m 10s & 71.28 \\
            Analytic-Splatting
                                    & 31.59 & 72.07 & 11m 37s & 69.62 \\
            \midrule
            Gaussian Blending-7K
                                    & 34.02 & 128.49 & 2m 3s & 55.17 \\
            Gaussian Blending-15K
                                    & 35.24 & 112.45 & 6m 6s & 72.82 \\
            Gaussian Blending-30K
                                    & \textbf{35.58} & 123.08 & 12m 34s & 72.82 \\
            \bottomrule
        \end{tabular}
    }
\end{table}

\subsection{Multi-Scale Training and Multi-Scale Testing}
To further demonstrate the robustness of Gaussian Blending in handling varying sampling rates, we also experiment with multi-scale training and multi-scale testing.
In this scenario, models are simultaneously trained on images across all four resolutions: $\times1$, $\times1/2$, $\times1/4$, and $\times1/8$, and then evaluated on the same scales to assess whether models can effectively learn from multi-scale data.

Table \ref{tab:mtmt_blender} and Table \ref{tab:mipnerf360}(c) show that our model outperforms all other real-time and unbounded NVS methods.
Our model consistently captures spatial variations without introducing dilation artifacts across diverse sampling rates, even when trained on complex scenes spanning a wide range of sampling rates.

\subsection{Drop-in Gaussian Blending}
To further demonstrate the generality of our formulation, we apply Gaussian Blending to various NVS frameworks with different rendering backbones, following the three integration types described in Section~\ref{sec:experiment_setup}.
Across all integration settings, Gaussian Blending consistently enhances rendering quality and anti-aliasing performance.
Remarkably, even when the splat size and spatial distribution are learned according to each model’s native renderer—such as in Mip-Splatting and Analytic-Splatting—simply replacing their rendering kernel with ours \textit{at test time only} prevents pixel-level dilation.
This demonstrates that Gaussian Blending offers a more physically consistent integration of alpha and transmittance, effectively reducing pixel-level dilation and preserving high-frequency details, all without retraining or modifying the underlying representation.

\begin{figure}[t!]
    \centering
    \renewcommand{\arraystretch}{0.0}
    \setlength{\tabcolsep}{2pt}
    \begin{tabular}{@{}ccc@{\hskip 1pt}c@{}}
        & 2DGS & Scaffold-GS &
        \raisebox{-3em}[0pt][0pt]{\multirow[c]{3}{*}{
            \shortstack[c]{
                Ground Truth\\
                \includegraphics[width=0.14\textwidth]{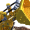}
            }
        }} \\ [3pt]
        \raisebox{3.4em}{\rotatebox[origin=c]{90}{Original}}
        & \includegraphics[width=0.14\textwidth]{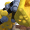}
        & \includegraphics[width=0.14\textwidth]{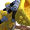}
        & \\ [1pt]
        \raisebox{3.4em}{\rotatebox[origin=c]{90}{+ GB}}
        & \includegraphics[width=0.14\textwidth]{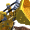}
        & \includegraphics[width=0.14\textwidth]{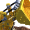}
        & \\
    \end{tabular}
    \caption{Qualitative comparison of Gaussian Blending applied to 2DGS and Scaffold-GS.} 
    \label{fig:plug_and_play}
\end{figure}

Moreover, Gaussian Blending exhibits strong robustness to unseen sampling rates, including both zoom-in and zoom-out scenarios.
While conventional scalar alpha blending tends to blur boundaries under high sampling rates and introduce staircase-like dilation under low rates, our method maintains stable transmittance accumulation and consistent image quality across scales.
When the model is trained directly using our kernel, the improvements become even more pronounced across all resolutions, confirming that Gaussian Blending functions as a unified rendering formulation suitable for both training and inference.
Qualitative comparisons in Figure~\ref{fig:plug_and_play} further support these findings: both \textit{Scaffold-GS+GB} and \textit{2DGS+GB} deliver sharper object boundaries compared to their original rendering kernels.
Our spatial alpha blending concept generalizes beyond 3DGS and can be applied to any neural rendering framework that employs an integrable alpha or density function.

\section{Conclusion}

We introduced \textit{Gaussian Blending}, a novel spatial alpha blending method to address intra-pixel aliasing artifacts arising from traditional scalar alpha blending used in existing novel view synthesis (NVS) methods.
By representing alpha and transmittance as spatially varying distributions within pixel regions, Gaussian Blending effectively mitigates erosion and dilation artifacts when rendering views at sampling rates unseen during training.
Leveraging the spatial coherence of Gaussian splats, our approach maintains identical computational complexity and memory usage to standard 3DGS.
Extensive experiments demonstrate that Gaussian Blending consistently synthesizes higher-quality views across both single-scale and multi-scale scenarios without additional priors or retraining.
Furthermore, it can be seamlessly integrated as a drop-in replacement for existing NVS frameworks, providing an efficient and practical anti-aliasing solution.

While Gaussian Blending performs well even on complex real scenes, it cannot resolve boundary blurring caused by real camera effects such as diffraction, chromatic aberration, and defocus blur.
Additionally, since the zoom-in setting is an ill-posed problem, high-frequency noise can occur when highly specular objects are present (e.g., the \textit{drums} scene).
Addressing these issues to further enhance performance in complex real-world scenarios remains future work.

\section*{Acknowledgments}
This work was supported by Institute of Information \& communications Technology Planning \& Evaluation (IITP) grant funded by the Korea government (MSIT) (No.~RS-2022-II220156, Fundamental research on continual meta-learning for quality enhancement of casual videos and their 3D metaverse transformation), Institute of Information \& Communications Technology Planning \& Evaluation(IITP) grant funded by the Korea government(MSIT) (RS-2025-25442338, AI star Fellowship Support Program(Seoul National Univ.)), Basic Science Research Program through the National Research Foundation of Korea(NRF) funded by the Ministry of Education(RS-2023-00274280), IITP(Institute of Information \& Communications Technology Planning \& Evaluation)-ITRC(Information Technology Research Center) grant funded by the Korea government(Ministry of Science and ICT)(IITP-2025-RS-2024-00437633), and Center for Applied Research in Artificial Intelligence(CARAI) grant funded by Defense Acquisition Program Administration(DAPA) and Agency for Defense Development(ADD) (UD230017TD).
Gunhee Kim is the corresponding author.

\bibliography{aaai2026}

\clearpage
\appendix

\section{Approach}

\subsection{Forward Derivation}
In this section, we provide a full derivation for computing the integral $\int w_i(x)dx$ and $T_{i+1}$ in the spatial alpha blending process.
To simplify integration over the 2D Gaussian, we align our coordinate system with the axes of each splat’s window, similar to Analytic-Splatting \cite{analytic_splatting}.
In this coordinate system, we can separate and independently evaluate the integral $\int w_i(x)dx$ along each axis.

We first define the zeroth-, first- and second-order moments of the 1D Gaussian ($I^0_{\sigma}(a, b), I^1_{\sigma}(a, b), I^2_{\sigma}(a, b)$) as
\begin{align}
    I^0_{\sigma}(a, b) & = \int_{a}^{b} \exp(-\frac{x^2}{2 \sigma^2}) dx \notag \\
            & = \sqrt{\frac{\pi}{2}} \sigma \left[\mathrm{erf}\left(\frac{b}{\sqrt{2} \sigma}\right) - \mathrm{erf}\left(\frac{a}{\sqrt{2} \sigma}\right)\right], \notag \\
    I^1_{\sigma}(a, b) & = \int_{a}^{b} x \exp(-\frac{x^2}{2 \sigma^2}) dx \notag \\
            & = \sigma^2 \left(\exp(-\frac{a^2}{2 \sigma^2}) - \exp(-\frac{b^2}{2 \sigma^2})\right), \notag \\
    I^2_{\sigma}(a, b) & = \int_{a}^{b} x^2 \exp(-\frac{x^2}{2 \sigma^2}) dx \notag \\
            & = \sigma^2 \left( I^0_{\sigma}(a, b) + a \exp(-\frac{a^2}{2 \sigma^2}) - b \exp(-\frac{b^2}{2 \sigma^2})\right).
\end{align}
Using the zeroth-order moment of the 1D Gaussian $I^0_{\sigma}(a, b)$, we express the integrated weight as
\begin{align}
    \int w_i(x)dx & = \int_{u_1}^{u_2} \int_{v_1}^{v_2} t_i o_i\exp(-\frac{x^2}{2\sigma_1^2} -\frac{y^2}{2\sigma_2^2}) dxdy \notag \\
                  & = t_i o_i \int_{u_1}^{u_2} \exp(-\frac{x^2}{2\sigma_1^2}) dx \int_{v_1}^{v_2}\exp(-\frac{y^2}{2\sigma_2^2}) dy \notag \\
                  & = t_i o_i I^0_{\sigma_1}(u_1, u_2) I^0_{\sigma_2}(v_1, v_2).
\end{align}

Next, we can use the integrated weight to compute the zeroth-order moment of $T_i(x)(1 - \alpha_i(x))$ as
\begin{align}
    M^0_i & = \int_{u_1}^{u_2} \int_{v_1}^{v_2} t_i (1 - o_i \exp(-\frac{x^2}{2\sigma_1^2} -\frac{y^2}{2\sigma_2^2})) dxdy \notag \\
             & = t_i l_{i, 1} l_{i, 2} - \int_{u_1}^{u_2} \int_{v_1}^{v_2} t_i o_i \exp(-\frac{x^2}{2\sigma_1^2} -\frac{y^2}{2\sigma_2^2}) dxdy \notag \notag \\
             & = t_i l_{i, 1} l_{i, 2} - \int w_i(x)dx \notag \\
             & = t_i l_{i, 1} l_{i, 2} - t_i o_i I^0_{\sigma_1}(u_1, u_2) I^0_{\sigma_2}(v_1, v_2).
\end{align}
The zeroth-order moment $M^0_i$ means the accumulated transmittance value after rendering the $i$-th splat, which plays a similar role to the transmittance value $T_i^s$ in the scalar alpha blending.
This simplication proves that our method always maintains $\int T_{i+1}(x) dx = \int T_i(x) - w_i(x) dx$, which is crucial for the physical correctness of the rendering process.

We can also simplify the first-, and second-order moments ($M^1_i, M^2_i$) of $T_i(x)(1 - \alpha_i(x))$ as
\begin{align}
    M^1_i & = \int_{u_1}^{u_2} \int_{v_1}^{v_2} t_i (1 - o_i \exp(-\frac{x^2}{2\sigma_1^2} -\frac{y^2}{2\sigma_2^2})) \begin{bmatrix}x\\y\end{bmatrix} dxdy \notag                  \\
             & = t_i l_{i, 1} l_{i, 2} \begin{bmatrix}u\\v\end{bmatrix} \notag \\
             & \quad - t_i o_i
            \begin{bmatrix}
                \int_{u_1}^{u_2} x \exp(-\frac{x^2}{2\sigma_1^2}) dx \int_{v_1}^{v_2} \exp(-\frac{y^2}{2\sigma_2^2}) dy \\
                \int_{u_1}^{u_2} \exp(-\frac{x^2}{2\sigma_1^2}) dx \int_{v_1}^{v_2} y \exp(-\frac{y^2}{2\sigma_2^2}) dy
            \end{bmatrix} \notag \\
             & = t_i l_{i, 1} l_{i, 2} \begin{bmatrix}u\\v\end{bmatrix} - t_i o_i
            \begin{bmatrix}
                I^1_{\sigma_1}(u_1, u_2) \cdot I^0_{\sigma_2}(v_1, v_2) \\
                I^0_{\sigma_1}(u_1, u_2) \cdot I^1_{\sigma_2}(v_1, v_2)
            \end{bmatrix},
\end{align}
\begin{align}
    M^2_i & = \int_{u_1}^{u_2} \int_{v_1}^{v_2} t_i (1 - o_i \exp(-\frac{x^2}{2\sigma_1^2} -\frac{y^2}{2\sigma_2^2})) \begin{bmatrix}x^2\\y^2\end{bmatrix} dxdy \notag              \\
             & = t_i l_{i, 1} l_{i, 2} \left(\begin{bmatrix}u^2\\v^2\end{bmatrix} + \frac{1}{12} l_i^2\right) \notag \\
             & \quad - t_i o_i
            \begin{bmatrix}
                \int_{u_1}^{u_2} x^2 \exp(-\frac{x^2}{2\sigma_1^2}) dx \int_{v_1}^{v_2} \exp(-\frac{y^2}{2\sigma_2^2}) dy \\
                \int_{u_1}^{u_2} \exp(-\frac{x^2}{2\sigma_1^2}) dx \int_{v_1}^{v_2} y^2 \exp(-\frac{y^2}{2\sigma_2^2}) dy
            \end{bmatrix} \notag \\
             & = t_i l_{i, 1} l_{i, 2} \left(\begin{bmatrix}u^2\\v^2\end{bmatrix} + \frac{1}{12} l_i^2\right) \notag \\
             & \quad - t_i o_i
            \begin{bmatrix}
                I^2_{\sigma_1}(u_1, u_2) \cdot I^0_{\sigma_2}(v_1, v_2) \\
                I^0_{\sigma_1}(u_1, u_2) \cdot I^2_{\sigma_2}(v_1, v_2)
            \end{bmatrix}.
\end{align}

Then, the mean of $T_i(x)(1 - \alpha_i(x))$ in the $\alpha_i$-aligned coordinate system can be calculated as $\frac{M^1_i}{M^0_i}$, the variance as $\frac{M^2_i}{M^0_i} - \left(\frac{M^1_i}{M^0_i}\right)^2$, and the accumulated transmittance value as $M^0_i$.
Leveraging the properties of a 2D uniform distribution, we derive the 2D screen-space expression for $T_{i+1}$ as
\begin{subequations}
    \begin{align}
        x_{i+1} & = \mu_i' + [e_1, e_2] \frac{M^1_i}{M^0_i},                                            \\
        l_{i+1} & = \sqrt{12\left(\frac{M^2_i}{M^0_i} - \left(\frac{M^1_i}{M^0_i}\right)^2\right)}, \\
        t_{i+1} & = \frac{M^0_i}{l_{i+1,1} l_{i+1,2}}.
    \end{align}
\end{subequations}
Note that $x_{i+1}$ is rotated back to the original coordinate system using the inverse rotation matrix $[e_1, e_2]$.
This explicitly preserves the total transmittance mass, mean, and variance in terms of $t_i$, $x_i$, and $l_i$, ensuring a physically accurate rendering process.

\subsection{Forward Visualization}

\begin{figure}[t!]
    \centering
    \begin{tabular}{c}
        \includegraphics[width=0.45\textwidth]{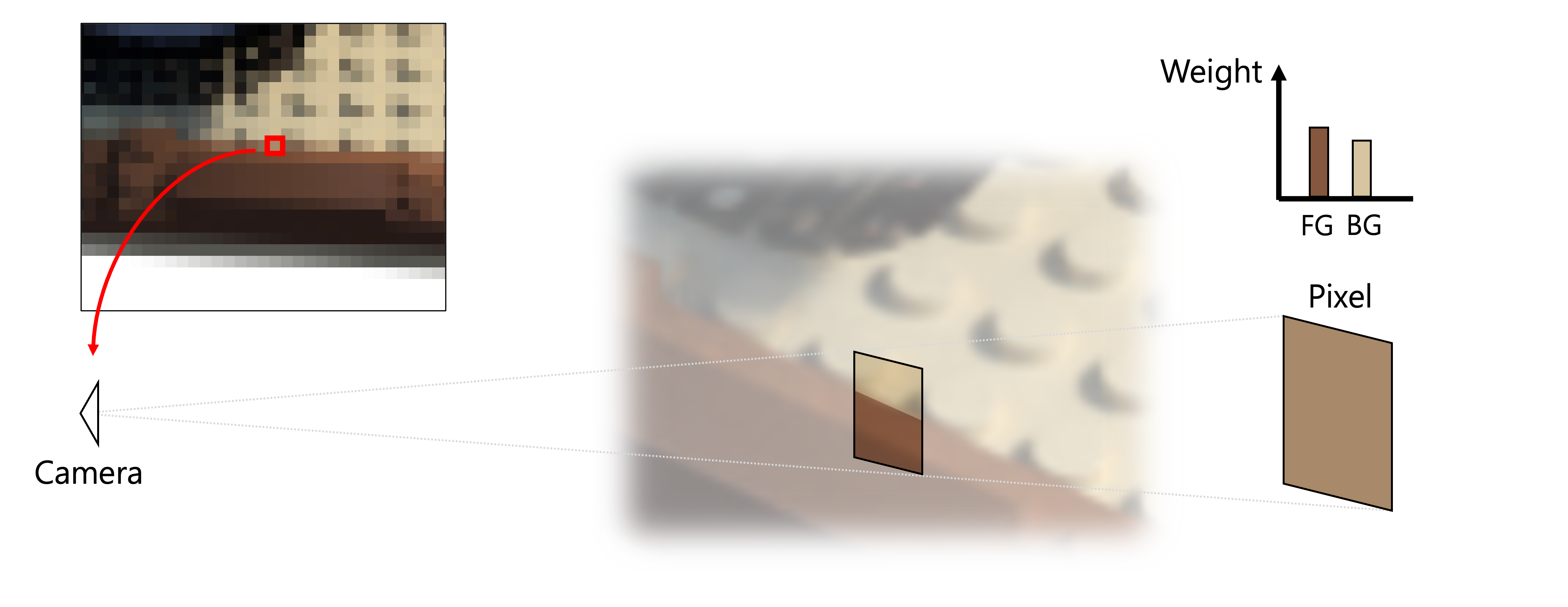} \\
        {(a) Real Camera} \\
        \includegraphics[width=0.45\textwidth]{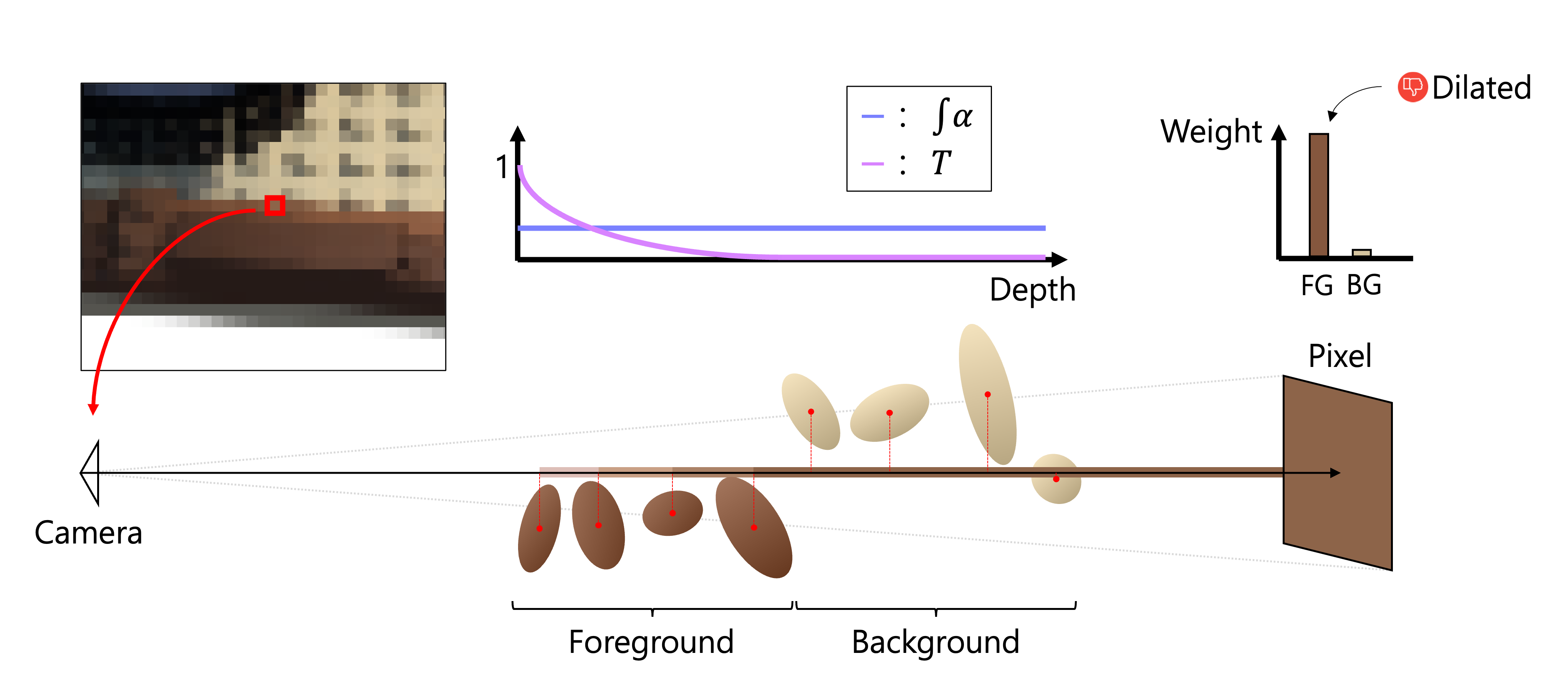} \\
        {(b) Scalar alpha blending} \\
        \includegraphics[width=0.45\textwidth]{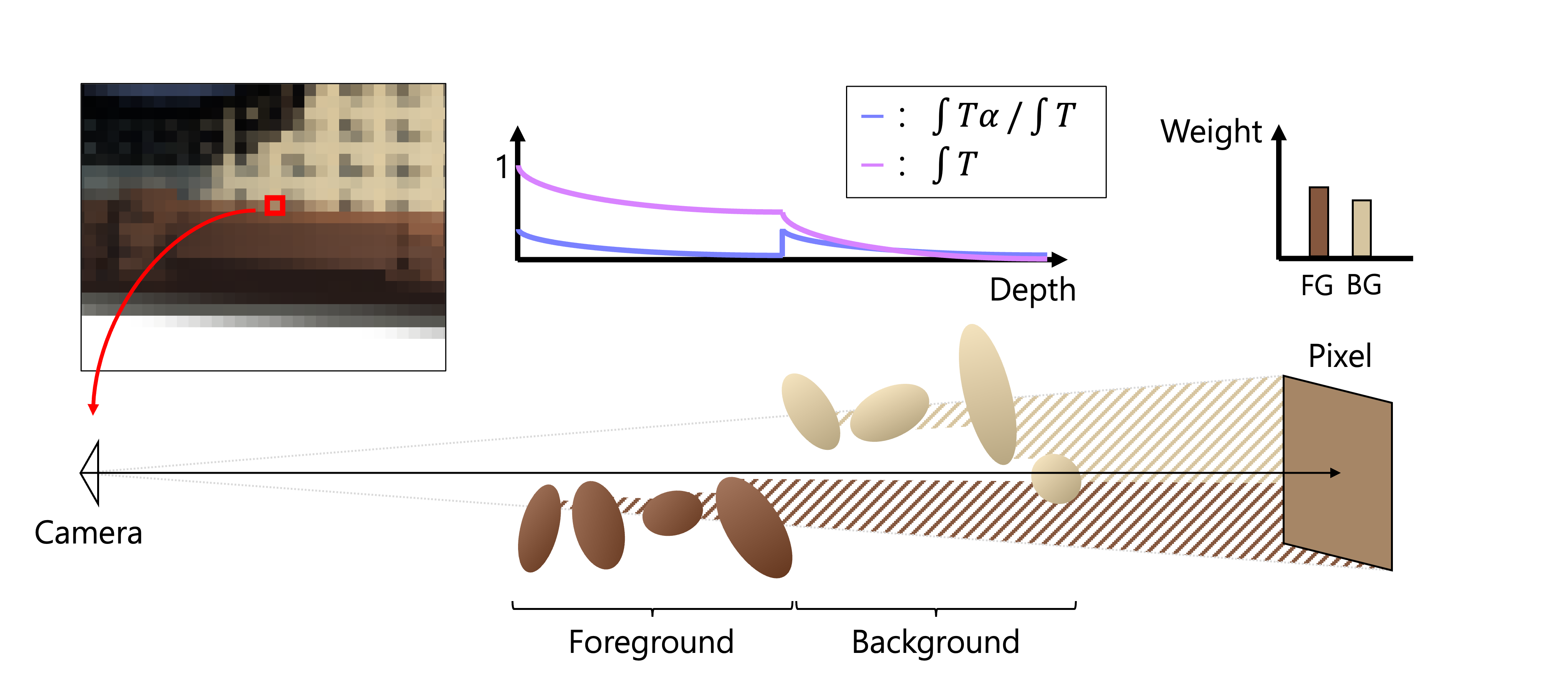} \\
        {(c) Gaussian Blending}
    \end{tabular}
    \caption{Comparison of rendering foreground splats and background splats using (a) a real camera, (b) scalar alpha blending, and (c) our Gaussian Blending.}
    \label{fig:rendering_detail}
\end{figure}

To provide deeper understanding of the difference between scalar alpha blending and our Gaussian Blending, we visualize the rendering process in Figure \ref{fig:rendering_detail}.
Scalar alpha blending first computes the scalar alpha value of each splat independently, and then blends the splats using this scalar alpha and scalar transmittance evaluated at the pixel center, which removes the spatial information of the splats.
As a result, splats with similar 2D Gaussian projections yield identical alpha values (denoted as $\int \alpha$ in Figure \ref{fig:rendering_detail}(b)), regardless of their spatial occlusions.
This approach causes undesired dilation of splats, rapidly saturating the transmittance value (denoted as $T$), and thus making background splats invisible in the final rendering.

In contrast, our Gaussian Blending considers alpha and transmittance as spatially varying distributions rather than scalar values.
During the rendering process, the transmittance is spatially consumed where each splat is located.
Consequently, overlapping splats have significantly lower average alpha values (denoted as $\int T\alpha / \int T$ in Figure \ref{fig:rendering_detail}(c)), ensuring that non-overlapping background splats remain clearly visible in the final rendering.

\begin{figure}[t!]
    \centering
    \begin{tabular}{cc}
        \raisebox{4em}{(a)} & \includegraphics[width=0.4\textwidth]{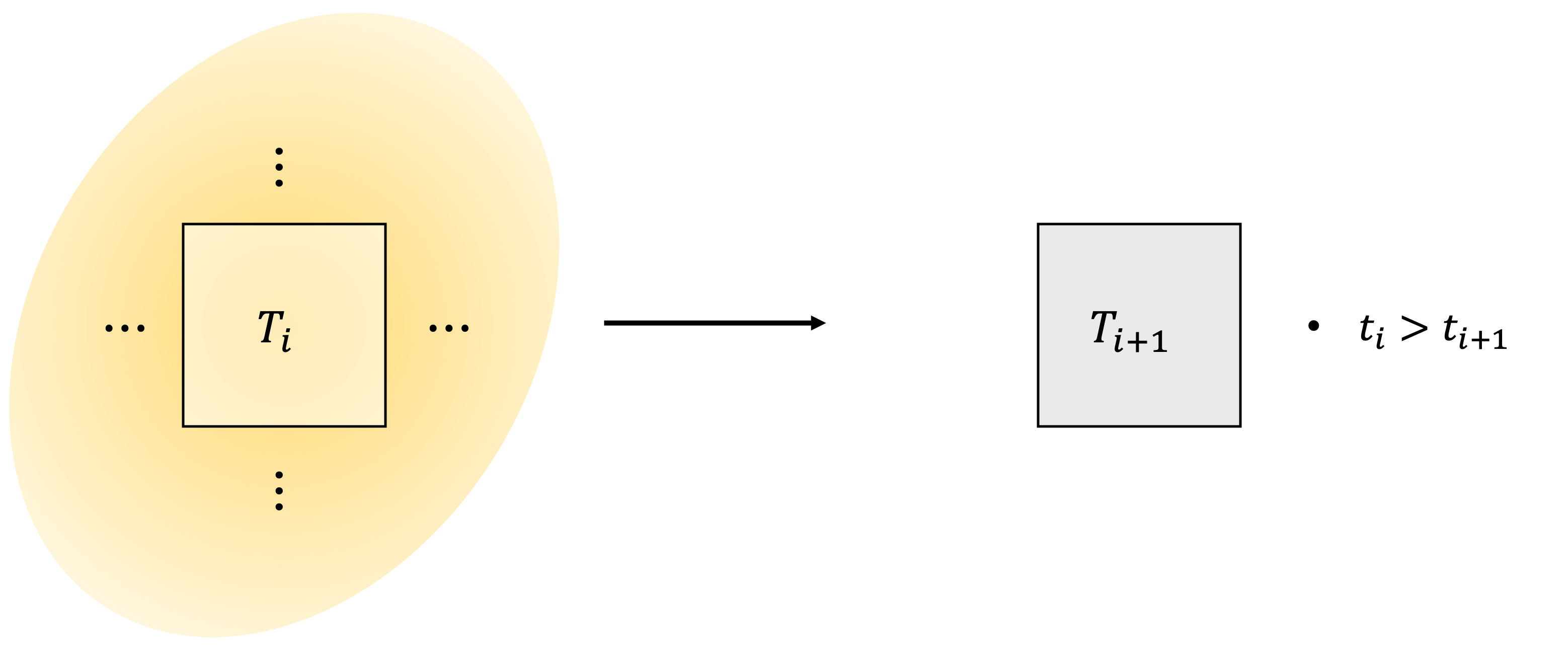} \\
        \raisebox{4em}{(b)} & \includegraphics[width=0.4\textwidth]{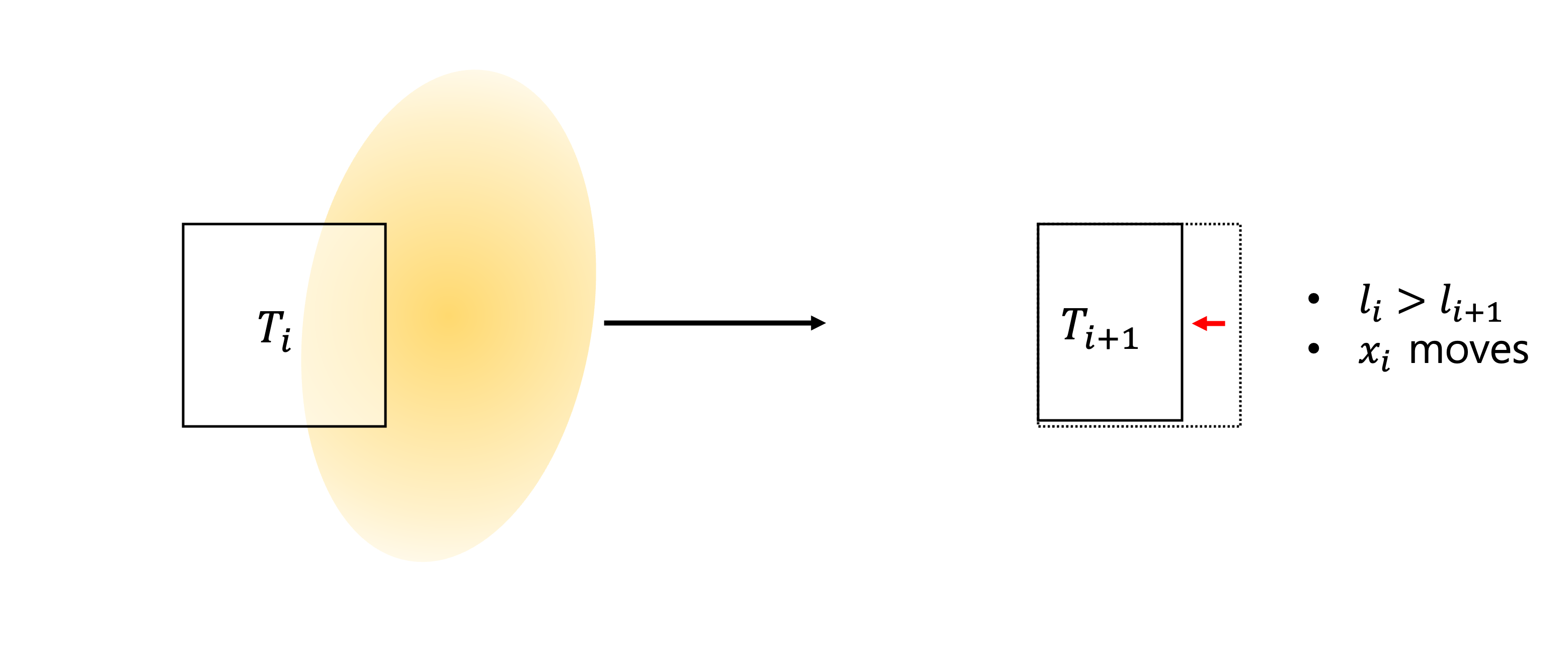} \\
        \raisebox{4em}{(c)} & \includegraphics[width=0.4\textwidth]{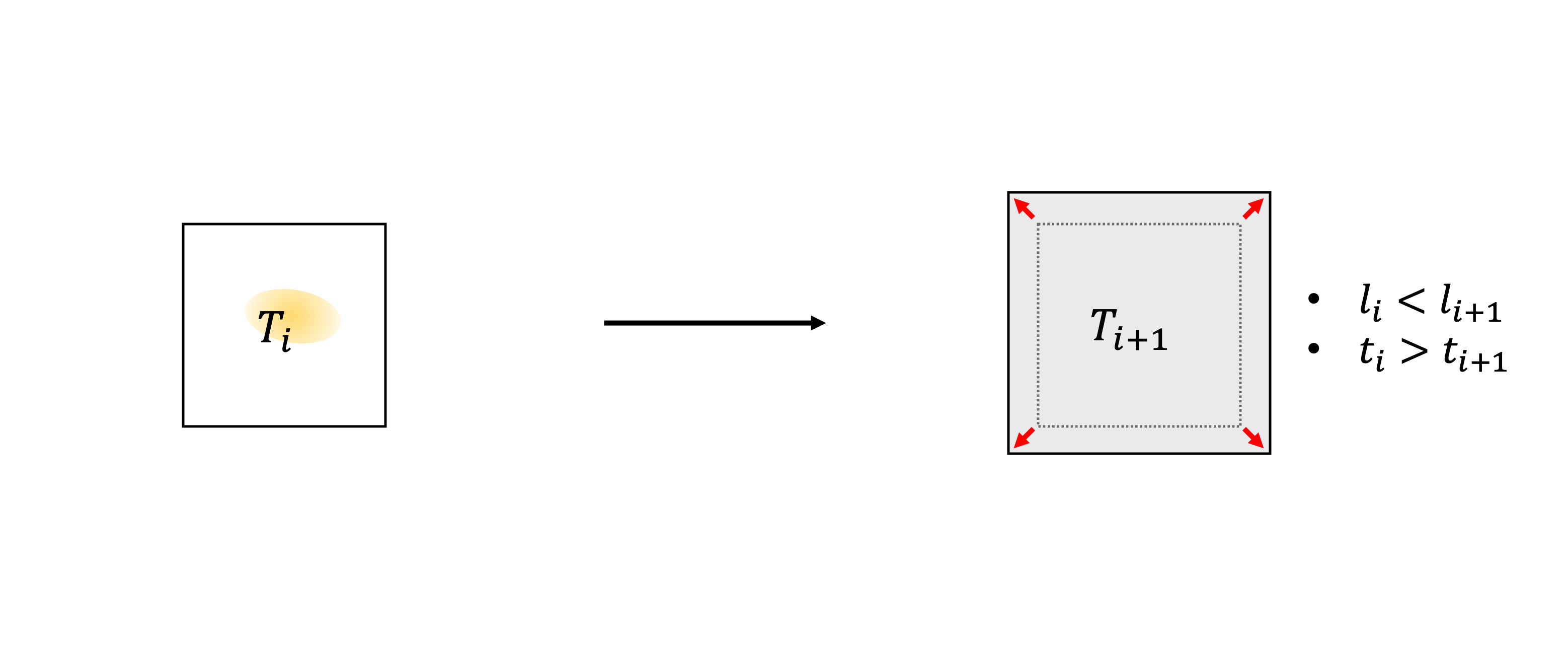} \\
    \end{tabular}
    \caption{Example of transmittance calculation for three different cases.}
    \label{fig:transmittance_example}
\end{figure}

We also illustrate several examples of the transition from $T_i$ to $T_{i+1}$ in Figure \ref{fig:transmittance_example}.
First, when a very large splat overlaps entirely with $T_i$, the splat does not affect the window range of the transmittance but lowers the overall transmittance value $t_i$.
Second, when a splat partially overlaps with $T_i$ from the side, the transmittance is reduced in the overlapped region, causing $T_{i+1}$ to narrow toward the non-overlapping region.
Moreover, despite using a 2D uniform distribution, our method can also handle cases where small splats create holes in the transmittance.
Specifically, when a very small splat intersects with $T_i$, the outer region without the splat dominates the transmittance, thus enlarging the window of $T_{i+1}$ and reducing the influence of the region occupied by the splat in subsequent rendering steps.

\section{Implementation Details}

\subsection{Numerical Stability and Optimization}
Analytic-Splatting approximates the Gaussian integral using a sigmoid-like function, as the integral of the Gaussian distribution cannot be expressed in a closed form.
In contrast, our method directly leverages the error function (erf), resulting in a simpler, more accurate, and numerically stable calculation of the Gaussian integral.
Additionally, when the window range of $T_i$ becomes extremely small or large relative to a splat, the calculation of $T_{i+1}$ may become numerically unstable.
Thus, when the range falls outside $[0.1\sigma, 10^6\sigma]$, we assume that the splat does not alter the window range of $T_i$ and instead perform scalar alpha blending at the center of $T_i$.

Although our Gaussian Blending adjusts the window from $T_i$ to $T_{i+1}$ during the forward pass, accurately reconstructing $T_i$ from $T_{i+1}$ in the backward pass becomes significantly challenging.
A fully accurate backward process requires intermediate $T_i$ buffers and sequentially propagating gradients using the chain rule from back to front.
This backward pass is both slow and memory-intensive, and also causes unstable training due to accumulated errors in the gradient computation.
To resolve this issue, we employ a stop-gradient operation on the window range calculation.
With this stop-gradient, as in 3DGS, each splat can independently compute both direct and indirect influences on pixels, lowering the gradient error accumulation.
Additionally, by calculating gradients from front to back, we can completely eliminate the intermediate buffers, making the backward pass equivalent to conventional 3DGS and its variants in terms of memory usage and time complexity.
Empirically, full backward and stop-gradient-based backward have no noticeable performance degradation, except in speed and stability.
As a result, our method achieves a comparable or even faster rendering speed to 3DGS and its variants, with approximately 123 FPS.

\subsection{Datasets}
We find that previous works, including 3DGS \cite{3dgs} and its variants such as Mip-Splatting \cite{mip_splatting} and Analytic-Splatting \cite{analytic_splatting}, compute LPIPS incorrectly due to improper normalization \cite{wrong_lpips}.
For a fair comparison, we recompute the metrics with intended normalization for all models using the same normalization.
We evaluate PSNR, SSIM, and LPIPS across all methods using identical hyperparameters and VGG backbone for LPIPS calculation.

Furthermore, the existing multi-scale Blender dataset \cite{nerf, mip_nerf} performs box downsampling without accounting for the alpha, causing darkening artifacts at object boundaries.
To resolve this issue, we incorporate alpha values into the box downsampling process, resulting in correct appearances at object boundaries regardless of background.
Additionally, we note that unlike the multi-scale Blender dataset, the multi-scale Mip-NeRF 360 dataset \cite{mip_nerf_360} uses interpolation for downsampling.
To better emulate real camera sensor behavior, we also apply box downsampling to the Mip-NeRF 360 dataset.
For fair comparison, we retrain and evaluate all models using these consistently downsampled datasets.

For further implementation details, we refer readers to our code.

\section{Experiments}

\subsection{Experiment Setup}
We run all experiments on a single NVIDIA RTX A6000 GPU with 48 GB of memory.
The baseline models are trained and evaluated using their publicly available official codebases and default hyperparameters.
For our proposed model, we individually validate each hyperparameter by comparing performance with values scaled by factors of $\times 2$ and $\times \frac{1}{2}$, selecting the hyperparameter value that yields the best or comparable performance.
We run experiments for each combination of scene and resolution setting, and the reported metrics are averaged per scene and/or resolution.

\subsection{Comparison with Supersampling}

\begin{figure*}[ht!]
    \centering
    \begin{tikzpicture}
        \begin{axis}[
            xlabel={Supersampling Scale},
            ylabel={PSNR},
            legend pos={south east},
            legend style={
                fill opacity=0.8,
                font=\small,
                row sep=0pt,
                inner sep=2pt,
            },
            grid=major,
            width=0.92\textwidth,
            height=0.35\textwidth,
            xtick={1,2,3,4,5},
            xmin=0.55, xmax=5.5,
        ]
        \addplot[color=orange,mark=x,thick] coordinates {
            (1, 35.81)
            (2, 39.41)
            (3, 39.71)
            (4, 39.45)
            (5, 39.02)
        };
        \addlegendentry{Gaussian Blending}
        \addplot[color=cyan,mark=x,thick] coordinates {
            (1, 27.22)
            (2, 33.62)
            (3, 36.71)
            (4, 38.39)
            (5, 39.36)
        };
        \addlegendentry{Analytic-Splatting}
        \addplot[color=blue,mark=x,thick] coordinates {
            (1, 17.74)
            (2, 22.41)
            (3, 25.50)
            (4, 27.72)
            (5, 29.42)
        };
        \addlegendentry{3DGS}
        \end{axis}
    \end{tikzpicture}
    \renewcommand{\arraystretch}{0.0}
    \begin{tabular}{@{\hskip 25pt}cccccc}
        \raisebox{1.2em}{\multirow{2}{*}{\rotatebox[origin=c]{90}{3DGS}}}
        & \includegraphics[width=0.145\textwidth]{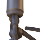}
        & \includegraphics[width=0.145\textwidth]{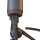}
        & \includegraphics[width=0.145\textwidth]{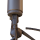}
        & \includegraphics[width=0.145\textwidth]{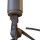}
        & \includegraphics[width=0.145\textwidth]{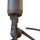} \\
        & \includegraphics[width=0.145\textwidth]{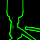}
        & \includegraphics[width=0.145\textwidth]{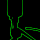}
        & \includegraphics[width=0.145\textwidth]{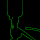}
        & \includegraphics[width=0.145\textwidth]{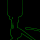}
        & \includegraphics[width=0.145\textwidth]{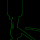} \\ [2pt]
        \raisebox{3.8em}{\multirow{2}{*}{\rotatebox[origin=c]{90}{Analytic-Splatting}}}
        & \includegraphics[width=0.145\textwidth]{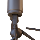}
        & \includegraphics[width=0.145\textwidth]{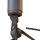}
        & \includegraphics[width=0.145\textwidth]{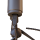}
        & \includegraphics[width=0.145\textwidth]{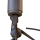}
        & \includegraphics[width=0.145\textwidth]{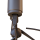} \\
        & \includegraphics[width=0.145\textwidth]{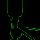}
        & \includegraphics[width=0.145\textwidth]{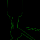}
        & \includegraphics[width=0.145\textwidth]{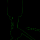}
        & \includegraphics[width=0.145\textwidth]{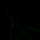}
        & \includegraphics[width=0.145\textwidth]{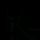} \\ [2pt]
        \raisebox{3.8em}{\multirow{2}{*}{\rotatebox[origin=c]{90}{Gaussian Blending}}}
        & \includegraphics[width=0.145\textwidth]{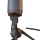}
        & \includegraphics[width=0.145\textwidth]{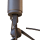}
        & \includegraphics[width=0.145\textwidth]{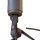}
        & \includegraphics[width=0.145\textwidth]{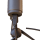}
        & \includegraphics[width=0.145\textwidth]{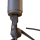} \\
        & \includegraphics[width=0.145\textwidth]{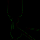}
        & \includegraphics[width=0.145\textwidth]{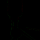}
        & \includegraphics[width=0.145\textwidth]{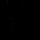}
        & \includegraphics[width=0.145\textwidth]{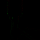}
        & \includegraphics[width=0.145\textwidth]{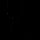} \\ [2pt]
        & {original} & {$2 \times 2$} & {$3 \times 3$} & {$4 \times 4$} & {$5 \times 5$}
    \end{tabular}
    \caption{Comparison of supersampling results on the multi-scale Blender dataset trained at $\times 1$ resolution and rendered at $\times 1/8$ resolution. The dilated region is marked in green and the eroded region is marked in red.}
    \label{fig:supersampling_qualitative}
\end{figure*}

Supersampling is widely used to enhance rendering quality and reduce aliasing artifacts by first rendering scenes at higher resolutions and then downsampling to the target resolution.
Rendering at higher resolutions allows models to independently integrate alpha and transmittance values in 2D screen space, thus reducing the pixel-level dilation issues caused by scalar alpha blending.
Nevertheless, supersampling significantly increases rendering time, posing a substantial drawback for real-time applications.

As shown qualitatively in Figure \ref{fig:supersampling_qualitative} and quantitatively in Table \ref{tab:supersampling_blender}, our Gaussian Blending demonstrates superior effectiveness and efficiency compared to existing methods with supersampling such as 3DGS and Analytic-Splatting.
Specifically, Analytic-Splatting requires at least $3 \times 3$ supersampling to achieve rendering quality comparable to our Gaussian Blending without supersampling, while 3DGS needs even higher levels of supersampling (over $5 \times 5$) for similar visual quality.
Furthermore, our Gaussian Blending is at least three times faster than Analytic-Splatting with $3 \times 3$ supersampling and 3DGS with $5 \times 5$ supersampling, all while maintaining comparable or better rendering quality.

Remarkably, our Gaussian Blending achieves high-quality rendering without relying on supersampling, demonstrating clear suitability for real-time applications.
Additionally, when employing $2 \times 2$ supersampling, Gaussian Blending can better handle the smaller splat case (in Figure \ref{fig:transmittance_example}(c)), achieving even higher rendering quality with a real-time rendering speed of 60 FPS.

\subsection{Additional Results}
In the main paper, to independently demonstrate zoom-out and zoom-in scenarios, we present results of models trained at the highest and lowest resolutions, respectively, and evaluate them in a multi-scale testing setting.
However, our model shows superior rendering performance even when trained at arbitrary resolutions and also tested at arbitrary resolutions.
As shown in Table \ref{tab:stmt2_blender} and Table \ref{tab:stmt4_blender}, our model consistently achieves state-of-the-art results when simultaneously performing zoom-out and zoom-in by training at intermediate resolutions (e.g., 1/2 or 1/4 resolution) and evaluating at multiple scales.
Moreover, our model also produces the highest rendering results when trained and tested at arbitrary sampling rates on the complex multi-scale Mip-NeRF 360 dataset (Table \ref{tab:stmt1_mipnerf360} - \ref{tab:stmt8_mipnerf360}).

\subsection{Limitations}

\begin{figure}[t!]
    \centering
    \includegraphics[width=0.47\textwidth]{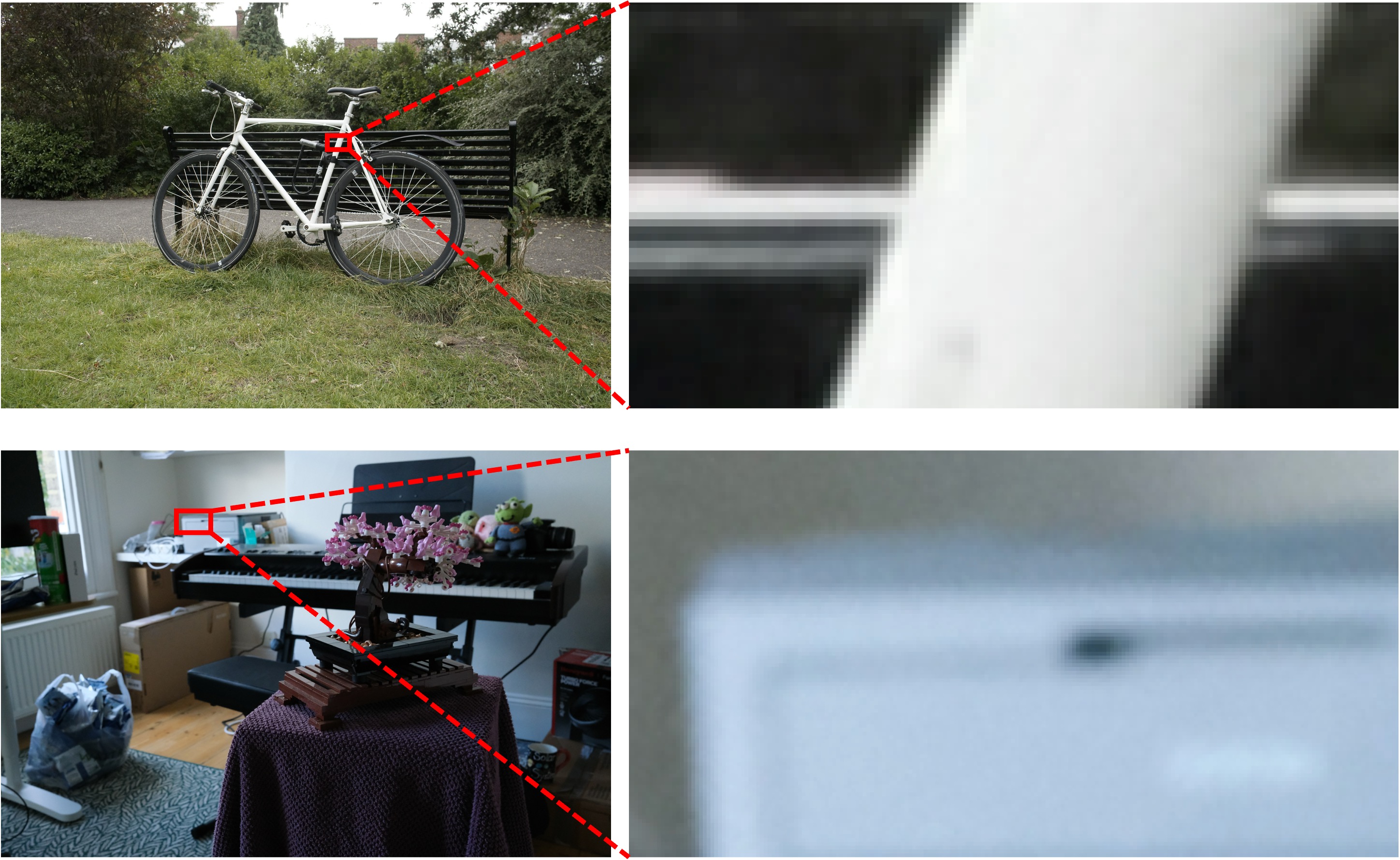}
    \caption{Due to the real camera effects, the Mip-NeRF 360 dataset contains inconsistent blur across different views.}
    \label{fig:mipnerf360_blur}
\end{figure}

Theoretically, the pinhole camera model should contain blurred appearance only when the object itself has smooth appearance, or when its edge is partially overlapping with the pixel.
However, in Mip-NeRF 360 dataset, the real camera effects, such as diffraction, chromatic aberration, and defocus blur, cause inconsistent blur across different views as in Figure \ref{fig:mipnerf360_blur}.
Specifically, object edges are blurred across the pixels that are not even overlapping with the object.
This negatively affects the scene reconstruction, where the inconsistent blur causes the splats to be misaligned with the object edges depending on the view.
As a result, our model's performance gap in the multi-scale Mip-NeRF 360 dataset is not as large as in the multi-scale Blender dataset.
However, our model still outperforms all other methods in the Mip-NeRF 360 dataset, demonstrating its robustness to real camera effects.

\begin{figure*}[t]
    \centering
\begin{tabular}{ccccc}
        & {$\times 1$} & {$\times 1/2$} & {$\times 1/4$} & {$\times 1/8$} \\
        \raisebox{4.2em}{\rotatebox[origin=c]{90}{Gaussian Blending}}
        & \includegraphics[width=0.18\textwidth]{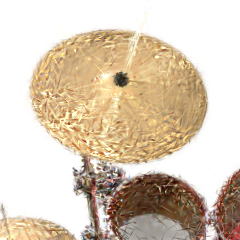}
        & \includegraphics[width=0.18\textwidth]{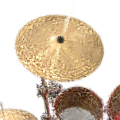}
        & \includegraphics[width=0.18\textwidth]{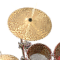}
        & \includegraphics[width=0.18\textwidth]{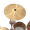} \\
        \raisebox{4.5em}{\rotatebox[origin=c]{90}{GT}}
        & \includegraphics[width=0.18\textwidth]{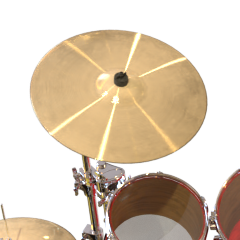}
        & \includegraphics[width=0.18\textwidth]{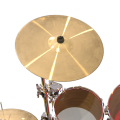}
        & \includegraphics[width=0.18\textwidth]{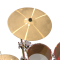}
        & \includegraphics[width=0.18\textwidth]{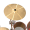} \\
        \midrule
        \raisebox{4.2em}{\rotatebox[origin=c]{90}{Gaussian Blending}}
        & \includegraphics[width=0.18\textwidth]{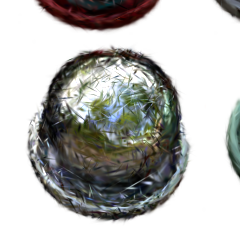}
        & \includegraphics[width=0.18\textwidth]{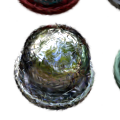}
        & \includegraphics[width=0.18\textwidth]{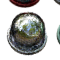}
        & \includegraphics[width=0.18\textwidth]{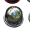} \\
        \raisebox{4.5em}{\rotatebox[origin=c]{90}{GT}}
        & \includegraphics[width=0.18\textwidth]{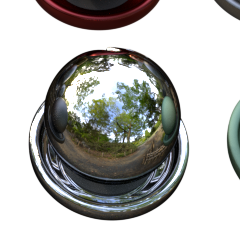}
        & \includegraphics[width=0.18\textwidth]{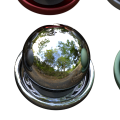}
        & \includegraphics[width=0.18\textwidth]{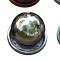}
        & \includegraphics[width=0.18\textwidth]{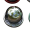} \\
    \end{tabular}
    \caption{Example of failure cases when we try to zoom-in on the scene containing highly specular objects. Our model is trained at $\times 1/8$ resolution.}
    \label{fig:failure_cases}
\end{figure*}

We also observe that our model struggles to reconstruct highly specular objects when zooming in on the scene.
Due to the ill-posed nature of the zoom-in task and the limited high-frequency information in the low-resolution training data, our model fails to accurately reconstruct the specular highlights as shown in Figure \ref{fig:failure_cases}.
However, our model still outperforms all other methods in this scenario, demonstrating its robustness to the zoom-in task.

We leave the improvement of real camera effects and zoom-in reconstruction of specular objects to future work, as these are orthogonal problems to our target of improving alpha blending to consider intra-pixel variations.

\begin{table*}[ht!]
    \centering
    \caption{Supersampling results on the multi-scale Blender dataset with single-scale training ($\times 1$ resolution) and multi-scale testing setting. Each model is represented with its supersampling scale.}
    \label{tab:supersampling_blender}
    \resizebox{\textwidth}{!}{
        \begin{tabular}{l|ccccc|ccccc|ccccc|c}
            \toprule
            \multirow{2}{*}{} & \multicolumn{5}{c|}{PSNR $\uparrow$} & \multicolumn{5}{c|}{SSIM $\uparrow$} & \multicolumn{5}{c|}{LPIPS $\downarrow$}                                                                                                                                                                                                             \\
                                    & $\times 1$                           & $\times 1/2$                         & $\times 1/4$                           & $\times 1/8$   & Avg.
                                    & $\times 1$                           & $\times 1/2$                         & $\times 1/4$                           & $\times 1/8$   & Avg.
                                    & $\times 1$                           & $\times 1/2$                         & $\times 1/4$                           & $\times 1/8$   & Avg.           & FPS                                                                                                                                                                     \\
            \midrule
            3DGS
                                    & 33.57                                & 27.04                                & 21.43                                  & 17.74          & 24.95          & 0.970          & 0.950          & 0.876          & 0.767          & 0.891          & 0.037          & 0.036          & 0.071          & 0.130          & 0.068          & 136.75 \\
            3DGS ($2 \times 2$)
                                    & 33.86                                & 31.47                                & 26.53                                  & 22.41          & 28.57          & 0.971          & 0.972          & 0.952          & 0.906          & 0.950          & 0.036          & 0.024          & 0.035          & 0.064          & 0.040          & 81.05  \\
\bottomrule
        \end{tabular}
    }
\end{table*}

\begin{table*}[ht!]
    \centering
    \caption{Single-scale training ($\times 1$ resolution) and multi-scale testing results on the multi-scale Blender dataset. The best results are highlighted in \textbf{bold}.}
    \label{tab:stmt1_blender}
    \resizebox{\textwidth}{!}{
        \begin{tabular}{l|ccccc|ccccc|ccccc}
            \toprule
            \multirow{2}{*}{} & \multicolumn{5}{c|}{PSNR $\uparrow$} & \multicolumn{5}{c|}{SSIM $\uparrow$} & \multicolumn{5}{c}{LPIPS $\downarrow$}                                                                                                                                                                                                             \\
                                    & $\times 1$                           & $\times 1/2$                         & $\times 1/4$                           & $\times 1/8$   & Avg.
                                    & $\times 1$                           & $\times 1/2$                         & $\times 1/4$                           & $\times 1/8$   & Avg.
                                    & $\times 1$                           & $\times 1/2$                         & $\times 1/4$                           & $\times 1/8$   & Avg.                                                                                                                                                                                     \\
            \midrule
            TensoRF
                                    & 33.22                                & 33.27                                & 30.23                                  & 26.77          & 30.87          & 0.964          & 0.972          & 0.968          & 0.953          & 0.964          & 0.047          & 0.034          & 0.049          & 0.079          & 0.052          \\
            3DGS
                                    & 33.57                                & 27.04                                & 21.43                                  & 17.74          & 24.95          & 0.970          & 0.950          & 0.876          & 0.767          & 0.891          & 0.037          & 0.036          & 0.071          & 0.130          & 0.068          \\
\bottomrule
        \end{tabular}
    }
\end{table*}

\begin{table*}[ht!]
    \centering
    \caption{Single-scale training ($\times 1/2$ resolution) and multi-scale testing results on the multi-scale Blender dataset. The best results are highlighted in \textbf{bold}.}
    \label{tab:stmt2_blender}
    \resizebox{\textwidth}{!}{
        \begin{tabular}{l|ccccc|ccccc|ccccc}
            \toprule
            \multirow{2}{*}{} & \multicolumn{5}{c|}{PSNR $\uparrow$} & \multicolumn{5}{c|}{SSIM $\uparrow$} & \multicolumn{5}{c}{LPIPS $\downarrow$}                                                                                                                                                                                                             \\
                                    & $\times 1$                           & $\times 1/2$                         & $\times 1/4$                           & $\times 1/8$   & Avg.
                                    & $\times 1$                           & $\times 1/2$                         & $\times 1/4$                           & $\times 1/8$   & Avg.
                                    & $\times 1$                           & $\times 1/2$                         & $\times 1/4$                           & $\times 1/8$   & Avg.                                                                                                                                                                                     \\
            \midrule
            TensoRF
                                    & 31.83                                & 34.11                                & 32.03                                  & 27.84          & 31.45          & 0.956          & 0.972          & 0.974          & 0.961          & 0.966          & 0.062          & 0.033          & 0.037          & 0.066          & 0.049          \\
            3DGS
                                    & 26.66                                & 34.08                                & 24.62                                  & 19.04          & 26.10          & 0.937          & 0.975          & 0.930          & 0.814          & 0.914          & 0.058          & 0.023          & 0.045          & 0.103          & 0.058          \\
\bottomrule
        \end{tabular}
    }
\end{table*}

\begin{table*}[ht!]
    \centering
    \caption{Single-scale training ($\times 1/4$ resolution) and multi-scale testing results on the multi-scale Blender dataset. The best results are highlighted in \textbf{bold}.}
    \label{tab:stmt4_blender}
    \resizebox{\textwidth}{!}{
        \begin{tabular}{l|ccccc|ccccc|ccccc}
            \toprule
            \multirow{2}{*}{} & \multicolumn{5}{c|}{PSNR $\uparrow$} & \multicolumn{5}{c|}{SSIM $\uparrow$} & \multicolumn{5}{c}{LPIPS $\downarrow$}                                                                                                                                                                                                             \\
                                    & $\times 1$                           & $\times 1/2$                         & $\times 1/4$                           & $\times 1/8$   & Avg.
                                    & $\times 1$                           & $\times 1/2$                         & $\times 1/4$                           & $\times 1/8$   & Avg.
                                    & $\times 1$                           & $\times 1/2$                         & $\times 1/4$                           & $\times 1/8$   & Avg.                                                                                                                                                                                     \\
            \midrule
            TensoRF
                                    & 28.51                                & 30.54                                & 33.73                                  & 30.49          & 30.82          & 0.922          & 0.951          & 0.974          & 0.972          & 0.955          & 0.111          & 0.070          & 0.032          & 0.044          & 0.064          \\
            3DGS
                                    & 21.56                                & 24.76                                & 34.49                                  & 22.70          & 25.88          & 0.866          & 0.920          & 0.979          & 0.908          & 0.918          & 0.114          & 0.061          & 0.019          & 0.059          & 0.063          \\
\bottomrule
        \end{tabular}
    }
\end{table*}

\begin{table*}[ht!]
    \centering
    \caption{Single-scale training ($\times 1/8$ resolution) and multi-scale testing results on the multi-scale Blender dataset. The best results are highlighted in \textbf{bold}.}
    \label{tab:stmt8_blender}
    \resizebox{\textwidth}{!}{
        \begin{tabular}{l|ccccc|ccccc|ccccc}
            \toprule
            \multirow{2}{*}{} & \multicolumn{5}{c|}{PSNR $\uparrow$} & \multicolumn{5}{c|}{SSIM $\uparrow$} & \multicolumn{5}{c}{LPIPS $\downarrow$}                                                                                                                                                                                                             \\
                                    & $\times 1$                           & $\times 1/2$                         & $\times 1/4$                           & $\times 1/8$   & Avg.
                                    & $\times 1$                           & $\times 1/2$                         & $\times 1/4$                           & $\times 1/8$   & Avg.
                                    & $\times 1$                           & $\times 1/2$                         & $\times 1/4$                           & $\times 1/8$   & Avg.                                                                                                                                                                                     \\
            \midrule
            TensoRF
                                    & 25.42                                & 26.57                                & 29.02                                  & 32.87          & 28.47          & 0.874          & 0.902          & 0.946          & 0.977          & 0.925          & 0.159          & 0.132          & 0.078          & 0.028          & 0.099          \\
            3DGS
                                    & 18.51                                & 19.90                                & 23.33                                  & 34.72          & 24.11          & 0.824          & 0.832          & 0.905          & 0.984          & 0.886          & 0.163          & 0.136          & 0.072          & 0.016          & 0.097          \\
\bottomrule
        \end{tabular}
    }
\end{table*}

\begin{table*}[ht!]
    \centering
    \caption{Multi-scale training and multi-scale testing results on the multi-scale Blender dataset. The best results are highlighted in \textbf{bold}.}
    \label{tab:mtmt_blender}
    \resizebox{\textwidth}{!}{
        \begin{tabular}{l|ccccc|ccccc|ccccc}
            \toprule
            \multirow{2}{*}{} & \multicolumn{5}{c|}{PSNR $\uparrow$} & \multicolumn{5}{c|}{SSIM $\uparrow$} & \multicolumn{5}{c}{LPIPS $\downarrow$}                                                                                                                                                                                                             \\
                                    & $\times 1$                           & $\times 1/2$                         & $\times 1/4$                           & $\times 1/8$   & Avg.
                                    & $\times 1$                           & $\times 1/2$                         & $\times 1/4$                           & $\times 1/8$   & Avg.
                                    & $\times 1$                           & $\times 1/2$                         & $\times 1/4$                           & $\times 1/8$   & Avg.                                                                                                                                                                                     \\
            \midrule
            TensoRF
                                    & 33.03                                & 33.92                                & 30.99                                  & 27.26          & 31.30          & 0.963          & 0.973          & 0.971          & 0.957          & 0.966          & 0.049          & 0.032          & 0.043          & 0.073          & 0.049          \\
            3DGS
                                    & 30.19                                & 31.38                                & 30.59                                  & 27.05          & 29.90          & 0.954          & 0.967          & 0.968          & 0.950          & 0.960          & 0.061          & 0.037          & 0.031          & 0.041          & 0.042          \\
\bottomrule
        \end{tabular}
    }
\end{table*}

\begin{table*}[ht!]
    \centering
    \caption{Single-scale training ($\times 1$ resolution) and multi-scale testing results on the multi-scale Mip-NeRF 360 dataset. The best results are highlighted in \textbf{bold}.}
    \label{tab:stmt1_mipnerf360}
    \resizebox{\textwidth}{!}{
        \begin{tabular}{l|ccccc|ccccc|ccccc}
            \toprule
            \multirow{2}{*}{} & \multicolumn{5}{c|}{PSNR $\uparrow$} & \multicolumn{5}{c|}{SSIM $\uparrow$} & \multicolumn{5}{c}{LPIPS $\downarrow$}                                                                                                                                                                                                             \\
                                    & $\times 1$                           & $\times 1/2$                         & $\times 1/4$                           & $\times 1/8$   & Avg.
                                    & $\times 1$                           & $\times 1/2$                         & $\times 1/4$                           & $\times 1/8$   & Avg.
                                    & $\times 1$                           & $\times 1/2$                         & $\times 1/4$                           & $\times 1/8$   & Avg.                                                                                                                                                                                     \\
            \midrule
            3DGS
                                    & 27.63                                & 25.76                                & 21.98                                  & 19.34          & 23.68          & 0.820          & 0.815          & 0.703          & 0.570          & 0.727          & 0.259          & 0.208          & 0.231          & 0.268          & 0.242          \\
            Scaffold-GS
                                    & 27.27                                & 26.05                                & 22.48                                  & 19.70          & 23.87          & 0.804          & 0.812          & 0.712          & 0.583          & 0.728          & 0.284          & 0.216          & 0.219          & 0.254          & 0.243          \\
\bottomrule
        \end{tabular}
    }
\end{table*}

\begin{table*}[ht!]
    \centering
    \caption{Single-scale training ($\times 1/2$ resolution) and multi-scale testing results on the multi-scale Mip-NeRF 360 dataset. The best results are highlighted in \textbf{bold}.}
    \label{tab:stmt2_mipnerf360}
    \resizebox{\textwidth}{!}{
        \begin{tabular}{l|ccccc|ccccc|ccccc}
            \toprule
            \multirow{2}{*}{} & \multicolumn{5}{c|}{PSNR $\uparrow$} & \multicolumn{5}{c|}{SSIM $\uparrow$} & \multicolumn{5}{c}{LPIPS $\downarrow$}                                                                                                                                                                                                             \\
                                    & $\times 1$                           & $\times 1/2$                         & $\times 1/4$                           & $\times 1/8$   & Avg.
                                    & $\times 1$                           & $\times 1/2$                         & $\times 1/4$                           & $\times 1/8$   & Avg.
                                    & $\times 1$                           & $\times 1/2$                         & $\times 1/4$                           & $\times 1/8$   & Avg.                                                                                                                                                                                     \\
            \midrule
            3DGS
                                    & 24.50                                & 28.55                                & 24.47                                  & 20.53          & 24.51          & 0.749          & 0.864          & 0.810          & 0.650          & 0.768          & 0.323          & 0.169          & 0.169          & 0.227          & 0.222          \\
            Scaffold-GS
                                    & 25.03                                & 28.41                                & 24.71                                  & 20.72          & 24.72          & 0.740          & 0.853          & 0.812          & 0.657          & 0.765          & 0.340          & 0.185          & 0.165          & 0.217          & 0.227          \\
\bottomrule
        \end{tabular}
    }
\end{table*}

\begin{table*}[ht!]
    \centering
    \caption{Single-scale training ($\times 1/4$ resolution) and multi-scale testing results on the multi-scale Mip-NeRF 360 dataset. The best results are highlighted in \textbf{bold}.}
    \label{tab:stmt4_mipnerf360}
    \resizebox{\textwidth}{!}{
        \begin{tabular}{l|ccccc|ccccc|ccccc}
            \toprule
            \multirow{2}{*}{} & \multicolumn{5}{c|}{PSNR $\uparrow$} & \multicolumn{5}{c|}{SSIM $\uparrow$} & \multicolumn{5}{c}{LPIPS $\downarrow$}                                                                                                                                                                                                             \\
                                    & $\times 1$                           & $\times 1/2$                         & $\times 1/4$                           & $\times 1/8$   & Avg.
                                    & $\times 1$                           & $\times 1/2$                         & $\times 1/4$                           & $\times 1/8$   & Avg.
                                    & $\times 1$                           & $\times 1/2$                         & $\times 1/4$                           & $\times 1/8$   & Avg.                                                                                                                                                                                     \\
            \midrule
            3DGS
                                    & 19.89                                & 23.11                                & 29.67                                  & 23.90          & 24.14          & 0.575          & 0.734          & 0.903          & 0.809          & 0.755          & 0.462          & 0.279          & 0.105          & 0.148          & 0.248          \\
            Scaffold-GS
                                    & 20.50                                & 23.39                                & 29.78                                  & 23.65          & 24.33          & 0.569          & 0.723          & 0.899          & 0.807          & 0.750          & 0.470          & 0.289          & 0.109          & 0.145          & 0.253          \\
\bottomrule
        \end{tabular}
    }
\end{table*}

\begin{table*}[ht!]
    \centering
    \caption{Single-scale training ($\times 1/8$ resolution) and multi-scale testing results on the multi-scale Mip-NeRF 360 dataset. The best results are highlighted in \textbf{bold}.}
    \label{tab:stmt8_mipnerf360}
    \resizebox{\textwidth}{!}{
        \begin{tabular}{l|ccccc|ccccc|ccccc}
            \toprule
            \multirow{2}{*}{} & \multicolumn{5}{c|}{PSNR $\uparrow$} & \multicolumn{5}{c|}{SSIM $\uparrow$} & \multicolumn{5}{c}{LPIPS $\downarrow$}                                                                                                                                                                                                             \\
                                    & $\times 1$                           & $\times 1/2$                         & $\times 1/4$                           & $\times 1/8$   & Avg.
                                    & $\times 1$                           & $\times 1/2$                         & $\times 1/4$                           & $\times 1/8$   & Avg.
                                    & $\times 1$                           & $\times 1/2$                         & $\times 1/4$                           & $\times 1/8$   & Avg.                                                                                                                                                                                     \\
            \midrule
            3DGS
                                    & 17.25                                & 18.73                                & 22.41                                  & 30.60          & 22.25          & 0.452          & 0.516          & 0.726          & 0.930          & 0.656          & 0.566          & 0.463          & 0.245          & 0.071          & 0.336          \\
            Scaffold-GS
                                    & 17.09                                & 18.43                                & 21.98                                  & 31.06          & 22.14          & 0.443          & 0.490          & 0.705          & 0.932          & 0.643          & 0.576          & 0.477          & 0.262          & 0.068          & 0.346          \\
\bottomrule
        \end{tabular}
    }
\end{table*}

\begin{table*}[ht!]
    \centering
    \caption{Multi-scale training and multi-scale testing results on the multi-scale Mip-NeRF 360 dataset. The best results are highlighted in \textbf{bold}.}
    \label{tab:mtmt_mipnerf360}
    \resizebox{\textwidth}{!}{
        \begin{tabular}{l|ccccc|ccccc|ccccc}
            \toprule
            \multirow{2}{*}{} & \multicolumn{5}{c|}{PSNR $\uparrow$} & \multicolumn{5}{c|}{SSIM $\uparrow$} & \multicolumn{5}{c}{LPIPS $\downarrow$}                                                                                                                                                                                                             \\
                                    & $\times 1$                           & $\times 1/2$                         & $\times 1/4$                           & $\times 1/8$   & Avg.
                                    & $\times 1$                           & $\times 1/2$                         & $\times 1/4$                           & $\times 1/8$   & Avg.
                                    & $\times 1$                           & $\times 1/2$                         & $\times 1/4$                           & $\times 1/8$   & Avg.                                                                                                                                                                                     \\
            \midrule
            3DGS
                                    & 26.85                                & 27.89                                & 28.28                                  & 27.17          & 27.55          & 0.790          & 0.847          & 0.882          & 0.876          & 0.849          & 0.311          & 0.205          & 0.132          & 0.111          & 0.190          \\
            Scaffold-GS
                                    & 26.61                                & 27.75                                & 28.42                                  & 27.23          & 27.50          & 0.773          & 0.835          & 0.879          & 0.875          & 0.841          & 0.333          & 0.223          & 0.136          & 0.112          & 0.201          \\
\bottomrule
        \end{tabular}
    }
\end{table*}

\begin{figure*}[ht!]
    \centering
    \setlength{\tabcolsep}{2pt}
    \begin{tabular}{ccccccc}
                                                                                                      & {Ground Truth} & {Tri-MipRF} & {3DGS} & {Mip-Splatting} & {Analytic-Splatting} & {Gaussian Blending} \\

        \raisebox{0.6em}{\rotatebox[origin=c]{90}{\textbf{1}}}                                              &
        \includegraphics[width=0.155\textwidth]{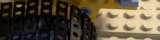}                 &
        \includegraphics[width=0.155\textwidth]{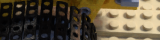}          &
        \includegraphics[width=0.155\textwidth]{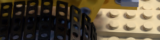}               &
        \includegraphics[width=0.155\textwidth]{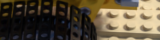}      &
        \includegraphics[width=0.155\textwidth]{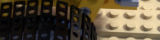} &
        \includegraphics[width=0.155\textwidth]{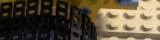} \\ [-2.5pt]
        \raisebox{0.6em}{\rotatebox[origin=c]{90}{1/2}} &
        \includegraphics[width=0.155\textwidth]{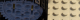} &
        \includegraphics[width=0.155\textwidth]{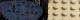} &
        \includegraphics[width=0.155\textwidth]{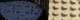} &
        \includegraphics[width=0.155\textwidth]{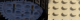} &
        \includegraphics[width=0.155\textwidth]{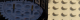} &
        \includegraphics[width=0.155\textwidth]{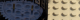} \\
\end{tabular}
    \caption{Qualitative results of single-scale training ($\times 1$ resolution) and multi-scale testing setting on the multi-scale Blender dataset. The training resolution is highlighted in \textbf{bold}.}
    \label{fig:qualitative_blender_zoomout}
\end{figure*}

\begin{figure*}[ht!]
    \centering
    \setlength{\tabcolsep}{2pt}
    \begin{tabular}{cccccccc}
                                                                                                     & {Ground Truth} & {Tri-MipRF} & {3DGS} & {Mip-Splatting} & {Analytic-Splatting} & {Gaussian Blending} \\

        \raisebox{0.6em}{\rotatebox[origin=c]{90}{1}}                                             &
        \includegraphics[width=0.155\textwidth]{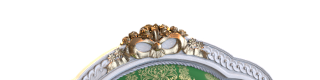}                 &
        \includegraphics[width=0.155\textwidth]{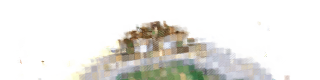}          &
        \includegraphics[width=0.155\textwidth]{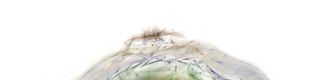}               &
        \includegraphics[width=0.155\textwidth]{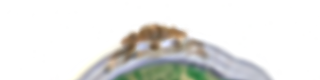}      &
        \includegraphics[width=0.155\textwidth]{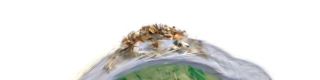} &
        \includegraphics[width=0.155\textwidth]{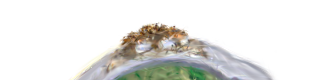} \\ [-2.5pt]
        \raisebox{0.6em}{\rotatebox[origin=c]{90}{1/2}} &
        \includegraphics[width=0.155\textwidth]{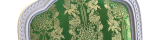} &
        \includegraphics[width=0.155\textwidth]{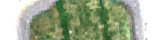} &
        \includegraphics[width=0.155\textwidth]{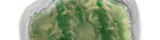} &
        \includegraphics[width=0.155\textwidth]{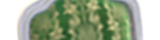} &
        \includegraphics[width=0.155\textwidth]{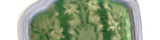} &
        \includegraphics[width=0.155\textwidth]{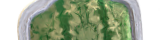} \\
\end{tabular}
    \caption{Qualitative results of single-scale training ($\times 1/8$ resolution) and multi-scale testing setting on the multi-scale Blender dataset. The training resolution is highlighted in \textbf{bold}.}
    \label{fig:qualitative_blender_zoomin}
\end{figure*}

\begin{figure*}[ht!]
    \centering
    \setlength{\tabcolsep}{2pt}
    \begin{tabular}{cccccc}
        {Ground Truth} & {3DGS} & {3DGS-SS} & {Mip-Splatting} & {Analytic-Splatting} & {Gaussian Blending} \\

        \includegraphics[width=0.155\textwidth]{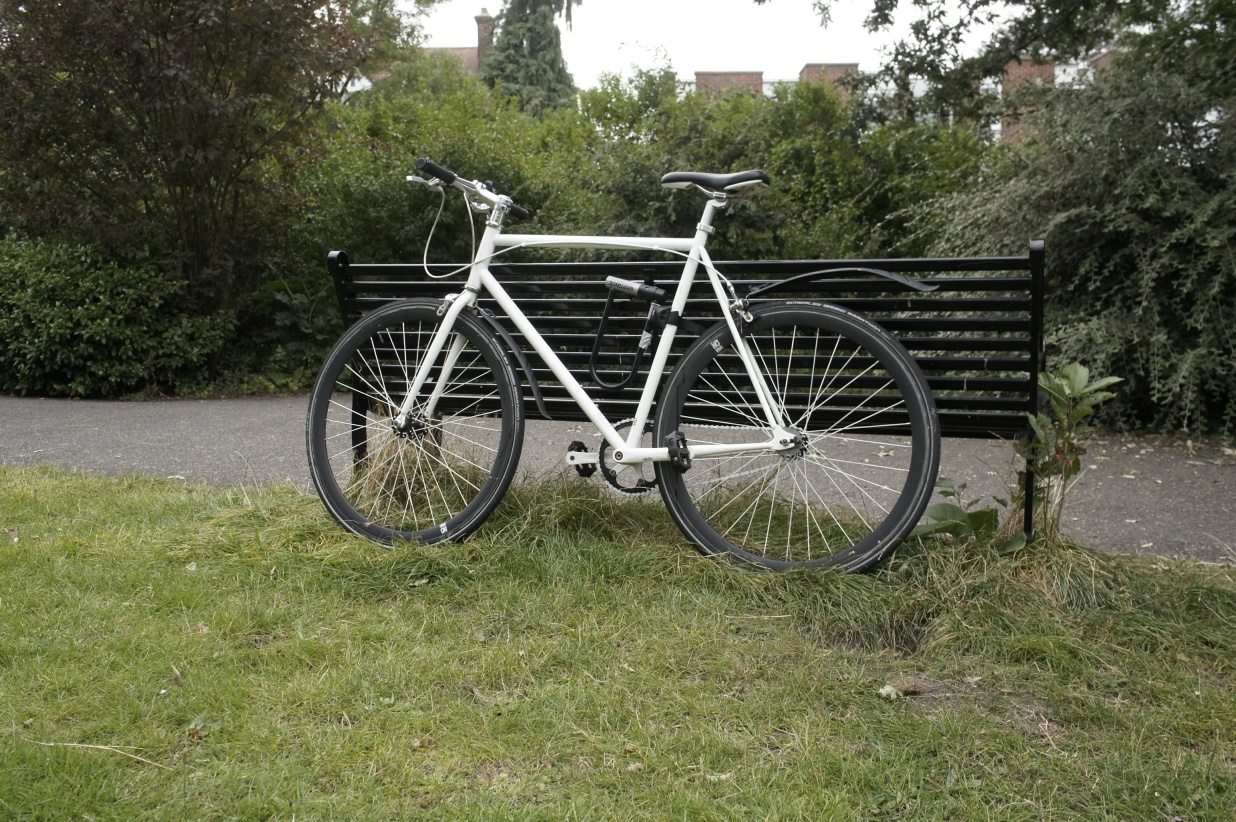}                 &
        \includegraphics[width=0.155\textwidth]{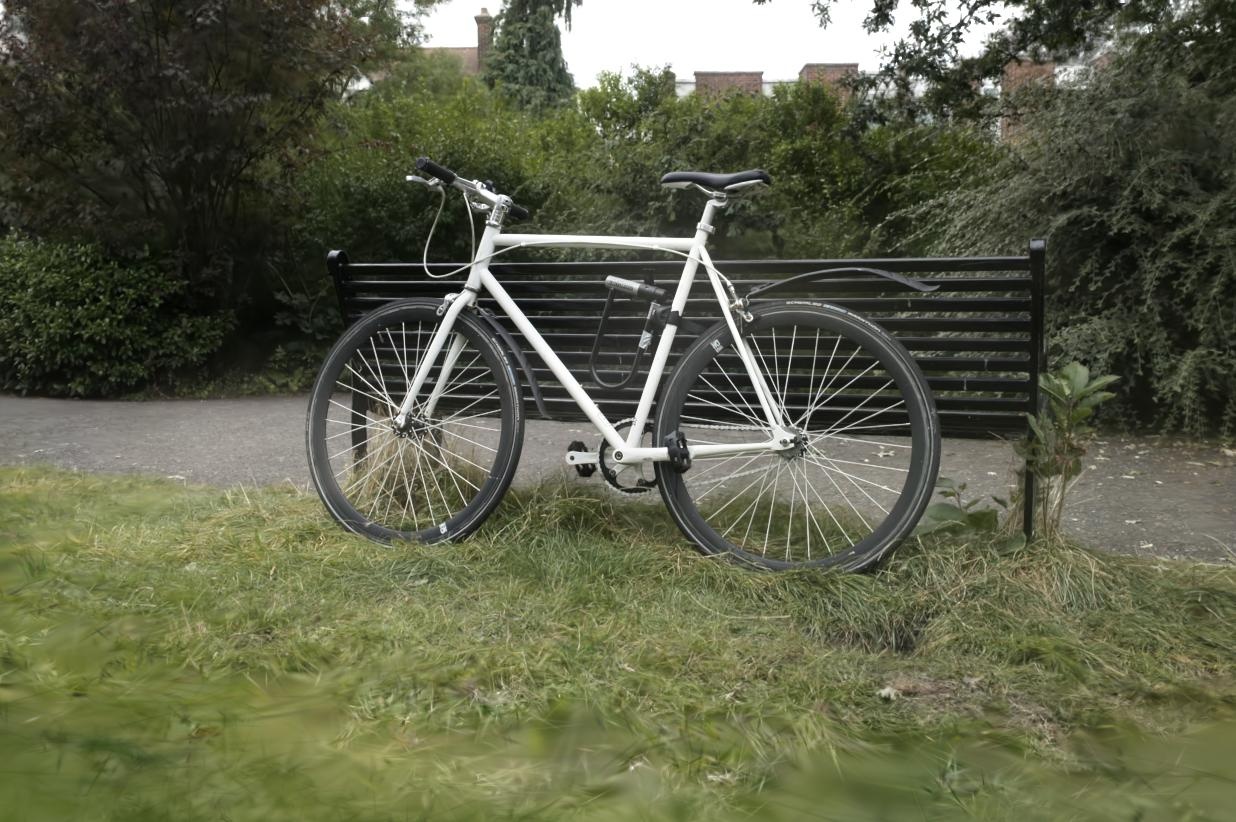}               &
        \includegraphics[width=0.155\textwidth]{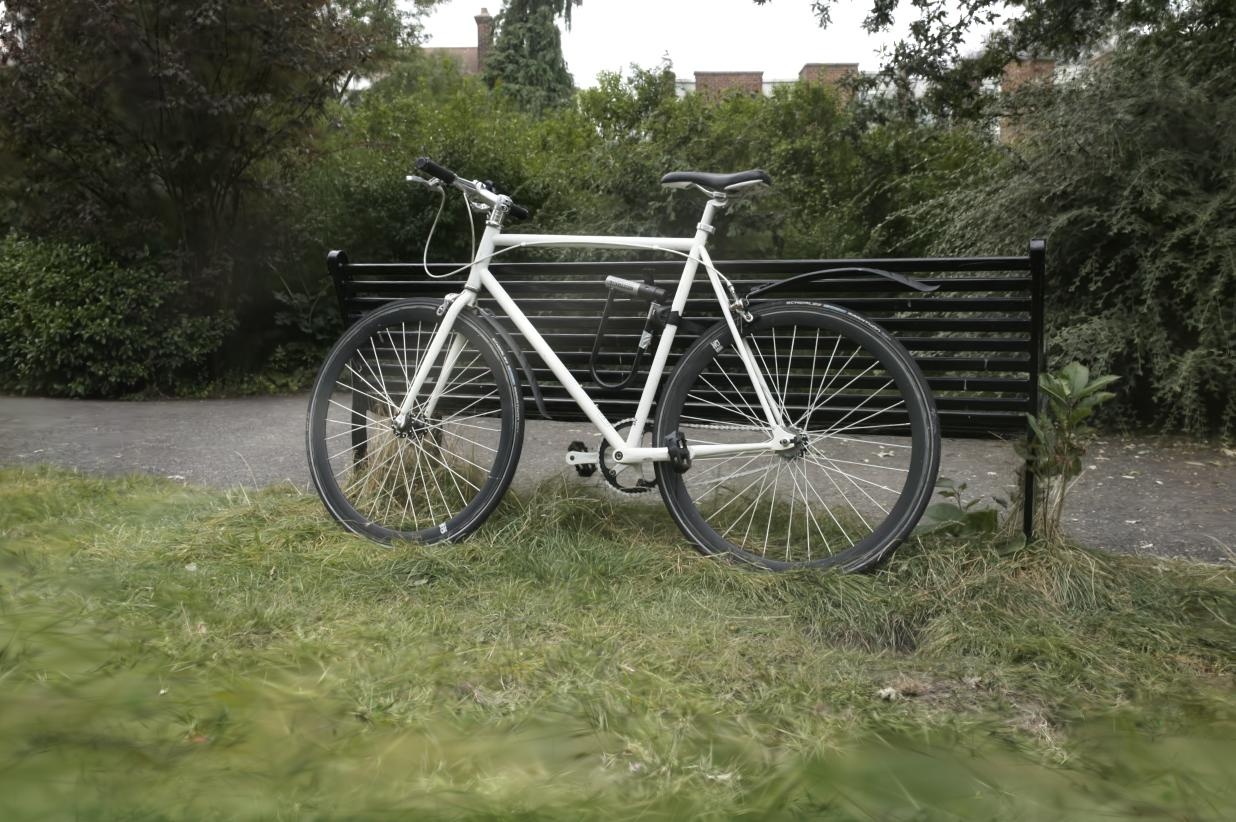}            &
        \includegraphics[width=0.155\textwidth]{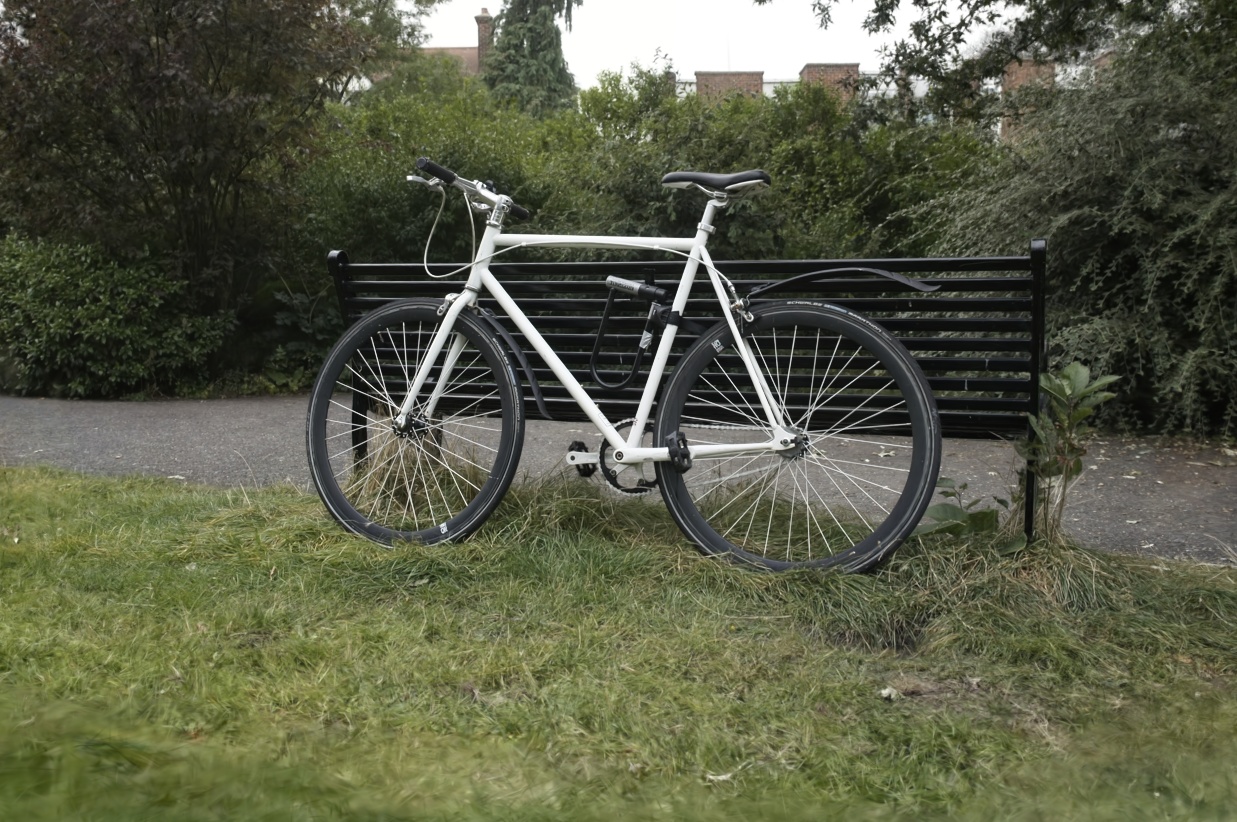}      &
        \includegraphics[width=0.155\textwidth]{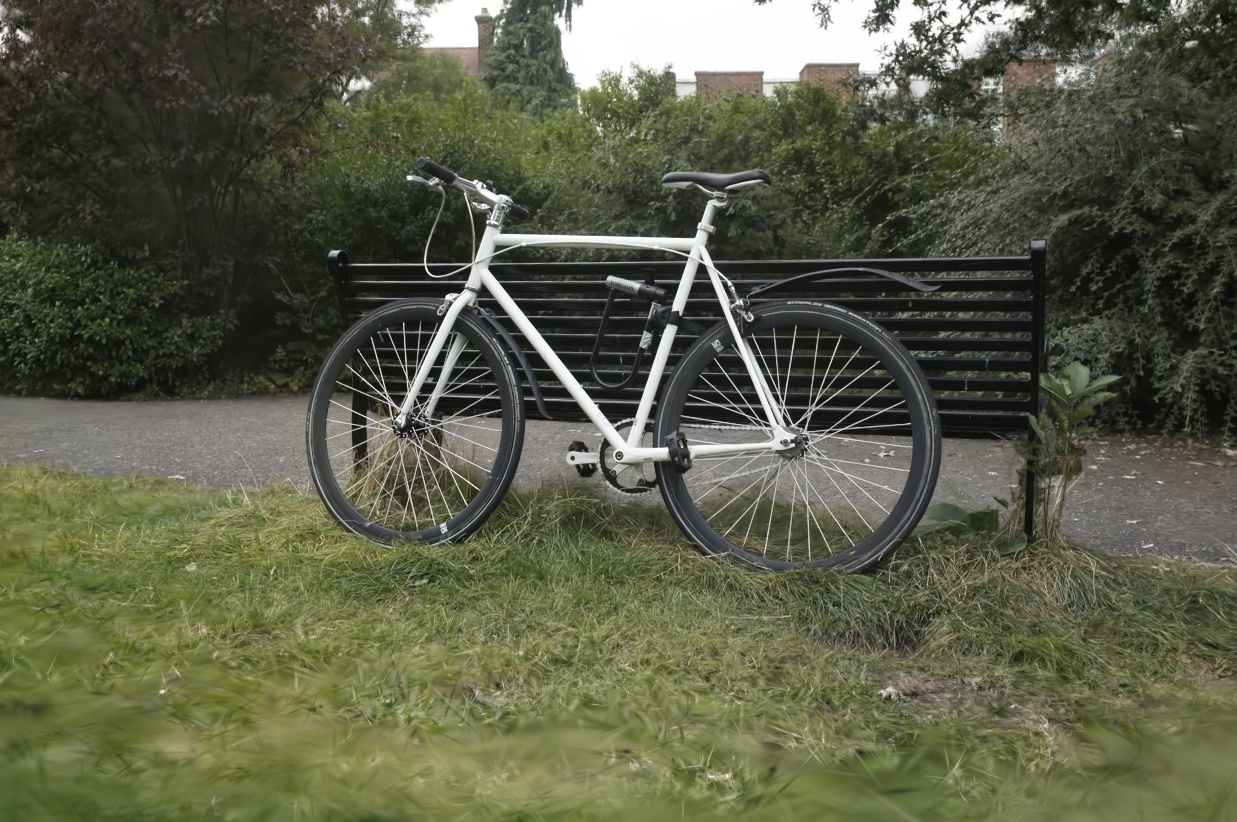} &
        \includegraphics[width=0.155\textwidth]{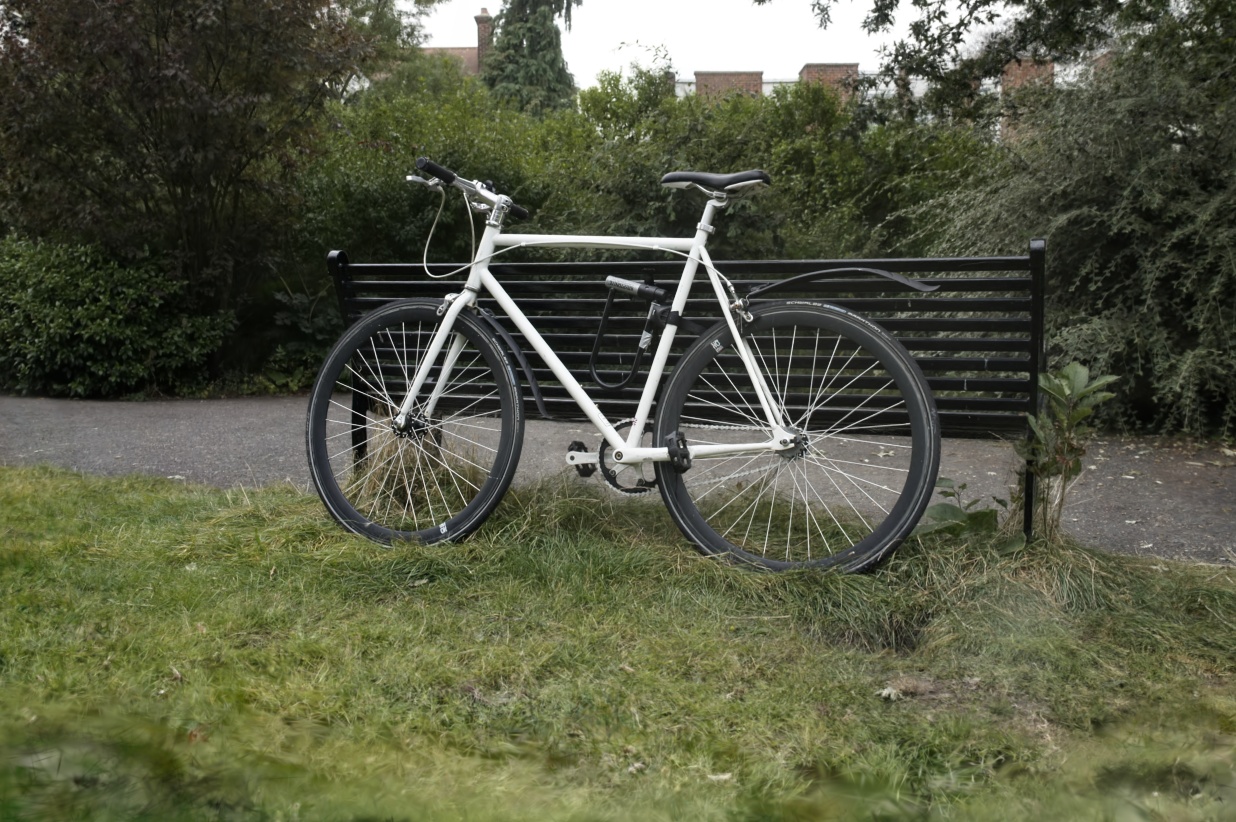}         \\

        \includegraphics[width=0.155\textwidth]{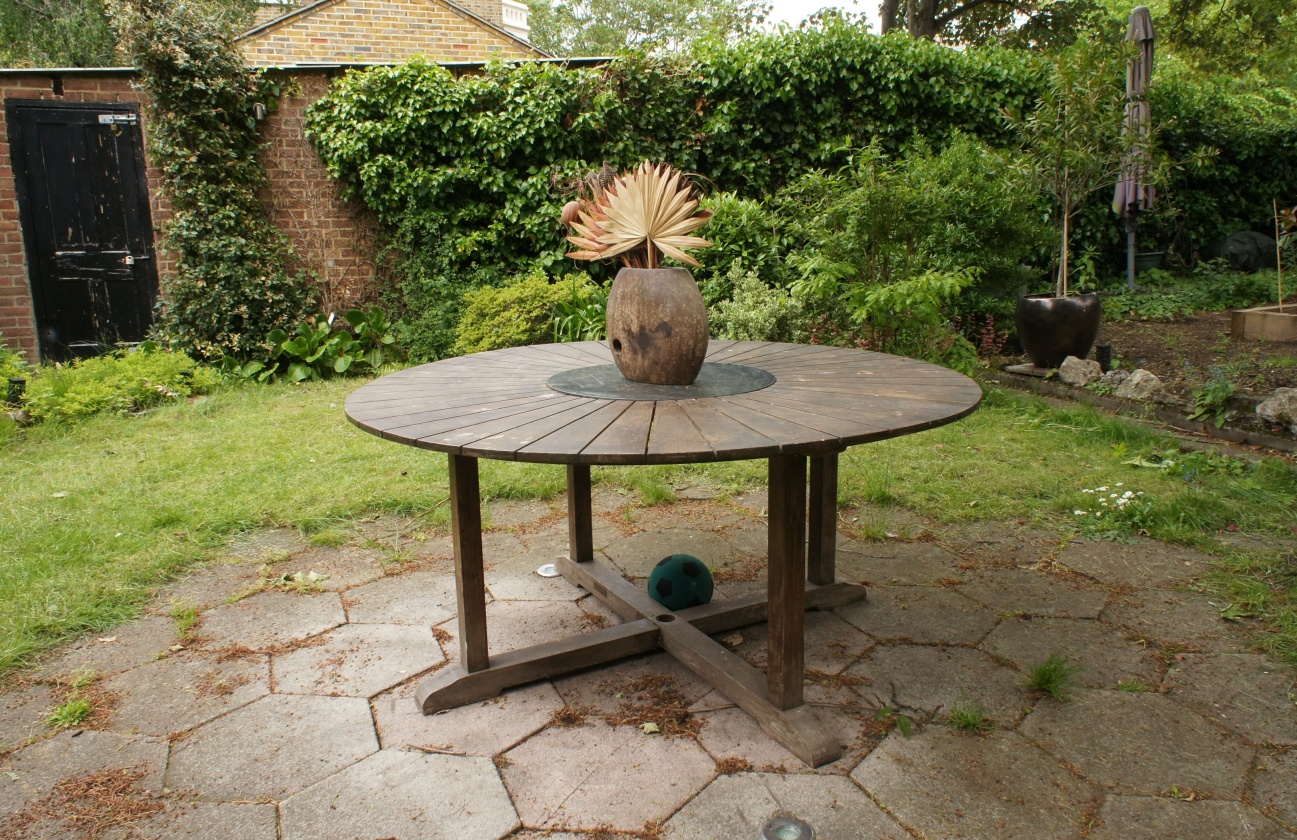}                 &
        \includegraphics[width=0.155\textwidth]{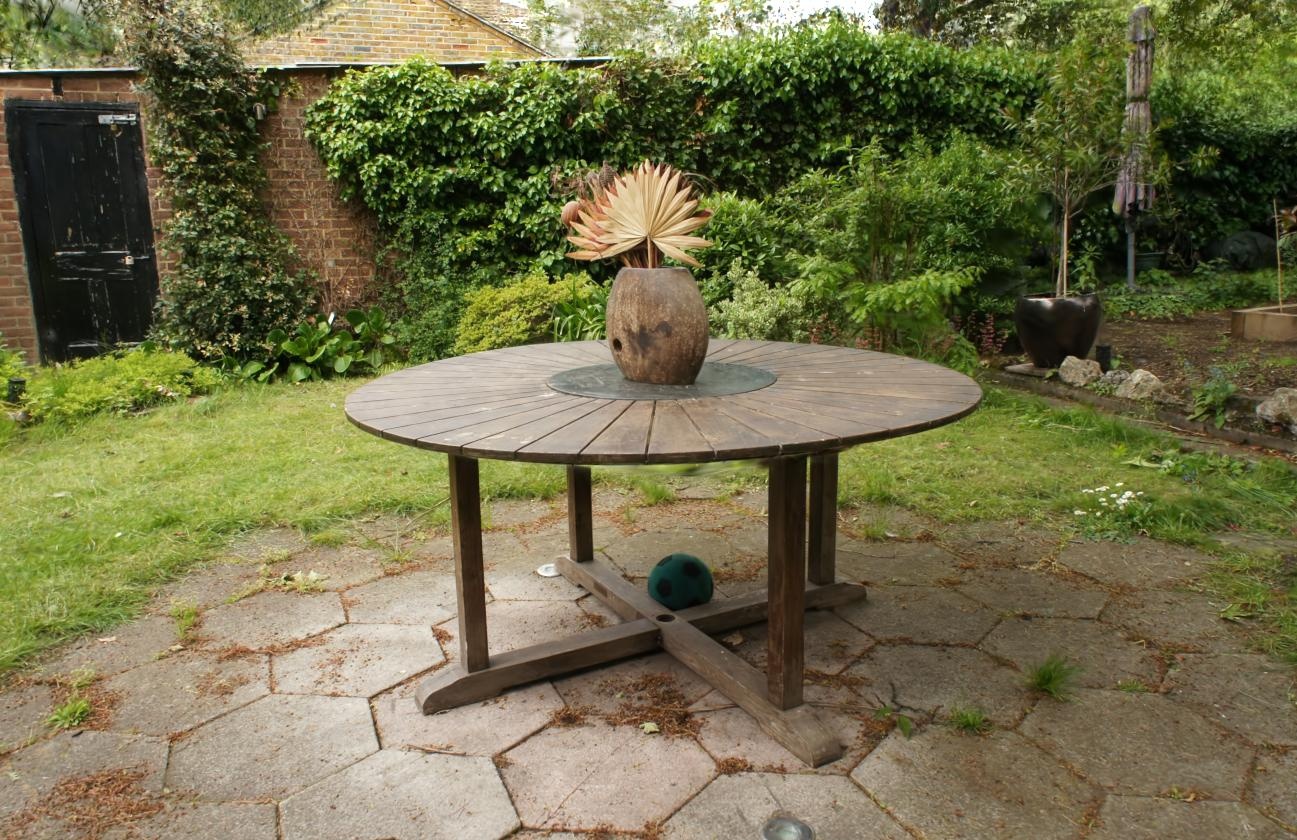}               &
        \includegraphics[width=0.155\textwidth]{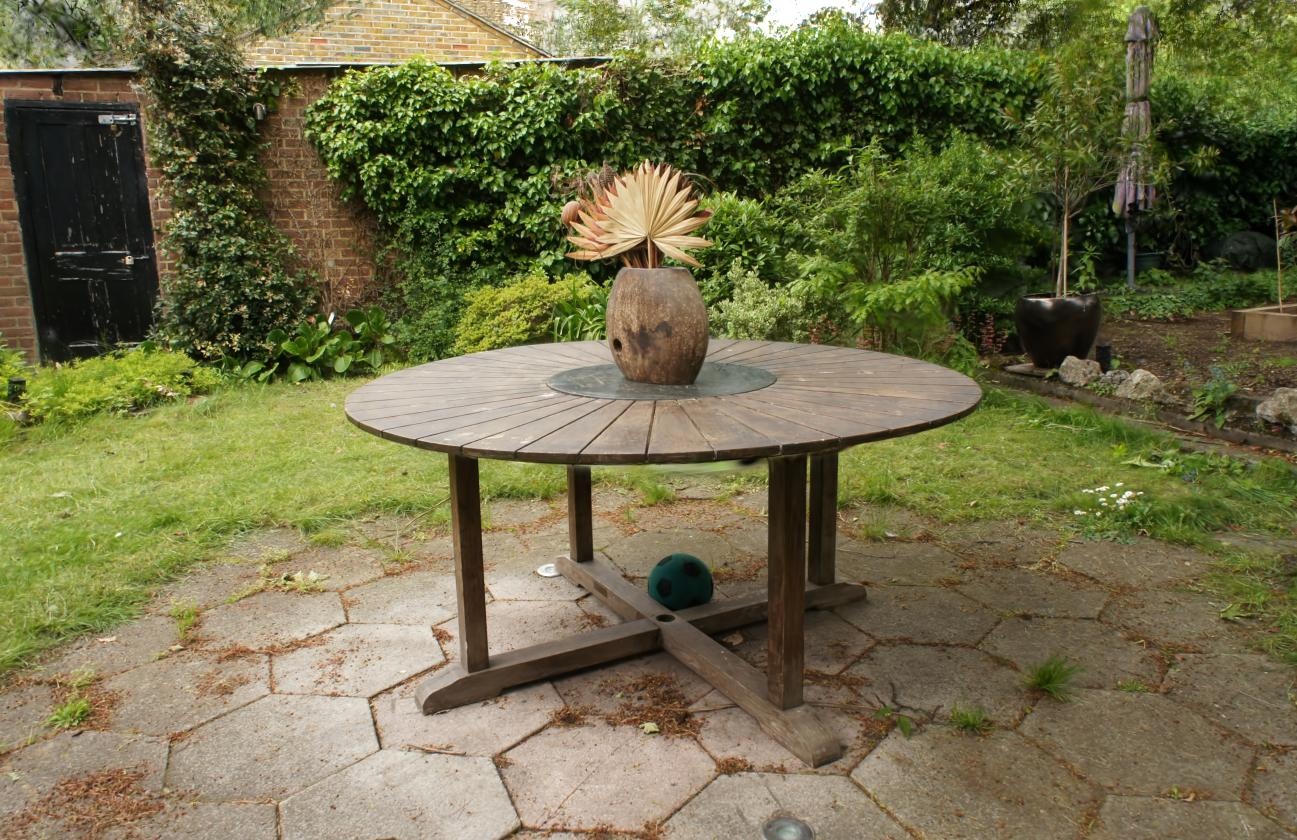}            &
        \includegraphics[width=0.155\textwidth]{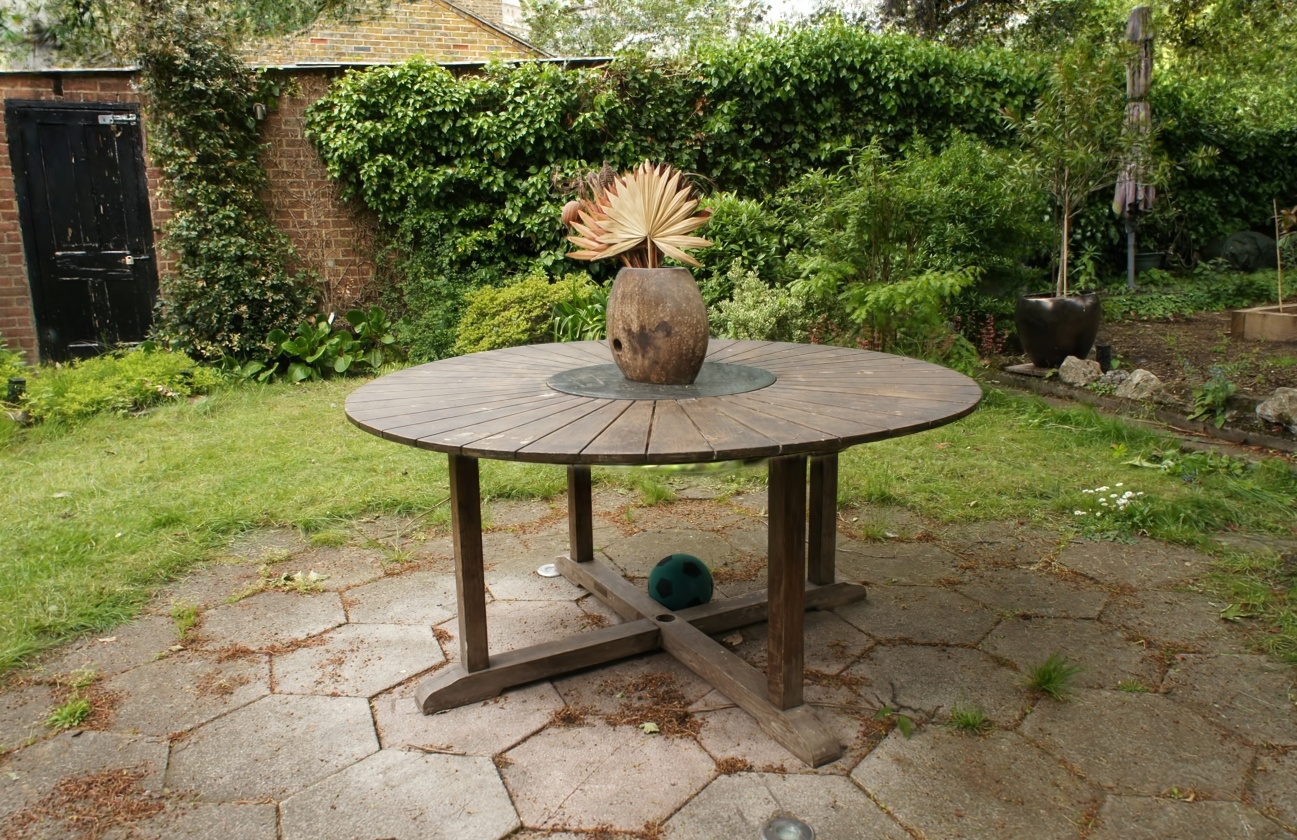}      &
        \includegraphics[width=0.155\textwidth]{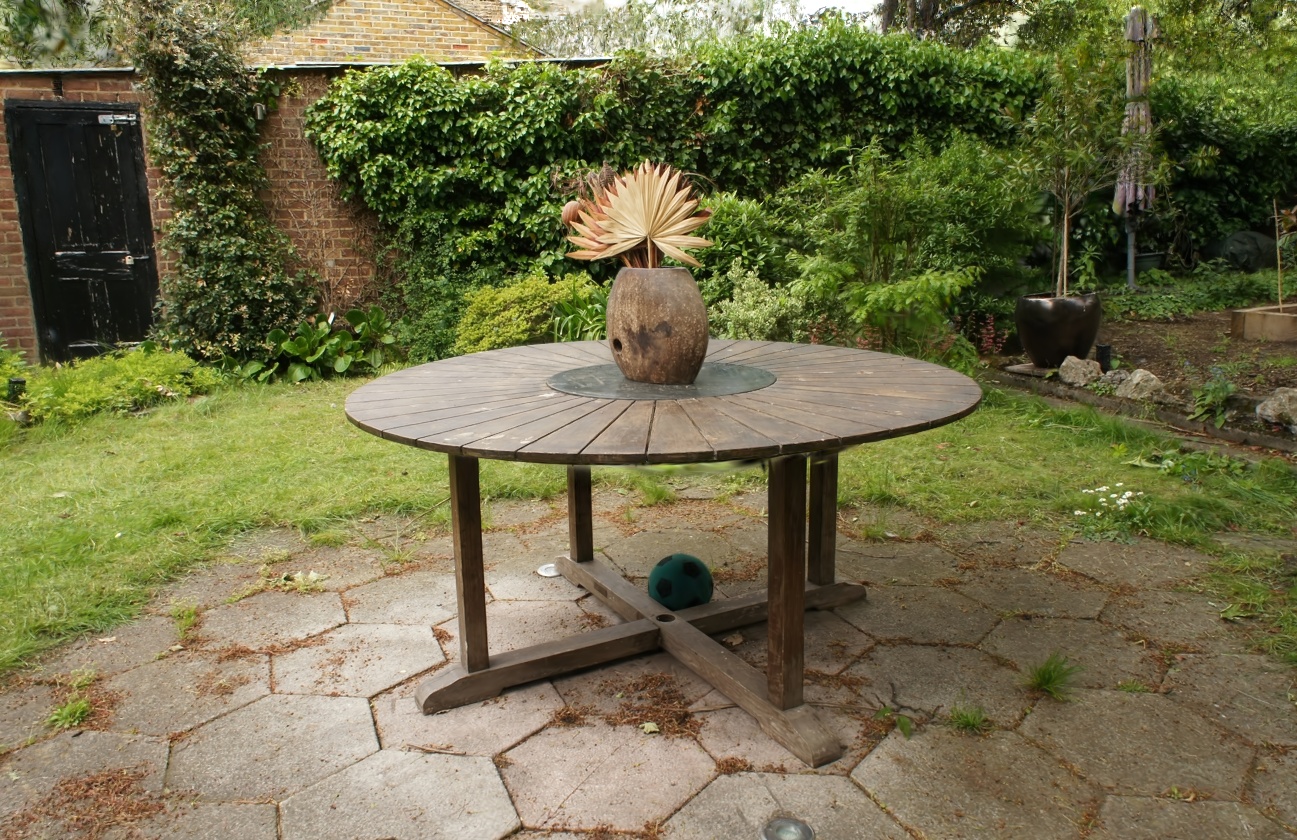} &
        \includegraphics[width=0.155\textwidth]{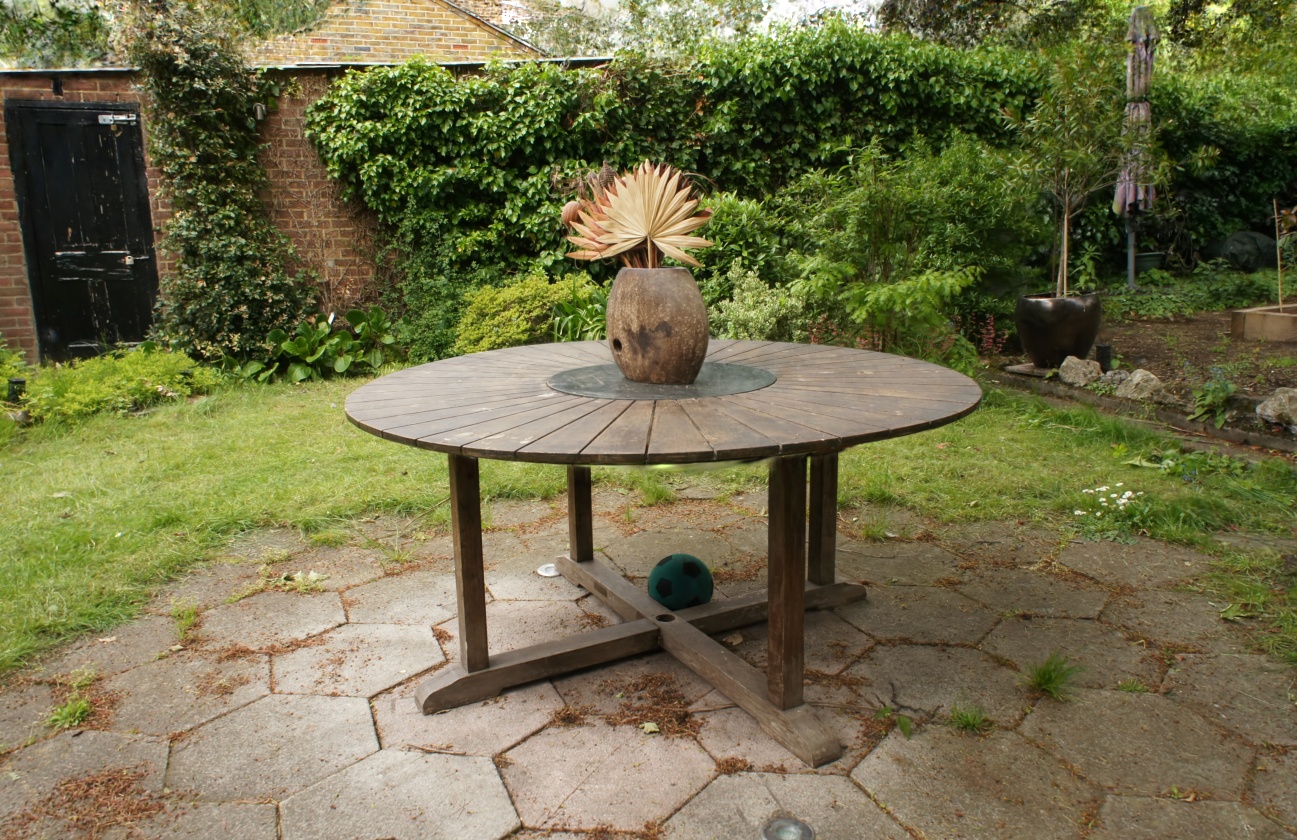}         \\
\end{tabular}
    \caption{Qualitative results of Novel View Synthesis (NVS) setting on the multi-scale Mip-NeRF 360 dataset.}
    \label{fig:qualitative_nvs}
\end{figure*}

\begin{figure*}[ht!]
    \centering
    \setlength{\tabcolsep}{2pt}
    \begin{tabular}{cccccccc}
        & {Ground Truth} & {Tri-MipRF} & {3DGS} & {Mip-Splatting} & {Analytic-Splatting} & {Gaussian Blending} \\

        \raisebox{3.5em}{\rotatebox[origin=c]{90}{$1 \rightarrow 1/2$}}                              &
        \includegraphics[width=0.155\textwidth]{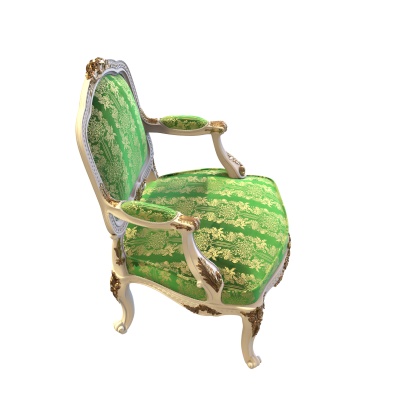}                 &
        \includegraphics[width=0.155\textwidth]{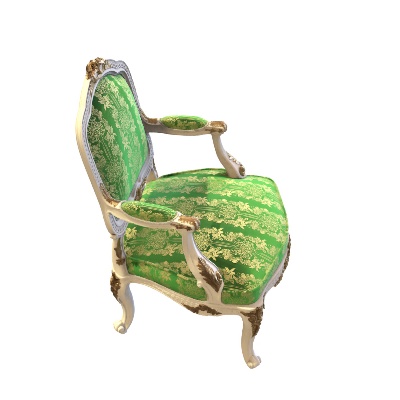}          &
        \includegraphics[width=0.155\textwidth]{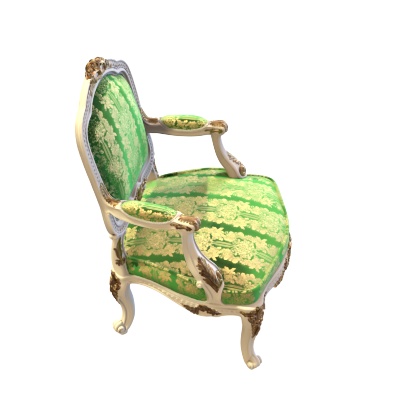}               &
        \includegraphics[width=0.155\textwidth]{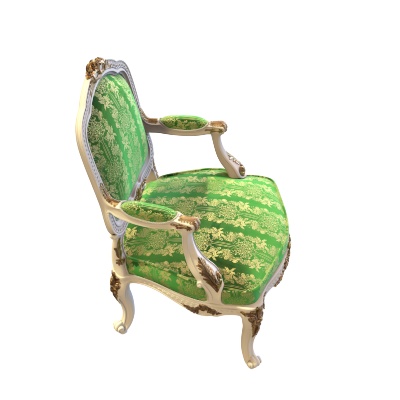}      &
        \includegraphics[width=0.155\textwidth]{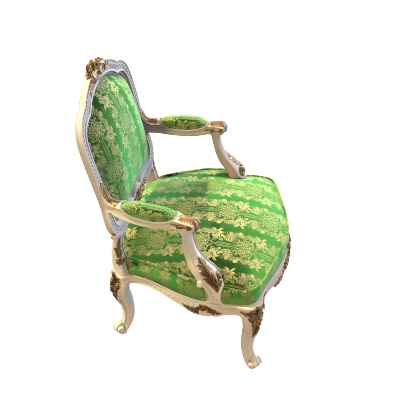} &
        \includegraphics[width=0.155\textwidth]{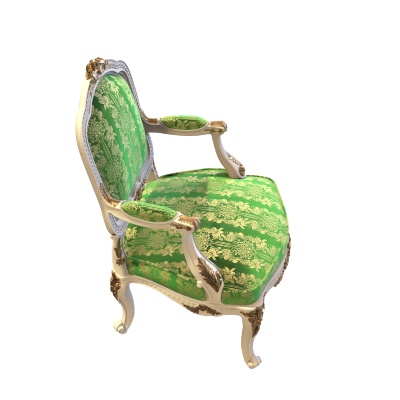}         \\

        \raisebox{3.5em}{\rotatebox[origin=c]{90}{$1 \rightarrow 1/4$}}                              &
        \includegraphics[width=0.155\textwidth]{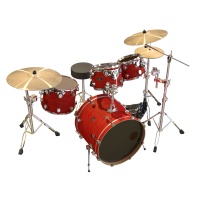}                 &
        \includegraphics[width=0.155\textwidth]{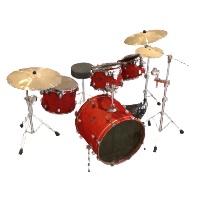}          &
        \includegraphics[width=0.155\textwidth]{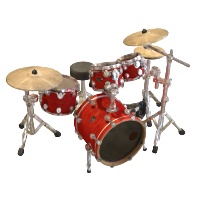}               &
        \includegraphics[width=0.155\textwidth]{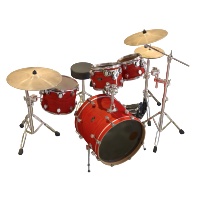}      &
        \includegraphics[width=0.155\textwidth]{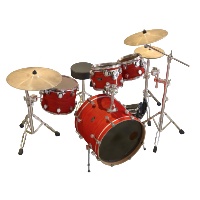} &
        \includegraphics[width=0.155\textwidth]{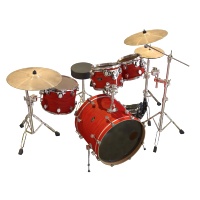}         \\
\end{tabular}
    \caption{Qualitative results of zoom-out setting on the multi-scale Blender dataset. The training resolution and rendering resolution is represented as (training resolution) $\rightarrow$ (rendering resolution).}
    \label{fig:qualitative_zoomout}
\end{figure*}

\begin{figure*}[ht!]
    \centering
    \setlength{\tabcolsep}{2pt}
    \begin{tabular}{cccccccc}
        & {Ground Truth} & {3DGS} & {3DGS-SS} & {Mip-Splatting} & {Analytic-Splatting} & {Gaussian Blending} \\

        \raisebox{2.5em}{\rotatebox[origin=c]{90}{$1/8 \rightarrow 1/4$}}                                &
        \includegraphics[width=0.155\textwidth]{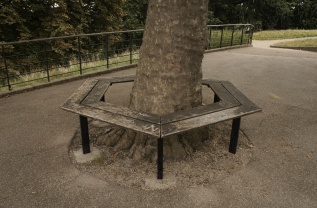}                 &
        \includegraphics[width=0.155\textwidth]{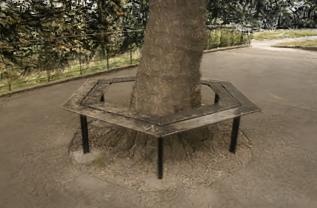}               &
        \includegraphics[width=0.155\textwidth]{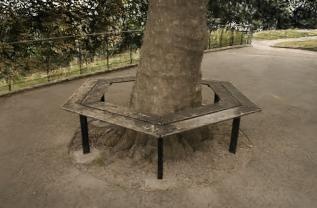}            &
        \includegraphics[width=0.155\textwidth]{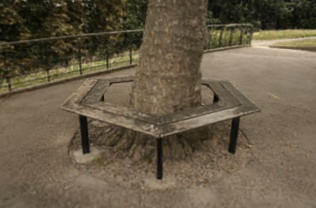}      &
        \includegraphics[width=0.155\textwidth]{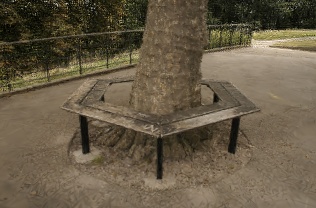} &
        \includegraphics[width=0.155\textwidth]{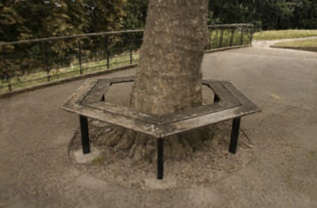}         \\

        \raisebox{2.5em}{\rotatebox[origin=c]{90}{$1/8 \rightarrow 1/2$}}                                &
        \includegraphics[width=0.155\textwidth]{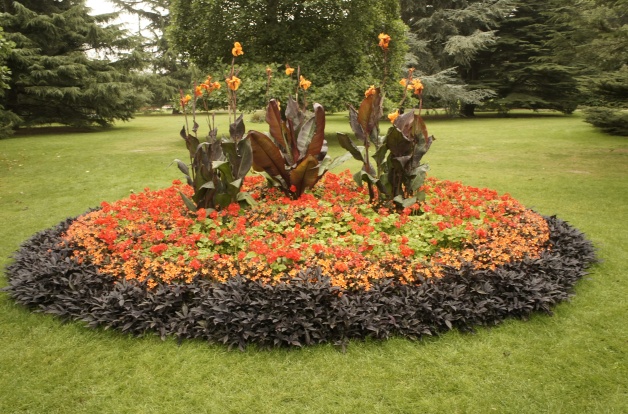}                 &
        \includegraphics[width=0.155\textwidth]{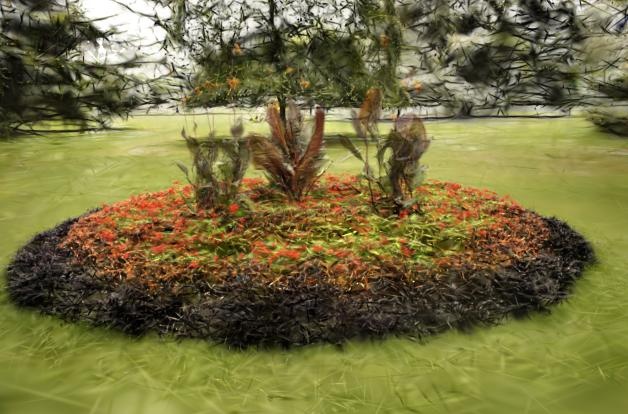}               &
        \includegraphics[width=0.155\textwidth]{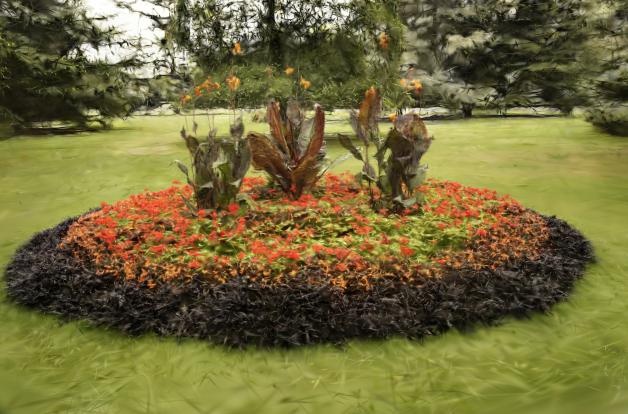}            &
        \includegraphics[width=0.155\textwidth]{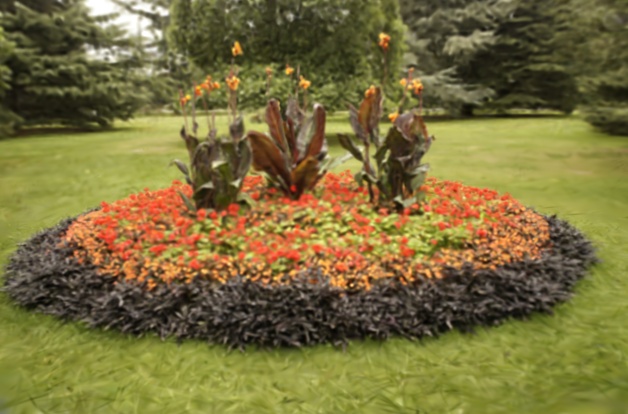}      &
        \includegraphics[width=0.155\textwidth]{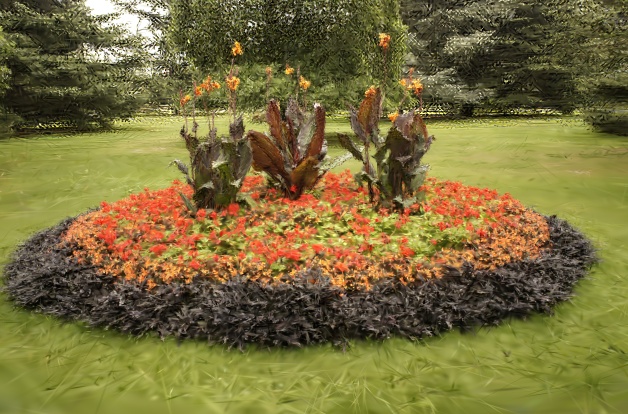} &
        \includegraphics[width=0.155\textwidth]{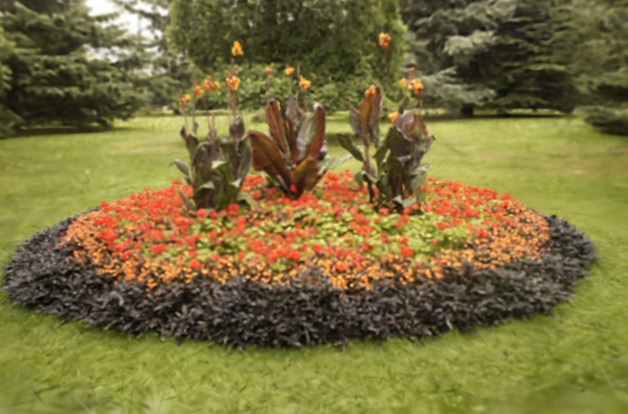}         \\
\end{tabular}
    \caption{Qualitative results of zoom-in setting on the multi-scale Mip-NeRF 360 dataset. The training resolution and rendering resolution is represented as (training resolution) $\rightarrow$ (rendering resolution).}
    \label{fig:qualitative_zoomin}
\end{figure*}

\begin{figure*}[ht!]
    \centering
    \setlength{\tabcolsep}{2pt}
    \begin{tabular}{ccccccc}
        {Ground Truth} & {3DGS} & {3DGS-SS} & {Mip-Splatting} & {Analytic-Splatting} & {Gaussian Blending} \\

        \includegraphics[width=0.155\textwidth]{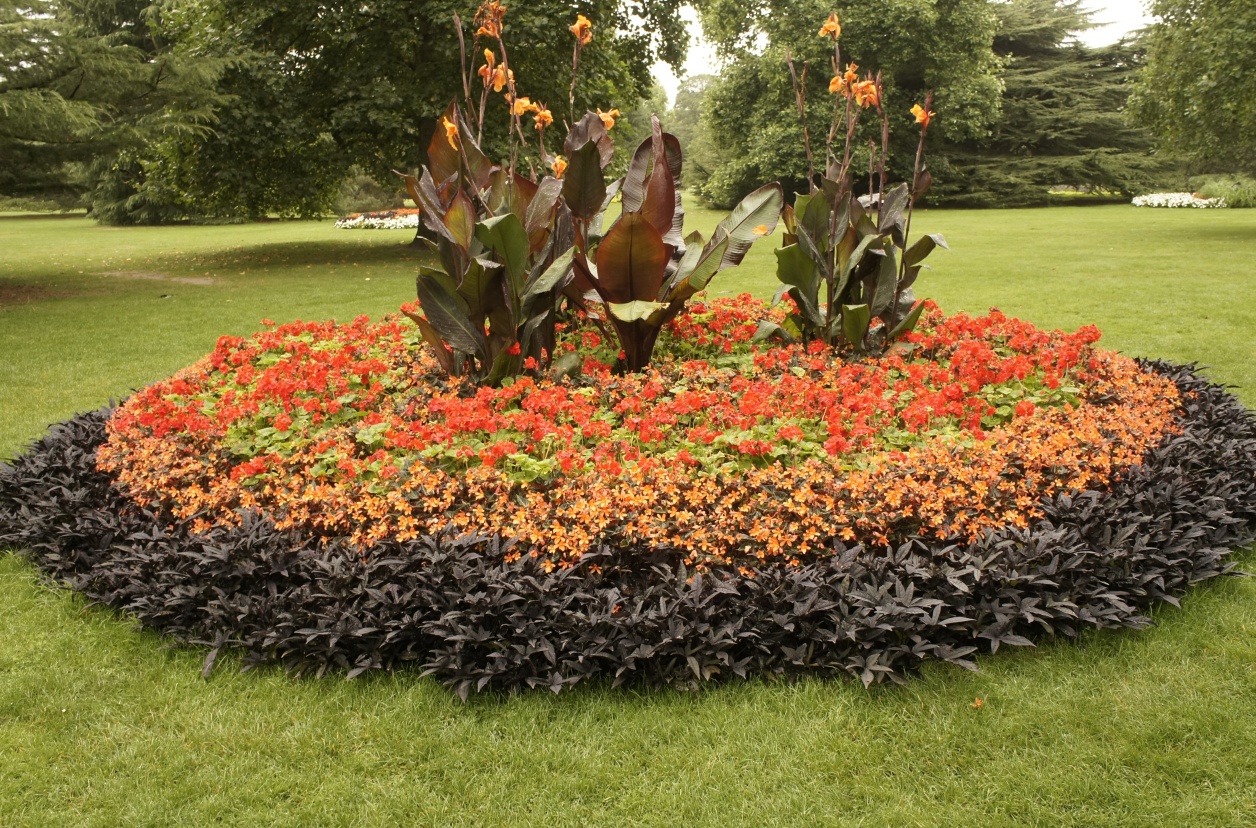}                 &
        \includegraphics[width=0.155\textwidth]{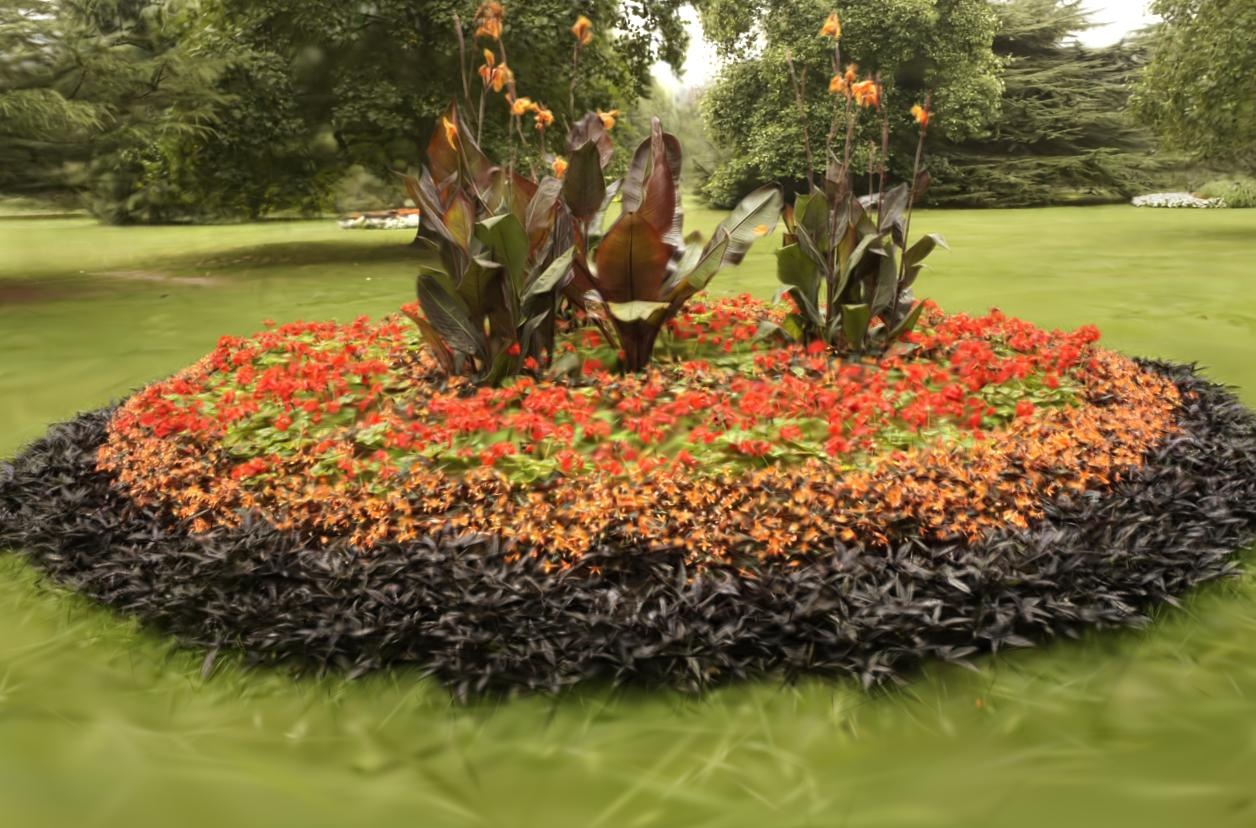}               &
        \includegraphics[width=0.155\textwidth]{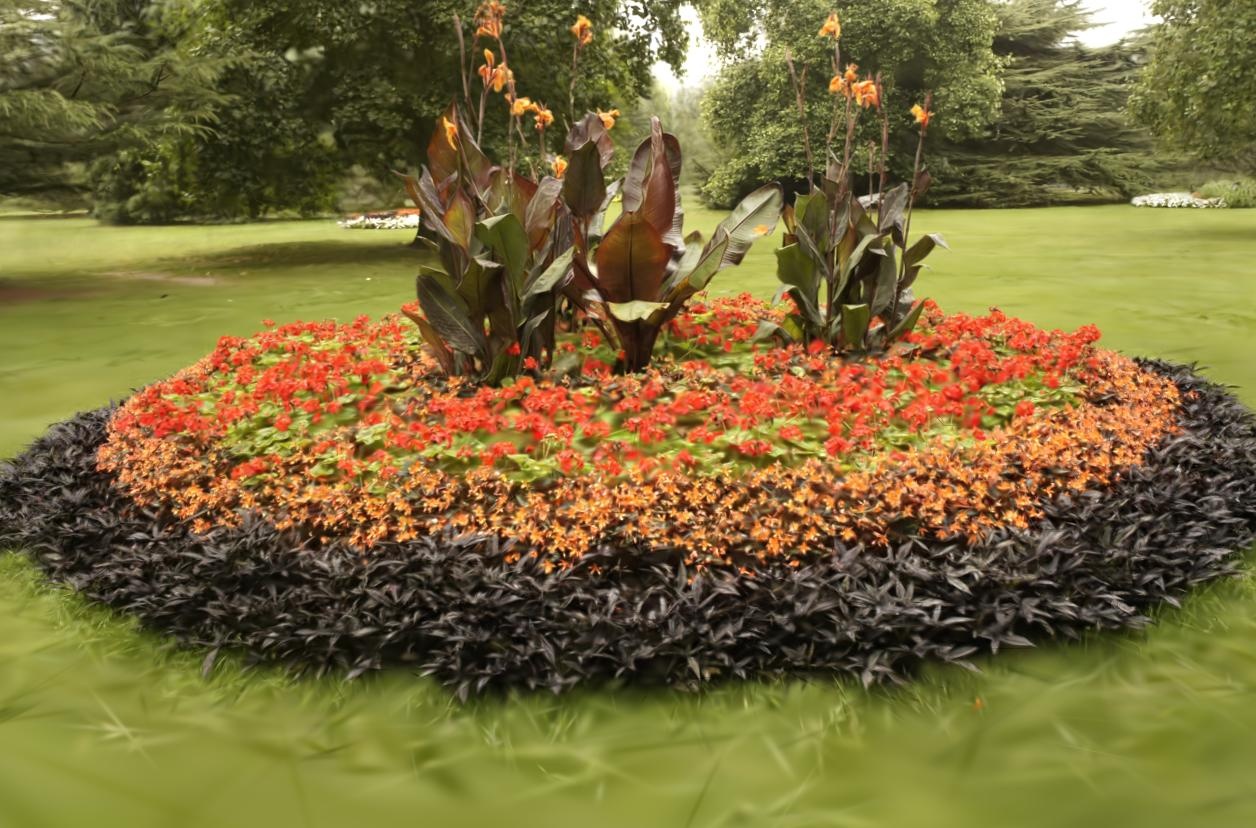}            &
        \includegraphics[width=0.155\textwidth]{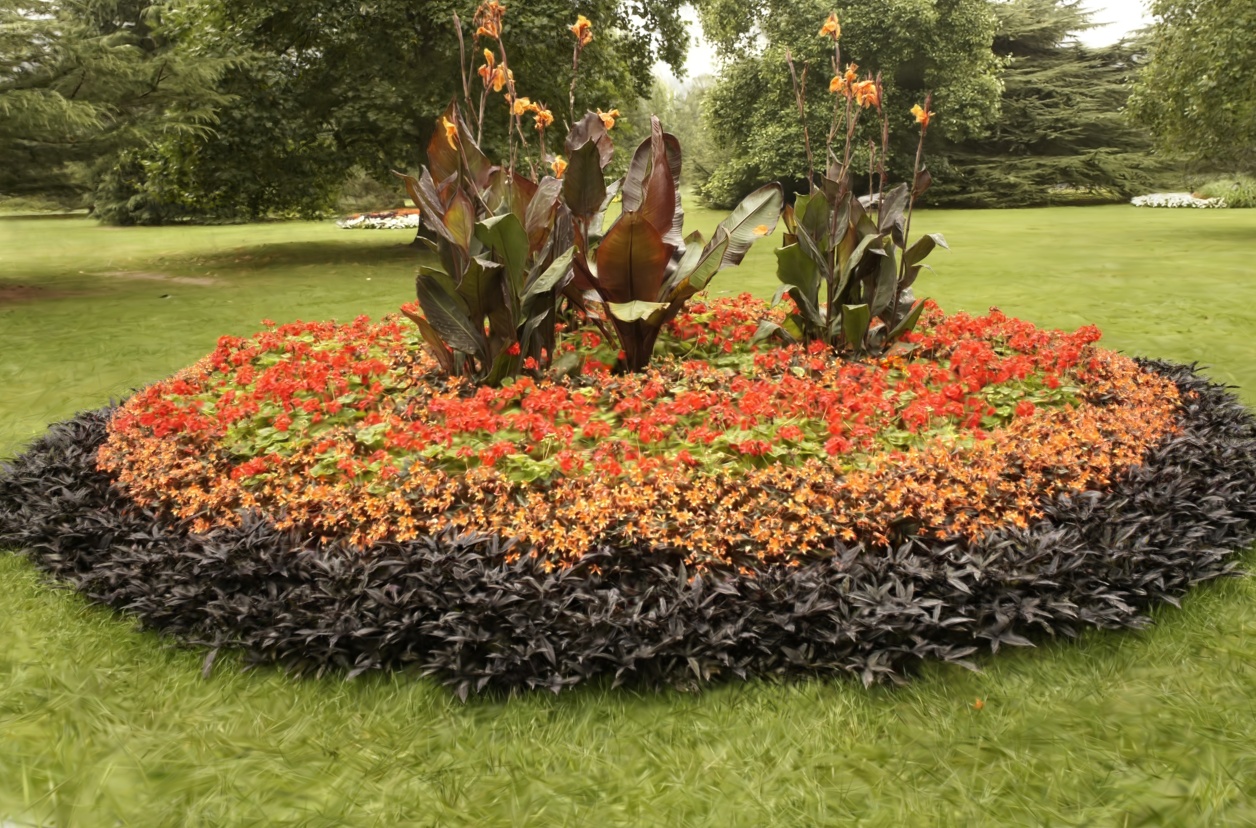}      &
        \includegraphics[width=0.155\textwidth]{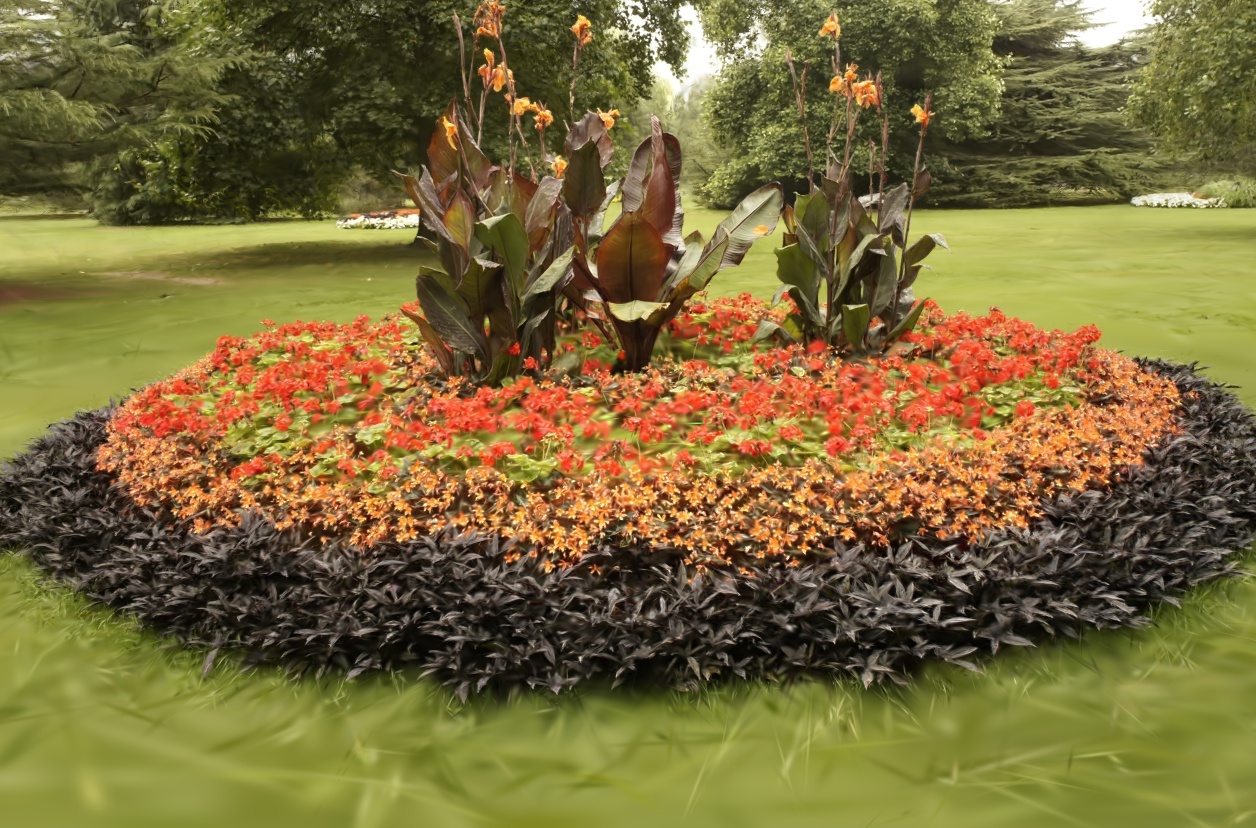} &
        \includegraphics[width=0.155\textwidth]{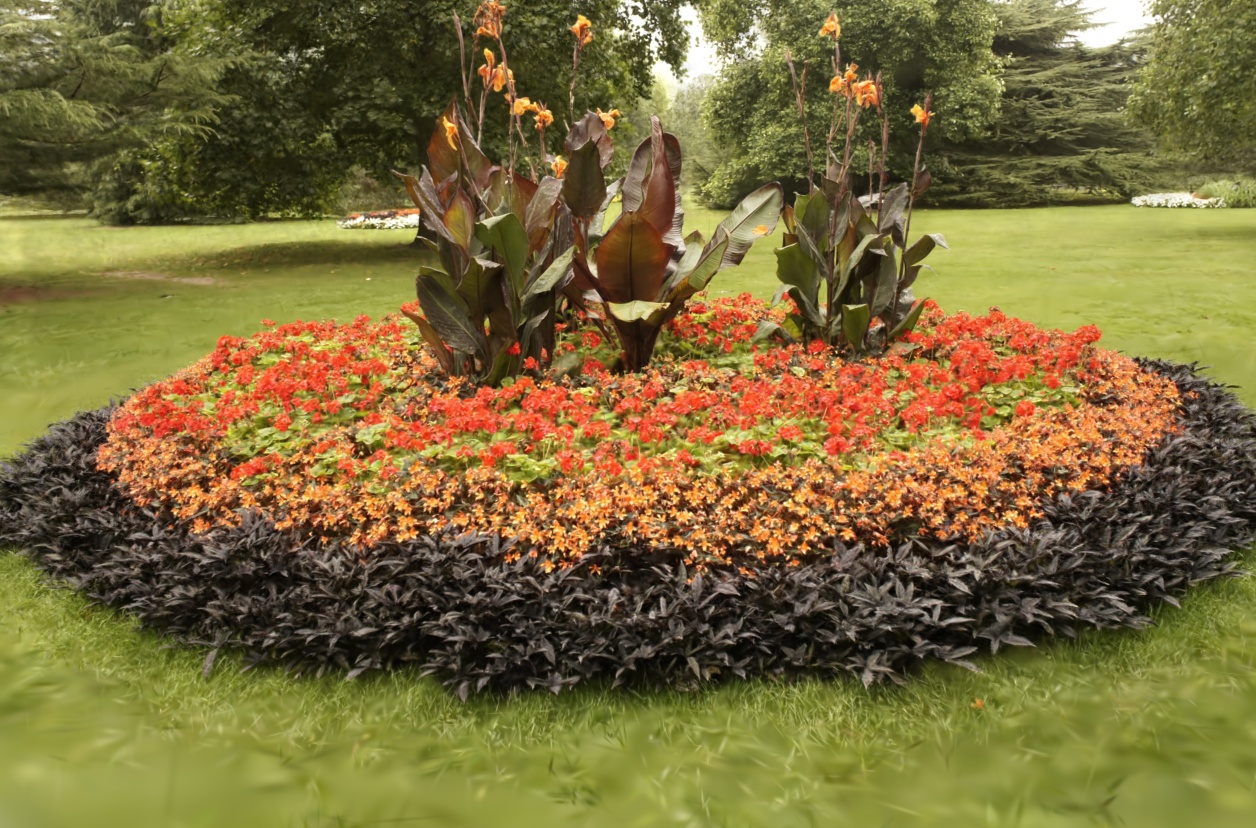}         \\

        \includegraphics[width=0.155\textwidth]{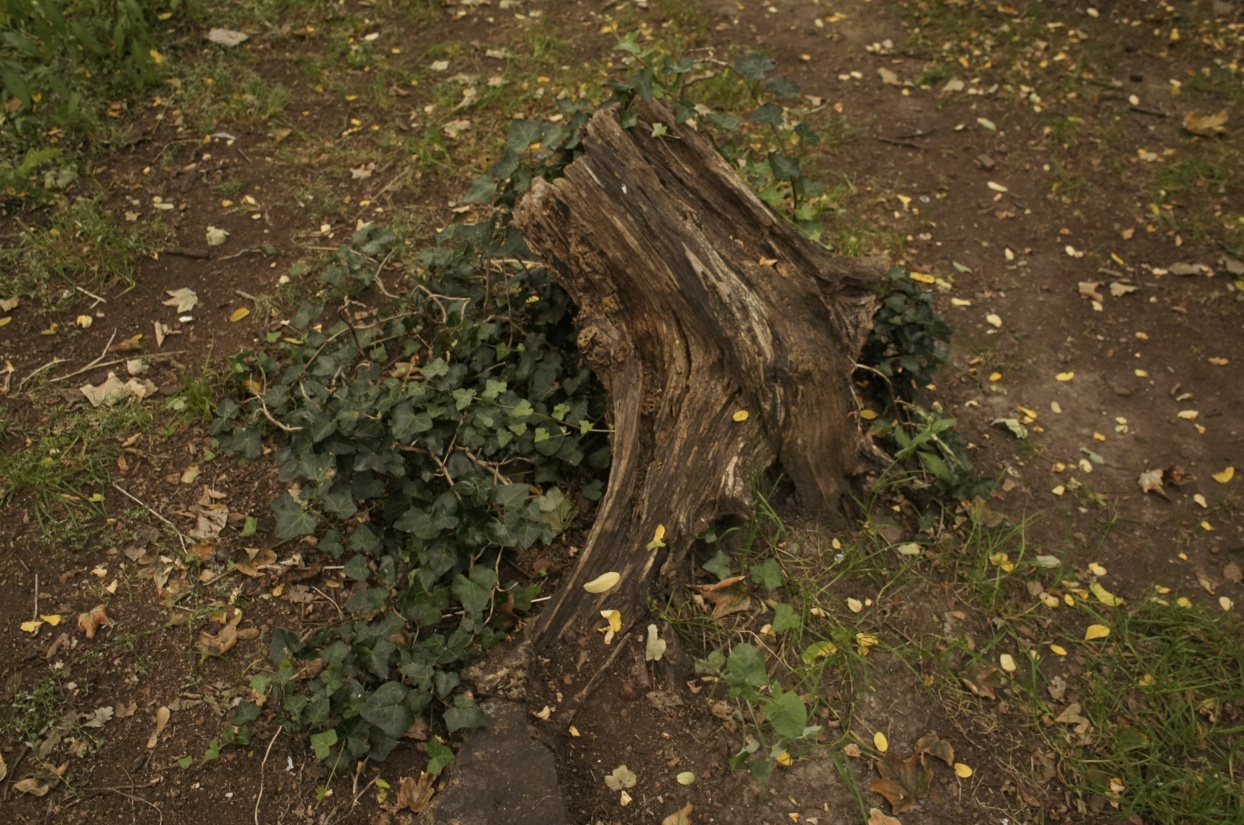}                 &
        \includegraphics[width=0.155\textwidth]{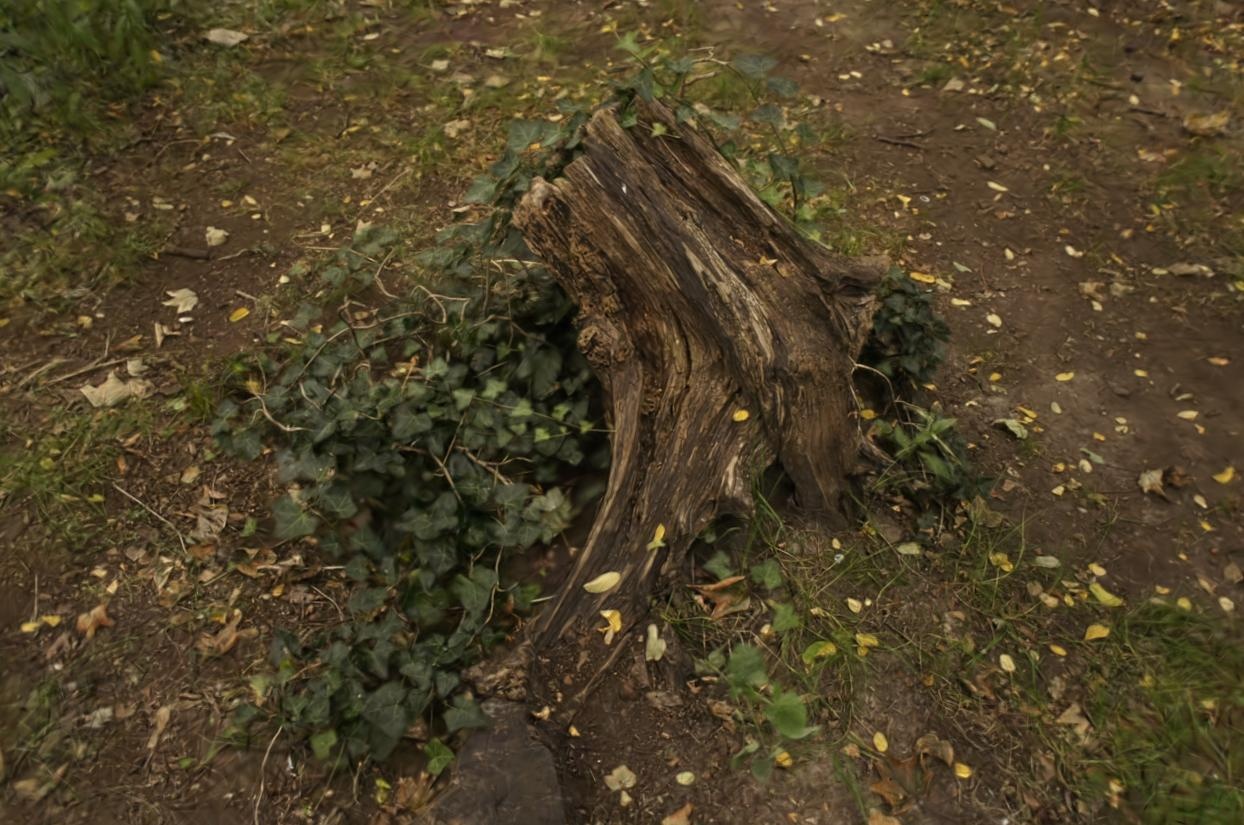}               &
        \includegraphics[width=0.155\textwidth]{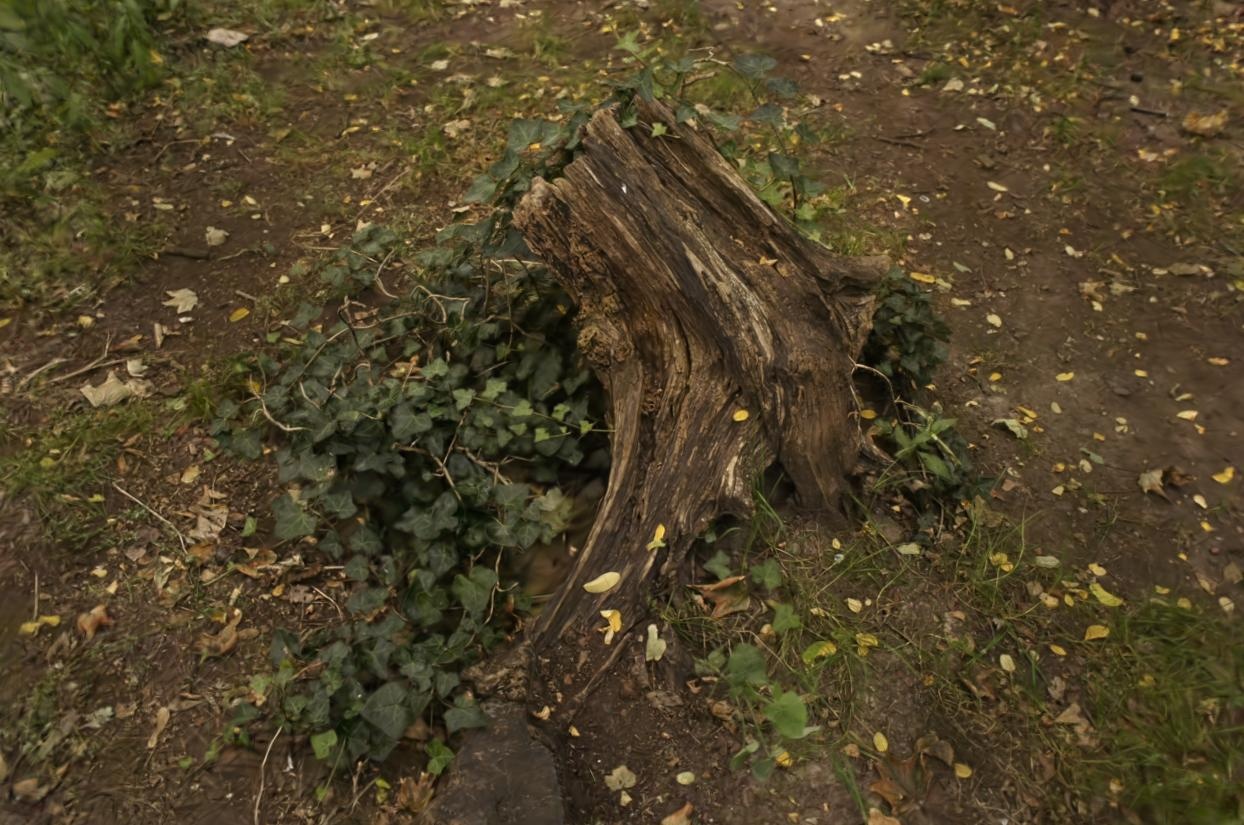}            &
        \includegraphics[width=0.155\textwidth]{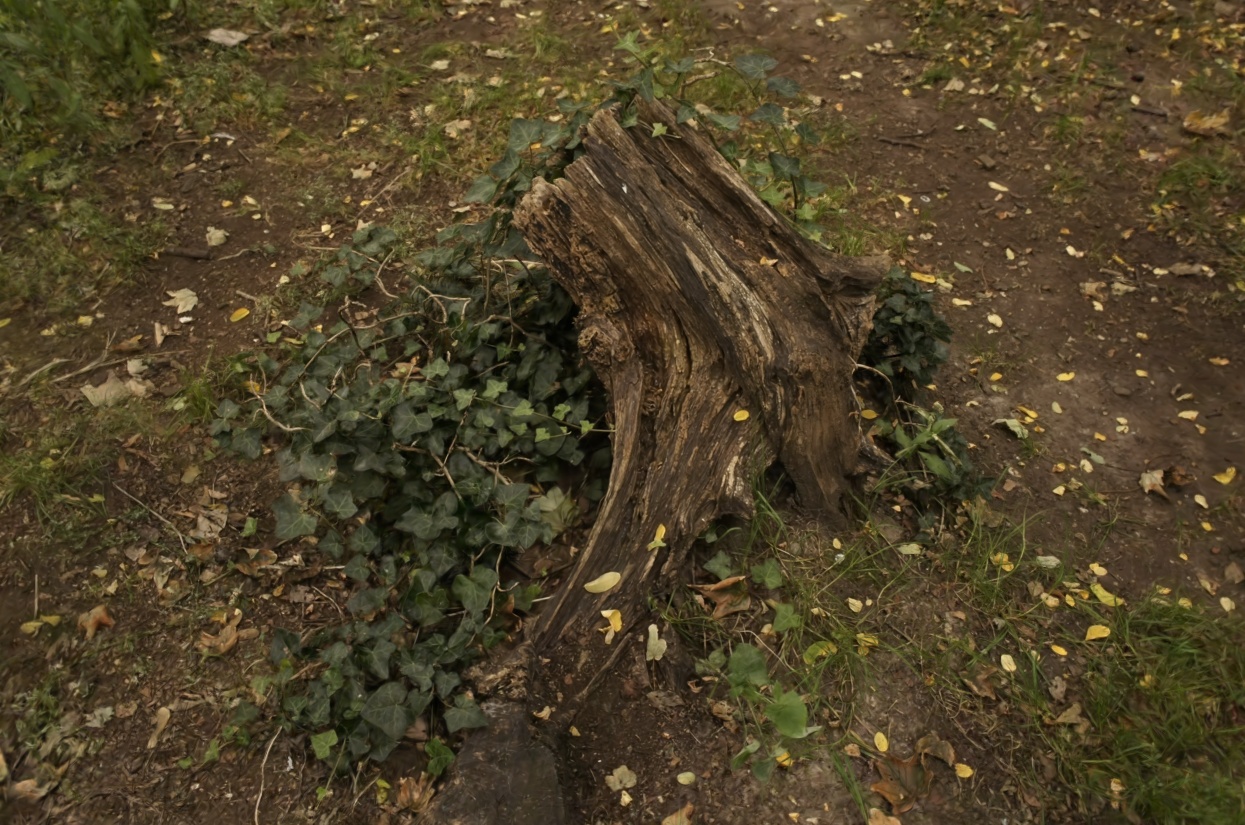}      &
        \includegraphics[width=0.155\textwidth]{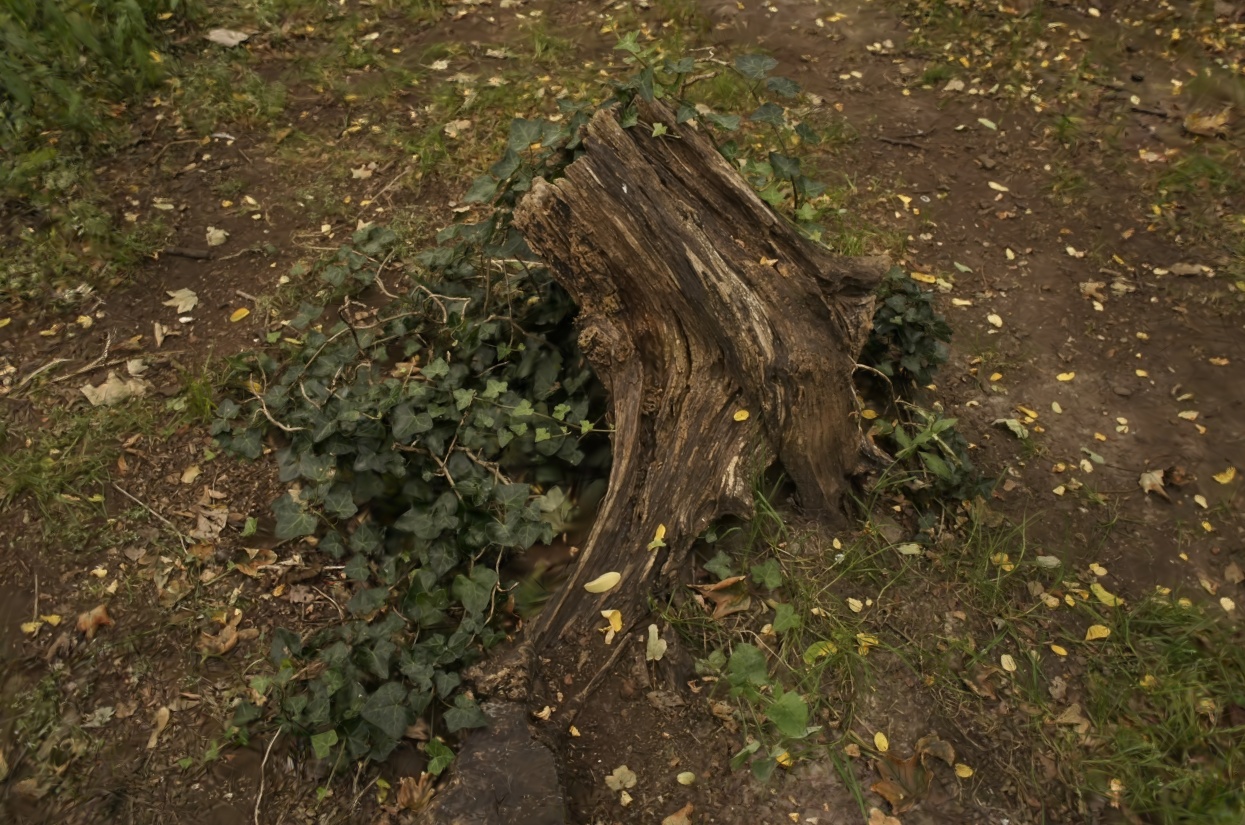} &
        \includegraphics[width=0.155\textwidth]{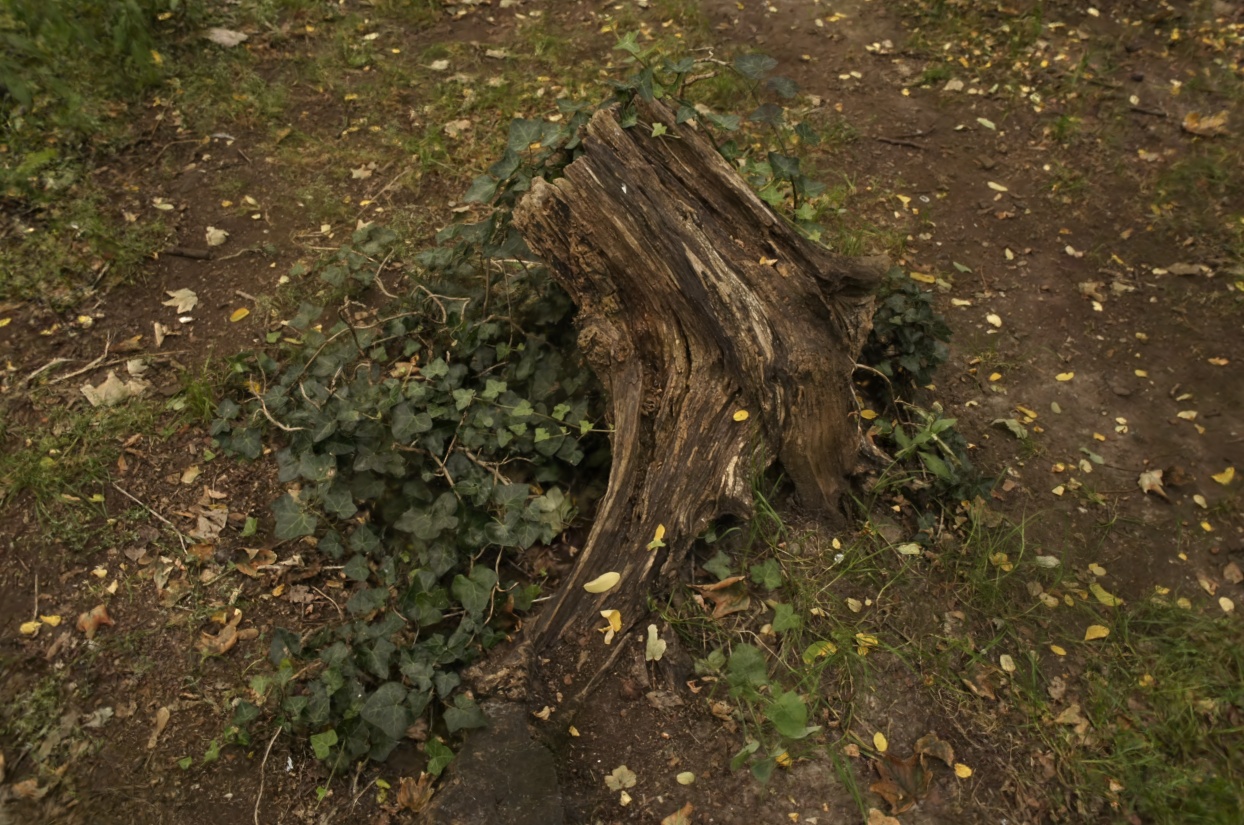}         \\
\end{tabular}
    \caption{Qualitative results of multi-scale training and multi-scale testing on the multi-scale Mip-NeRF 360 dataset. All images are rendered at $\times 1$ resolution.}
    \label{fig:qualitative_multi}
\end{figure*}

\begin{table*}[ht!]
    \centering
    \caption{Single-scale training ($\times 1$ resolution) and multi-scale testing results on the multi-scale Blender dataset. The best results are highlighted in \textbf{bold}.}
    \label{tab:stmt1_blender_scene}
    \resizebox{\textwidth}{!}{
        \begin{tabular}{l|cccccccc|c}
            \toprule
            \multirow{2}{*}{} & \multicolumn{9}{c}{PSNR $\uparrow$} \\
                                    & chair          & drums          & ficus          & hotdog         & lego           & materials      & mic            & ship           & Avg.           \\
            \midrule
            TensoRF
                                    & 32.57          & 25.63          & 31.70          & 35.16          & 32.05          & 28.80          & 31.77          & 29.30          & 30.87          \\
            3DGS
                                    & 26.59          & 21.24          & 26.25          & 28.85          & 25.59          & 23.06          & 24.81          & 23.15          & 24.94          \\
            Scaffold-GS
                                    & 26.82          & 21.14          & 25.93          & 29.62          & 25.33          & 23.36          & 24.28          & 22.93          & 24.92          \\
            2DGS
                                    & 23.04          & 20.52          & 25.99          & 28.25          & 24.47          & 22.61          & 22.75          & 22.69          & 23.79          \\
            3D-HGS
                                    & 26.74          & 21.53          & 26.33          & 29.27          & 25.71          & 23.38          & 25.19          & 23.34          & 25.19          \\
            \midrule
            Tri-MipRF
                                    & 32.61          & 24.11          & 28.09          & 34.94          & 30.67          & 28.05          & 31.08          & 27.09          & 29.58          \\
\bottomrule
        \end{tabular}
    }
    \resizebox{\textwidth}{!}{
        \begin{tabular}{l|cccccccc|c}
            \toprule
            \multirow{2}{*}{} & \multicolumn{9}{c}{SSIM $\uparrow$} \\
                                    & chair          & drums          & ficus          & hotdog         & lego           & materials      & mic            & ship           & Avg.           \\
            \midrule
            TensoRF
                                    & 0.975          & 0.941          & 0.980          & 0.982          & 0.974          & 0.964          & 0.982          & 0.914          & 0.964          \\
            3DGS
                                    & 0.914          & 0.850          & 0.920          & 0.931          & 0.887          & 0.882          & 0.913          & 0.829          & 0.891          \\
            Scaffold-GS
                                    & 0.918          & 0.847          & 0.922          & 0.948          & 0.892          & 0.893          & 0.910          & 0.828          & 0.895          \\
            2DGS
                                    & 0.879          & 0.821          & 0.908          & 0.915          & 0.864          & 0.863          & 0.879          & 0.805          & 0.867          \\
            3D-HGS
                                    & 0.917          & 0.856          & 0.921          & 0.936          & 0.891          & 0.889          & 0.920          & 0.835          & 0.896          \\
            \midrule
            Tri-MipRF
                                    & 0.978          & 0.915          & 0.959          & 0.978          & 0.963          & 0.956          & 0.977          & 0.884          & 0.951          \\
\bottomrule
        \end{tabular}
    }
    \resizebox{\textwidth}{!}{
        \begin{tabular}{l|cccccccc|c}
            \toprule
            \multirow{2}{*}{} & \multicolumn{9}{c}{LPIPS $\downarrow$} \\
                                    & chair          & drums          & ficus          & hotdog         & lego           & materials      & mic            & ship           & Avg.           \\
            \midrule
            TensoRF
                                    & 0.040          & 0.073          & 0.028          & 0.036          & 0.037          & 0.058          & 0.039          & 0.105          & 0.052          \\
            3DGS
                                    & 0.051          & 0.095          & 0.056          & 0.042          & 0.068          & 0.060          & 0.048          & 0.128          & 0.069          \\
            Scaffold-GS
                                    & 0.051          & 0.099          & 0.057          & 0.038          & 0.067          & 0.057          & 0.053          & 0.136          & 0.070          \\
            2DGS
                                    & 0.109          & 0.120          & 0.069          & 0.058          & 0.086          & 0.070          & 0.090          & 0.156          & 0.095          \\
            3D-HGS
                                    & 0.051          & 0.089          & 0.055          & 0.040          & 0.065          & 0.057          & 0.044          & 0.124          & 0.066          \\
            \midrule
            Tri-MipRF
                                    & 0.026          & 0.090          & 0.042          & 0.029          & 0.039          & 0.055          & 0.029          & 0.131          & 0.055          \\
\bottomrule
        \end{tabular}
    }
\end{table*}

\begin{table*}[ht!]
    \centering
    \caption{Single-scale training ($\times 1/2$ resolution) and multi-scale testing results on the multi-scale Blender dataset. The best results are highlighted in \textbf{bold}.}
    \label{tab:stmt2_blender_scene}
    \resizebox{\textwidth}{!}{
        \begin{tabular}{l|cccccccc|c}
            \toprule
            \multirow{2}{*}{} & \multicolumn{9}{c}{PSNR $\uparrow$} \\
                                    & chair          & drums          & ficus          & hotdog         & lego           & materials      & mic            & ship           & Avg.           \\
            \midrule
            TensoRF
                                    & 33.74          & 25.99          & 31.27          & 35.94          & 32.80          & 28.93          & 32.75          & 30.22          & 31.45          \\
            3DGS
                                    & 28.18          & 22.34          & 26.81          & 30.17          & 26.34          & 24.29          & 25.88          & 24.79          & 26.10          \\
            Scaffold-GS
                                    & 28.28          & 22.10          & 25.76          & 30.77          & 25.96          & 24.35          & 25.29          & 24.17          & 25.84          \\
            2DGS
                                    & 24.62          & 22.14          & 27.39          & 29.99          & 25.83          & 24.44          & 25.62          & 24.82          & 25.61          \\
            3D-HGS
                                    & 28.60          & 22.51          & 26.91          & 30.55          & 26.56          & 24.60          & 26.23          & 24.84          & 26.35          \\
            \midrule
            Tri-MipRF
                                    & 32.81          & 24.41          & 29.22          & 35.11          & 31.32          & 28.74          & 31.42          & 28.07          & 30.14          \\
\bottomrule
        \end{tabular}
    }
    \resizebox{\textwidth}{!}{
        \begin{tabular}{l|cccccccc|c}
            \toprule
            \multirow{2}{*}{} & \multicolumn{9}{c}{SSIM $\uparrow$} \\
                                    & chair          & drums          & ficus          & hotdog         & lego           & materials      & mic            & ship           & Avg.           \\
            \midrule
            TensoRF
                                    & 0.979          & 0.943          & 0.979          & 0.983          & 0.977          & 0.963          & 0.984          & 0.917          & 0.966          \\
            3DGS
                                    & 0.932          & 0.884          & 0.941          & 0.946          & 0.909          & 0.910          & 0.931          & 0.860          & 0.914          \\
            Scaffold-GS
                                    & 0.936          & 0.878          & 0.933          & 0.957          & 0.910          & 0.912          & 0.927          & 0.851          & 0.913          \\
            2DGS
                                    & 0.905          & 0.868          & 0.932          & 0.934          & 0.892          & 0.897          & 0.913          & 0.845          & 0.898          \\
            3D-HGS
                                    & 0.935          & 0.886          & 0.940          & 0.950          & 0.913          & 0.914          & 0.936          & 0.861          & 0.917          \\
            \midrule
            Tri-MipRF
                                    & 0.977          & 0.917          & 0.968          & 0.979          & 0.966          & 0.958          & 0.978          & 0.890          & 0.954          \\
\bottomrule
        \end{tabular}
    }
    \resizebox{\textwidth}{!}{
        \begin{tabular}{l|cccccccc|c}
            \toprule
            \multirow{2}{*}{} & \multicolumn{9}{c}{LPIPS $\downarrow$} \\
                                    & chair          & drums          & ficus          & hotdog         & lego           & materials      & mic            & ship           & Avg.           \\
            \midrule
            TensoRF
                                    & 0.034          & 0.071          & 0.035          & 0.033          & 0.033          & 0.062          & 0.033          & 0.096          & 0.049          \\
            3DGS
                                    & 0.040          & 0.077          & 0.044          & 0.036          & 0.059          & 0.052          & 0.041          & 0.112          & 0.058          \\
            Scaffold-GS
                                    & 0.043          & 0.089          & 0.052          & 0.036          & 0.064          & 0.052          & 0.045          & 0.123          & 0.063          \\
            2DGS
                                    & 0.089          & 0.099          & 0.053          & 0.047          & 0.075          & 0.058          & 0.054          & 0.126          & 0.075          \\
            3D-HGS
                                    & 0.039          & 0.077          & 0.044          & 0.034          & 0.056          & 0.050          & 0.038          & 0.111          & 0.056          \\
            \midrule
            Tri-MipRF
                                    & 0.032          & 0.093          & 0.043          & 0.032          & 0.040          & 0.056          & 0.029          & 0.120          & 0.056          \\
\bottomrule
        \end{tabular}
    }
\end{table*}

\begin{table*}[ht!]
    \centering
    \caption{Single-scale training ($\times 1/4$ resolution) and multi-scale testing results on the multi-scale Blender dataset. The best results are highlighted in \textbf{bold}.}
    \label{tab:stmt4_blender_scene}
    \resizebox{\textwidth}{!}{
        \begin{tabular}{l|cccccccc|c}
            \toprule
            \multirow{2}{*}{} & \multicolumn{9}{c}{PSNR $\uparrow$} \\
                                    & chair          & drums          & ficus          & hotdog         & lego           & materials      & mic            & ship           & Avg.           \\
            \midrule
            TensoRF
                                    & 33.45          & 25.48          & 29.92          & 35.36          & 31.48          & 28.64          & 32.39          & 29.84          & 30.82          \\
            3DGS
                                    & 28.54          & 22.37          & 25.36          & 30.28          & 25.12          & 24.09          & 26.29          & 24.98          & 25.88          \\
            Scaffold-GS
                                    & 28.06          & 21.77          & 24.15          & 30.37          & 24.56          & 23.66          & 25.15          & 23.57          & 25.16          \\
            2DGS
                                    & 25.06          & 23.36          & 27.58          & 31.03          & 26.07          & 25.56          & 28.15          & 26.08          & 26.61          \\
            3D-HGS
                                    & 29.07          & 22.62          & 25.61          & 30.51          & 25.53          & 24.46          & 26.51          & 25.02          & 26.17          \\
            \midrule
            Tri-MipRF
                                    & 31.60          & 24.06          & 27.62          & 34.20          & 29.92          & 27.87          & 30.54          & 28.07          & 29.23          \\
\bottomrule
        \end{tabular}
    }
    \resizebox{\textwidth}{!}{
        \begin{tabular}{l|cccccccc|c}
            \toprule
            \multirow{2}{*}{} & \multicolumn{9}{c}{SSIM $\uparrow$} \\
                                    & chair          & drums          & ficus          & hotdog         & lego           & materials      & mic            & ship           & Avg.           \\
            \midrule
            TensoRF
                                    & 0.970          & 0.926          & 0.968          & 0.977          & 0.967          & 0.954          & 0.979          & 0.897          & 0.955          \\
            3DGS
                                    & 0.935          & 0.897          & 0.944          & 0.951          & 0.899          & 0.915          & 0.944          & 0.861          & 0.918          \\
            Scaffold-GS
                                    & 0.935          & 0.884          & 0.931          & 0.954          & 0.892          & 0.909          & 0.933          & 0.836          & 0.909          \\
            2DGS
                                    & 0.915          & 0.905          & 0.952          & 0.956          & 0.910          & 0.928          & 0.952          & 0.870          & 0.923          \\
            3D-HGS
                                    & 0.938          & 0.898          & 0.944          & 0.952          & 0.905          & 0.919          & 0.945          & 0.860          & 0.920          \\
            \midrule
            Tri-MipRF
                                    & 0.957          & 0.901          & 0.951          & 0.969          & 0.946          & 0.934          & 0.968          & 0.872          & 0.937          \\
\bottomrule
        \end{tabular}
    }
    \resizebox{\textwidth}{!}{
        \begin{tabular}{l|cccccccc|c}
            \toprule
            \multirow{2}{*}{} & \multicolumn{9}{c}{LPIPS $\uparrow$} \\
                                    & chair          & drums          & ficus          & hotdog         & lego           & materials      & mic            & ship           & Avg.           \\
            \midrule
            TensoRF
                                    & 0.045          & 0.101          & 0.050          & 0.046          & 0.052          & 0.075          & 0.036          & 0.111          & 0.064          \\
            3DGS
                                    & 0.045          & 0.077          & 0.043          & 0.042          & 0.077          & 0.062          & 0.041          & 0.118          & 0.063          \\
            Scaffold-GS
                                    & 0.052          & 0.098          & 0.058          & 0.045          & 0.085          & 0.068          & 0.050          & 0.134          & 0.074          \\
            2DGS
                                    & 0.096          & 0.091          & 0.046          & 0.046          & 0.082          & 0.059          & 0.043          & 0.125          & 0.074          \\
            3D-HGS
                                    & 0.044          & 0.080          & 0.044          & 0.041          & 0.076          & 0.061          & 0.040          & 0.118          & 0.063          \\
            \midrule
            Tri-MipRF
                                    & 0.057          & 0.122          & 0.069          & 0.057          & 0.069          & 0.090          & 0.051          & 0.133          & 0.081          \\
\bottomrule
        \end{tabular}
    }
\end{table*}

\begin{table*}[ht!]
    \centering
    \caption{Single-scale training ($\times 1/8$ resolution) and multi-scale testing results on the multi-scale Blender dataset. The best results are highlighted in \textbf{bold}.}
    \label{tab:stmt8_blender_scene}
    \resizebox{\textwidth}{!}{
        %
    }
\end{table*}

\begin{table*}[ht!]
    \centering
    \caption{Multi-scale training and multi-scale testing results on the multi-scale Blender dataset. The best results are highlighted in \textbf{bold}.}
    \label{tab:mtmt_blender_scene}
    \resizebox{\textwidth}{!}{
        %
    }
\end{table*}

\begin{table*}[ht!]
    \centering
    \caption{Single-scale training ($\times 1$ resolution) and multi-scale testing results on the multi-scale Mip-NeRF 360 dataset. The best results are highlighted in \textbf{bold}.}
    \label{tab:stmt1_mipnerf360_scene}
    \resizebox{\textwidth}{!}{
        %
    }
\end{table*}

\begin{table*}[ht!]
    \centering
    \caption{Single-scale training ($\times 1/2$ resolution) and multi-scale testing results on the multi-scale Mip-NeRF 360 dataset. The best results are highlighted in \textbf{bold}.}
    \label{tab:stmt2_mipnerf360_scene}
    \resizebox{\textwidth}{!}{
        %
    }
\end{table*}

\begin{table*}[ht!]
    \centering
    \caption{Single-scale training ($\times 1/4$ resolution) and multi-scale testing results on the multi-scale Mip-NeRF 360 dataset. The best results are highlighted in \textbf{bold}.}
    \label{tab:stmt4_mipnerf360_scene}
    \resizebox{\textwidth}{!}{
        %
    }
\end{table*}

\begin{table*}[ht!]
    \centering
    \caption{Single-scale training ($\times 1/8$ resolution) and multi-scale testing results on the multi-scale Mip-NeRF 360 dataset. The best results are highlighted in \textbf{bold}.}
    \label{tab:stmt8_mipnerf360_scene}
    \resizebox{\textwidth}{!}{
        %
    }
\end{table*}

\begin{table*}[ht!]
    \centering
    \caption{Multi-scale training and multi-scale testing results on the multi-scale Mip-NeRF 360 dataset. The best results are highlighted in \textbf{bold}.}
    \label{tab:mtmt_mipnerf360_scene}
    \resizebox{\textwidth}{!}{
        %
    }
\end{table*}

\end{document}